# Individual health–disease phase diagrams for disease prevention based on machine learning

Kazuki Nakamura[1,2], Eiichiro Uchino[1], Noriaki Sato[1], Ayano Araki[1], Kei Terayama[1,3], Ryosuke Kojima[1], Koichi Murashita[4], Ken Itoh[5], Tatsuya Mikami[6], Yoshinori Tamada[6] and Yasushi Okuno[1,*]

[1] Department of Biomedical Data Intelligence, Graduate School of Medicine, Kyoto University, Kyoto, 606-8507, Japan.

[2] Research and Business Development Department, Kyowa Hakko Bio Co., Ltd., Tokyo, 100-0004, Japan.

[3] Graduate School of Medical Life Science, Yokohama City University, Kanagawa, 230-0045, Japan.

[4] Center of Innovation Research Initiatives Organization (The Center of Healthy Aging Innovation), Graduate School of Medicine, Hirosaki University, Aomori, 036-8562, Japan.

[5] Department of Stress Response Science, Graduate School of Medicine, Hirosaki University, Aomori, 036-8562, Japan.

[6] Innovation Center for Health Promotion, Graduate School of Medicine, Hirosaki University, Aomori, 036-8562, Japan.

## Abstract

Early disease detection and prevention methods based on effective interventions are gaining attention. Machine learning technology has enabled precise disease prediction by capturing individual differences in multivariate data. Progress in precision medicine has revealed that substantial heterogeneity exists in health data at the individual level and that complex health factors are involved in the development of chronic diseases. However, it remains a challenge to identify individual physiological state changes in cross-disease onset processes because of the complex relationships among multiple biomarkers. Here, we present the health–disease phase diagram (HDPD), which represents a personal health state by visualizing the boundary values of multiple biomarkers that fluctuate early in the disease progression process. In HDPDs, future onset predictions are represented by perturbing multiple biomarker values while accounting for dependencies among variables. We constructed HDPDs for 11 non-communicable diseases (NCDs) from a longitudinal health checkup cohort of 3,238 individuals, comprising 3,215 measurement items and genetic data. Improvement of biomarker values to the non-onset region in HDPD significantly prevented future disease onset in 7 out of 11 NCDs. Our results demonstrate that HDPDs can represent individual physiological states in the onset process and be used as intervention goals for disease prevention.

* Correspondence to Yasushi Okuno

Author email addresses:

Kazuki Nakamura (nakamura.kazuki.88m@st.kyoto-u.ac.jp), Yasushi Okuno (okuno.yasushi.4c@kyoto-u.ac.jp).

# Introduction

Early detection of disease signs and preemptive prevention of onset based on effective interventions are essential for reducing chronic diseases[1]. Non-communicable diseases (NCDs) are non-infectious chronic diseases such as cardiovascular diseases and diabetes mellitus, which are caused by various health-related factors[2]. NCDs are the leading cause of death worldwide and cause cumulative economic loss[3,4]. Recent advances in machine learning technology have contributed to the prediction of disease onset by precisely capturing individual differences in multivariate health data[5–7]. For example, onset prediction models for NCDs such as chronic kidney disease (CKD) and diabetes have been constructed using health data by capturing biomarkers that fluctuate early individual disease progression processes[8–11].

Progress in precision medicine has revealed substantial heterogeneity in the physiological factors and progression processes associated with diseases at the individual level[12–15]. Biomarker values, such as physiological measurements or laboratory tests, can indicate a high risk for certain individuals, even when the values fall within the reference range of clinical guidelines[16]. It has been reported that biomarker values should be evaluated using 'personal thresholds' based on individual-level variability rather than 'population thresholds' based on population-level variability to realize early diagnosis in some diseases[17,18]. Changes in the physiological state from health to disease are complex and have not yet been comprehensively explained. In particular, it is unclear how diverse the changes in an individual physiological state are in complex biomarkers during the progression process to the onset of various diseases.

In this study, we propose a health–disease phase diagram (HDPD) to represent individual health status at each time point by visualizing the boundary values of multiple biomarkers that fluctuate early in the progression process of NCDs. We constructed HDPDs by plotting onset/non-onset predictions for data points where the input biomarkers of disease onset prediction models were virtually perturbed. A personal boundary was defined as the boundary that separates the future onset and non-onset states in the HDPD. Physiological state changes from health to disease can be determined by examining the boundaries of multiple biomarkers. Since multiple biomarkers have complex interdependencies, we introduced the projected-multivariate individual conditional expectation (p-mICE), which extends ICE[19–21], an explainable artificial intelligence (XAI) method, to multiple dimensions while accounting for multivariate dependence.

We applied the HDPDs to Iwaki Health Promotion Project (IHPP) data, which are annual health checkup data with a wide range of health-related factors in healthy people. We constructed HDPDs to represent personal boundaries based on predictive models of 11 NCDs, including CKD and diabetes, and examined individual physiological states in disease progression processes. Furthermore, we retrospectively analyzed the clinical validity of personal boundaries as improvement goals for early intervention at the individual level.

# Results

To investigate biomarkers using HDPDs based on NCDs prediction models, we analyzed the health checkup dataset collected in the IHPP. More than 3,000 items of extensive medical checkup data for participants have been acquired annually in the IHPP. An overview of the IHPP dataset is provided in Table 1 and Supplementary Table 1. In this study, we built onset-prediction models for multiple NCDs using the IHPP dataset. Subsequently, HDPDs were plotted using disease onset prediction models.

**Construction of disease onset prediction models**

We constructed onset prediction models for the following 11 NCDs: arteriosclerosis, CKD, chronic obstructive pulmonary disease (COPD), dementia, diabetes mellitus, dyslipidemia, hypertension, locomotive syndrome (LS), obesity, knee osteoarthritis (KOA), and osteopenia. Records of the IHPP dataset were labeled based on the disease onset criteria to predict the current disease onset or future onset within 1-3 years for each disease. After applying pre-processing to the missing values and discrete variables, the explanatory variables were reduced using recursive feature elimination (RFE)[22], a popular method for selecting reasonable variables in a machine learning model. RFE was performed with five-fold cross-validation in the training data. The explanatory variables were reduced to 25 for each disease, which had little impact on the predictive scores of the validation data (Fig. 1 and Supplementary Figs. 1–4). After determining the hyperparameters and predict-probability threshold values, we built disease onset prediction models using these explanatory variables. For the machine learning algorithm, we used XGBoost[23], which is generally capable of high-performance prediction owing to its non-linear decision process. The model scores for the test data are summarized in Table 2 and Supplementary Table 2. Since scores were retained in the future-onset prediction models, we focused on the onset prediction models within three years to investigate early markers in the subsequent analysis.

In addition, we performed an experiment to demonstrate the construction of appropriate prediction models. We analyzed whether the prediction models captured the early signs of NCDs by comparing three groups of records: the non-onset group within three years, the onset group within three years, and the current onset group (Supplementary Fig. 5). Regarding the explanatory variables, the distributions were different between the three groups. For example, in addition to HbA1c and blood glucose, which are part of the diagnostic criteria, waist circumference, white blood cell count, and pulse wave velocity (PWV) in the onset group within three years exhibited intermediate values compared to the other two groups in diabetes (Supplementary Fig. 5e).

**Construction of p-mICE-based HDPD for disease onset prediction models**

Subsequently, we plotted personal HDPDs based on p-mICE. A schematic representation of p-mICE-based HDPD construction is shown in Fig. 2 and detailed in the Methods section. In this paper, we referred to intervention variables as explanatory variables that were explicitly varied on the plot, that is, variables assigned to the x- and y-axes. The p-mICE process consists of two steps. In Step 1, the intervention variables were perturbed from the original record, corresponding to an extension of the conventional ICE[19]. In Step 2, the non-intervention variables are projected to follow the actual data distribution. The nearest neighbor records to the intervention variable-perturbed records were retrieved from the training dataset using the $k$-nearest neighbor ($k$-NN) algorithm[24] (Step 2-i). With the exception of intervention variables, all variable values were changed using these records to project the actual data distribution (Step 2-ii). Finally, the HDPD was constructed by plotting the predicted values of the p-mICE-applied data points. The personal boundary was defined as the boundary between the onset and non-onset predictions in the HDPD.

We constructed p-mICE-based HDPDs at the individual level using NCDs onset-prediction models to represent personal boundaries. All bivariate combinations of the explanatory variables used in the predictive model were investigated as intervention variables. We first describe the detailed results regarding HDPD of three typical lifestyle diseases to be prevented: CKD, diabetes, and hypertension. Subsequently, a wider analysis of the 11 NCDs was performed.

**Evaluation of p-mICE-based HDPDs in CKD prediction model**

In the CKD onset prediction model, the estimated glomerular filtration rate (eGFR), an indicator of current kidney function, was selected as the most contributing variable, followed by other variables regarding known risk factors for CKD, such as diabetes, hypertension, arteriosclerosis, and obesity-related items[25,26] (Supplementary Fig. 3b).

Examples of HDPDs are shown in Fig. 3a–c and Supplementary Fig. 6. In the HDPDs of eGFR and blood glucose level, eGFR had a dominant effect, but a decrease in blood glucose levels that were too high improved the predicted value. Since low eGFR values are a well-known risk factor for CKD, we conducted a more detailed examination of the upper limit eGFR values that yielded CKD onset prediction based on eGFR–blood glucose HDPDs (Supplementary Fig. 7a). The eGFR values that define the personal boundaries vary between records (Fig. 3d). Higher immunoglobulin A (IgA) levels were associated with a higher risk in the HDPDs of IgA and total cholesterol (Fig. 3a–b). High serum IgA levels may reflect the involvement of IgA nephropathy, which is a common cause of CKD[27]. Some HDPDs exhibited no personal boundaries (Supplementary Fig. 6), whereas others displayed non-linear personal boundaries (Fig. 3a–c). In addition, there were cases in which the known knowledge and risk value were reversed up and down, such as total cholesterol (Fig. 3a). Even for the same participant, the personal boundaries in the HDPDs changed over time (Fig. 3c). In this example, a clinically valid combination of decreased eGFR and increased blood glucose level forms a boundary in the direction of onset. Before and after the baseline year (X + 2), when the participant entered the future onset area, the actual measured values of eGFR and blood glucose showed fluctuations in the surrounding area. The personal boundaries also changed from year to year, depending on the state of the other parameters, indicating that the target values in the intervention changed.

We also constructed phase diagrams based on the perturbation of only the intervention variables (only Step 1 in p-mICE was applied), which we call 2d-ICE-based HDPDs (Supplementary Fig. 8). The p-mICE-based substitution might relax unnatural input values, such as low eGFR values (< 60), which is a diagnostic level of CKD. In addition, we examined efficient boundary construction using active learning. We performed uncertainty sampling[28] for active learning, and 50 data points were searched in each HDPD instead of a full search. It was shown that the personal boundary could be visualized with fewer search points using active learning (Supplementary Fig. 9).

We then investigated individual differences in personal boundaries. First, we examined individual differences in the intervention variables that contributed to personal boundaries. For each record, we constructed HDPDs for all bivariate combinations of the explanatory variables. The feature contribution in the HDPDs was introduced as the proportion of the HDPDs where the variable contributed to boundary formation (Supplementary Fig. 10, detailed in the Methods section). We performed hierarchical clustering of feature contributions from the records (Fig. 3e). The variables contributing to personal boundaries in the HDPD differed between individuals. Most records showed a high eGFR contribution to HDPD. eGFR was of the highest importance in the prediction model (Supplementary Fig. 3b). It was noted that many other variables also contributed to boundary formation in some clusters. Disease onset prediction labels tended to be concentrated in specific clusters. In contrast, it was difficult to capture meaningful differences in the clustering and onset label patterns between records from the clustering of the original explanatory variables (Fig. 3f).

**Evaluation of p-mICE-based HDPDs in diabetes prediction model**

The most important variables in the diabetes onset prediction model were HbA1c and serum blood glucose level. HbA1c is part of the diagnostic criteria and represents a long-term blood glucose trend[29]. As with CKD, known risk factors, such as triglycerides and obesity-related items, were selected for other variables. Immunoglobulin M (IgM), which has been reported to be associated with prediabetes in recent years, was also ranked high[30]. Examples of HDPDs are presented in Supplementary Figs. 11–13. The values of the explanatory variables that defined personal boundaries and boundary patterns were diverse between the records. Although the most typical cases were those in which high HbA1c levels were dominant, serum blood glucose levels combined with HbA1c determined the personal boundaries in some cases (Supplementary Fig. 11b). A broad analysis of the lower limit of HbA1c, which yielded diabetes onset predictions, showed that HbA1c values defining personal boundaries varied across the records (Supplementary Fig. 14). The various individual characteristics resulted in no boundaries in the combinations of the specific variables (Supplementary Fig. 11a–b). Supplementary Fig. 11e shows an example of personal boundary changes in triglycerides and waist circumference over time. Triglycerides and waist circumference are more easily intervened than HbA1c, and serum blood glucose is linked to the diagnostic criteria. In this example, the participant was judged to have developed diabetes in the year (X + 5). A personal boundary mainly due to triglycerides appeared in the base year (X + 1), which was predicted to be a high onset risk. Subsequently, an improvement was observed once, and the high-risk range disappeared. However, in year (X + 4), a severe boundary appeared that required significant reductions in both the variables.

Subsequently, we analyzed the variables that contributed to the boundary formation among the records. The feature contribution to HDPDs was calculated for each variable and then hierarchically clustered across records (Supplementary Figs. 15d and 16e). Although the variables contributing to boundary formation varied by record, HbA1c and blood glucose levels had dominant effects. These variables also presented high feature importance in the onset prediction model (Supplementary Fig. 3e). Supplementary Fig. 17d shows the clustering results with the original explanatory variables.

**Evaluation of p-mICE-based HDPDs in hypertension prediction model**

In the hypertension onset prediction model, clinically appropriate variables such as PWVs, obesity-related items, and diabetes-related items were selected in addition to systolic blood pressure (SBP), diastolic blood pressure (DBP), and mean arterial pressure (MAP)[31–34]. PWVs are indicators of arteriosclerosis and have been associated with the development of hypertension[32]. Examples of HDPDs are presented in Supplementary Figs. 18–20. In the HDPD of SBP and left brachial-ankle PWV (LbaPWV), even if SBP was not high at a certain point, the risk of onset was higher as arteriosclerosis progressed (Supplementary Fig. 18d). These results suggest that personal boundaries depend on the presence of multiple biomarkers. In the HDPD of leg fat percentage and HbA1c, the dominant variable in boundary formation was distinct between records (Supplementary Fig. 18b–d), suggesting that individual differences were expressed.

Subsequently, hierarchical clustering was performed on the feature contribution of each variable (Supplementary Figs. 15f and 16g). Records were separated into two clusters. One cluster showed a wide contribution of variables, mainly blood pressure and limb blood pressure, to boundary formation. The other cluster contained a small number of contributing variables. The relationship between the original explanatory variables and predicted labels is shown in Supplementary Fig. 17f.

**Cross-disease analysis of p-mICE-based HDPDs**

We analyzed biomarkers across diseases based on their personal boundaries. First, we investigated the differences in personal boundary patterns (Supplementary Table 3). HDPDs can be divided into three patterns depending on the number of intervention variables that contribute to the personal boundary (Supplementary Fig. 10). The first is a no-boundary pattern, in which the entire region consists of only onset or non-onset points. The second is a univariate boundary pattern, in which the boundary is defined by only one variable. The third is a bivariate boundary pattern, in which both variables form the boundary. The proportions of bivariate boundary patterns to the univariate and bivariate boundary patterns were calculated (Fig. 4a and Supplementary Figs. 21–22). Personal boundaries tend to be formed by multiple variables rather than single variables.

In addition, biomarker values defining personal boundaries were analyzed for dominant biomarkers in the onset-prediction models. For biomarkers with the highest feature importance in each disease prediction model, the value that yielded disease onset prediction at the furthest value from the abnormal values was extracted based on the HDPD for each record (Fig. 4b). Even when focusing on disease criteria, such as eGFR in CKD and HbA1c in diabetes, there was variation in the value of the boundary across samples. For other items, a wider distribution of boundaries was observed.

**Validation of personal boundaries as individual intervention goals**

We evaluated whether personal boundaries would work as individual-level intervention goals for disease prevention. A retrospective time-series analysis was conducted on the HDPDs of the predicted disease-onset records (Fig. 5a). Among the predicted disease onset records, some did not develop the disease in the future. We defined a prevented onset record as a record that did not develop onset in the future and an actual onset record as a developed record, which corresponded to false positive (FP) and true positive (TP), respectively (Table 2). We hypothesized that disease onset could be prevented if the future values of the predicted disease onset record improved beyond the personal boundaries. We introduced the improved HDPD proportion as the proportion of HDPDs in which future biomarker values would be improved beyond personal boundaries. We calculated the improved HDPD proportion based on the HDPDs of all combinations of intervention variables for each predicted-onset record. The improved HDPD proportion was significantly higher in the onset prevention records than in the actual onset records for 7 of the 11 NCDs (Fig. 5b and Supplementary Fig. 23). It tended to be higher for the remaining NCDs. Furthermore, we evaluated whether rational changes in non-intervention variables, dependent on perturbations of intervention variables, were simulated using p-mICE. We calculated the distance from the 2d-ICE-based data point to the actual future record because of the effect of the projection step in p-mICE. Compared with the 2d-ICE-applied data point, it was demonstrated that the p-mICE-applied data points were significantly closer to the actual future records in most NCDs (Table 3).

# Discussion

In this study, we constructed multiple disease onset prediction models based on a health checkup dataset using a machine-learning method. The model performance statistics, such as area under the curve (AUC), were comparable to those of the well-studied conventional models for CKD and diabetes prediction[8–11]. In addition, items consistent with previous studies and clinical knowledge were selected as important features of the predictions (Fig. 1). Future onset prediction models were constructed to capture the characteristics that may fluctuate during the prodromal stage of the disease onset process (Supplementary Figs. 2–5). In addition to age, common lifestyle-related disease factors, such as blood glucose and triglycerides, contributed to many models. Interestingly, serum immunoglobulins also ranked high in many conditions, which may be related to their involvement in lifestyle, metabolic abnormalities, and pathophysiology such as chronic inflammation[35,36]. In addition to diagnostic markers, early markers are involved in disease onset. The boundary values for the prevention of disease onset should be interpreted based on multiple biomarkers.

HDPDs were constructed based on p-mICE, using predictive models to represent multivariate-based personal boundaries. In most cases in the 11 NCDs analyses, personal boundaries were formed with bivariate intervention variables instead of univariate variables (Fig. 4a and Supplementary Figs. 21–22). The application of p-mICE reduced the no-boundary patterns and increased the bivariate boundary patterns compared with simple 2d-ICE (Supplementary Table 3). As multiple health-related factors cause NCDs, it is reasonable to assume that personal boundaries are formed by multivariate contributions. There were cases where health improvement beyond the personal boundary was not achieved with univariate intervention but was possible with bivariate intervention (Supplementary Table 4). High-dimensional personal boundary construction will lead to the establishment of more effective intervention goals that can reflect the patient's preferences. However, computational costs with a full search in higher-dimensional phase diagrams exponentially increase when constructing personal boundaries. Active learning is promising for solving this problem (Supplementary Figs. 9, 13, and 20). Compared to a full search, a personal boundary can be efficiently formed with fewer search points using active learning.

Individual differences in biomarkers contributing to the personal boundaries and boundary values of the biomarkers were investigated. The feature contribution of each explanatory variable to the HDPDs differed at the individual level (Fig. 3e and Supplementary Fig. 15). Variables with high feature importance in prediction models generally had a high feature contribution in the individual-level HDPD analysis, for example, eGFR in CKD, HbA1c and blood glucose in diabetes, and blood pressure and limb blood pressure-related items in hypertension. These variables are also known disease-related biomarkers, suggesting that a high feature contribution to personal boundaries is reasonable. In the hierarchical clustering of feature contributions in HDPDs among records, the disease onset prediction labels tended to concentrate in specific clusters. In contrast, such differences were unclear in the clustering of the explanatory variable values (Fig. 3f and Supplementary Fig. 17). Thus, personal boundaries may reflect the individual disease onset process based on the complex relationships between biomarkers captured by disease onset prediction models, which are difficult to interpret using popular XAI methods in the medical field[19,37–42]. In addition, biomarker values that defined personal boundaries and boundary patterns in HDPDs differed among individuals, even in the same intervention variables. For biomarkers closely related to disease definitions (e.g., eGFR for CKD), biomarker values that defined personal boundaries were comprehensively investigated among the records (Fig. 3d and Supplementary Fig 14). Considering eGFR as an example of a quantitative expression of personal boundary, individual differences were observed in the risk range of eGFR for the onset of CKD within three years, ranging from near the diagnostic criterion of 60 mL/min/1.73 m$^2$ to larger values, depending on the status of other measurements. In other diseases, the boundaries also appeared for a wide range of biomarker values (Fig. 4b), suggesting the importance of our approach in considering multiple variables. In some HDPDs, the personal boundaries were inconsistent with previous clinical knowledge (Fig. 3a). The model may capture individual differences that have not been previously clarified by expressing a non-linear relationship between variables. Furthermore, personal boundaries, which should be the intervention goal, change over time at the individual level (Fig. 3c and Supplementary Fig. 11e). Although HDPDs can change drastically over time, making further quantitative evaluations of changes difficult in this study, HDPDs may represent an individual's physiological state.

Furthermore, as another notable point in our study, we retrospectively demonstrated that the improved HDPD proportion in the prevented onset records was significantly higher than in the actual onset records for most diseases (Fig. 5b). This result indicates that health improvements beyond personal boundaries can prevent disease onset. Thus, based on machine learning models, personal boundaries can be reasonable individual treatment goals. Intervention goals were set for most of the onset-predicted records based on personal boundaries (Supplementary Table 4). For practical use, it is assumed that, for example, the clinician and patient will consult to select a set of variables based on the patient's preferences and clinical practice, followed by confirming HDPDs to set improvement goals for the overall prevention of multiple diseases (Supplementary Fig. 24). To the best of our knowledge, this is the first study to demonstrate the validity of intervention goals based on disease onset prediction models using an actual health dataset. In addition, the p-mICE-applied data points were closer to the actual future records than were the 2d-ICE-applied data points (Table 3). Because the biomarkers are dependent on each other in medical data, data points where only specific biomarker values are virtually perturbed should deviate from the valid health data distribution and not be suitable as input for prediction models[43,44]. This result suggests that p-mICE can simulate changes in multiple biomarker values to approach the data distribution. As an idea similar to this study, a method using neighborhood data points obtained by $k$-NN has been reported to improve the robustness of prediction for data that deviate from the training data distribution in the research field of adversarial defense, which supports the validity of our approach[45].

This study had some limitations. First, the IHPP dataset was obtained from a healthy population. While this enabled the construction of future onset prediction models for multiple diseases, it was not possible to acquire a sufficient number of onset-label records for some diseases. There was no significant difference in the time-series analysis of some NCDs (Fig. 5b), which may be partly due to the small number of onset-label records of patients with dementia. Second, we performed a retrospective time-series analysis to verify the validity of personal boundaries as a treatment goal. However, prospective studies on personal boundary-based interventions and their outcomes are required to precisely demonstrate their clinical utility. The prospective verification of XAI technologies on medical machine learning models is yet to be addressed, but is essential for disease prevention and should be examined in future studies. The third is the problem of verifying whether p-mICE can project virtually perturbed health data points onto the actual data distribution. Since multi-item medical checkup data were collected annually in the IHPP, we could evaluate the validity of p-mICE using the future biomarker values of the records. It is difficult to evaluate p-mICE using other datasets because of the lack of public datasets with such characteristics. Although we have validated p-mICE in multiple diseases using the IHPP dataset, future studies should examine other time-series health checkup datasets.

In summary, a comprehensive analysis of the early biomarkers of NCDs was conducted using longitudinal, multi-item cohort data from healthy individuals. We introduced p-mICE-based HDPDs to construct personal boundaries. Personal boundaries differed individually regarding the biomarker values defining boundaries, boundary patterns, and feature contributions. Time-series changes in HDPDs could be relevant to disease onset processes. HDPDs will provide opportunities to understand the individual physiological states expressed in machine learning models in the medical field. Furthermore, we proved that personal boundaries could be reasonable intervention goals through a retrospective analysis using actual health data. This result suggests that HDPDs are beneficial for planning treatment goals for preventing NCDs in precision medicine. Future studies should consider prospective validation of HDPDs for clinical applications.

## Methods

### Dataset

In this study, we used the data acquired by the IHPP. This project has obtained a wide range of health checkup data that comprises the molecular biology, physiology, biochemistry, personal lifestyle, and socio-environmental aspects of residents of Iwaki district, Hirosaki City, Aomori Prefecture, Japan. We targeted 13,922 health checkup records of adult participants across 14 years (2005–2018), comprising 3,215 measurements and genetic data. The number of unique participants was 3,238 because the same person had participated for several years (Table 1). This study was approved by the Ethics Committee of Hirosaki University School of Medicine (annual approval, the latest approval number: 2019-009) and conducted according to the recommendations of the Declaration of Helsinki. All the participants provided written informed consent.

### Disease onset criteria and each disease dataset generation

We determined the presence of 11 NCDs for each participant at each time point using disease-related items in the dataset. The criteria for each disease are shown in the Supplementary Methods. Based on the presence of disease labels, we constructed datasets to predict current disease status or future disease onset. Datasets for future disease onset included records with future records eligible for determining the condition in the targeted onset periods but without the current disease. A record was labeled as positive if the disease state was determined at least once within each period.

## Data pre-processing and feature selection

We performed the following pre-processing steps on each disease dataset. Ambiguous items, such as answers to questionnaires and items with missing values of 25% or more, were excluded. Regarding single-nucleotide polymorphism (SNP) data, we created a list of known SNPs from the GWAS catalog by querying the name of each disease[46]. A total of 5,489 SNPs were identified. Subsequently, we created a table representing the genotypes of each SNP in each patient. SNPs known to be relevant to each disease were selected to construct each disease dataset (Supplementary Table 5). Missing values were complemented with median values, and one-hot encoding was applied for categorical variables. Participants were randomly divided into training (80%) and test (20%) groups. Each disease dataset was divided into training and test data based on participants' training/test assignment (Supplementary Table 6).

To construct disease onset prediction models with meaningful variables, we performed an XGBoost[23]-based RFE[22] to reduce the explanatory variables using the training data. RFE was performed with five-fold cross-validation by the participants, and the explanatory variables were reduced to 25.

## Construction of disease onset prediction models

We used XGBoost[23], which is based on a gradient boosting decision tree algorithm, to create the disease onset-prediction models. Hyperparameters of the models and threshold values in predict-probabilities used to identify onset/non-onset were determined with five-fold cross-validation split by participants in the training data. Predict-probability values that provided the best f1 scores in the validation data of each fold were acquired, and the threshold values were determined using the median values of five folds. Finally, disease onset prediction models were constructed using all the training data.

## p-mICE-based HDPD construction

We constructed phase diagrams in which onset/non-onset predictions were plotted by varying multiple input biomarkers in the disease prediction models. We introduced p-mICE to simulate changes in the input biomarkers for the prediction model. A schematic representation of p-mICE is shown in Fig. 2. In this paper, we referred to intervention variables as explanatory variables that were explicitly varied on the plot, that is, variables assigned to the x- and y-axes in the phase diagram. p-mICE consists of two steps. In Step 1, the intervention variables were perturbed from the original record, corresponding to an extension of the conventional ICE to two dimensions (2d-ICE). The range of variation in the continuous variables was restricted to reduce the impact of outliers and unrealistic variations. The range excluding the upper and lower 0.5th percentile in the training data was treated as the domain of the biomarker. Biomarker values increased or decreased from their original values within the domain in 4% increments, up to 40% of the domain. One-hot categorical variables were varied with a value of 0 or 1. In Step 2, the non-intervention variables are projected to follow the data distribution. Based on the $k$-NN algorithm, $k$ neighbor records of the 2d-ICE-applied data points were retrieved from the original record and any training data consistent with the predicted labels. When applying the $k$-NN algorithm, continuous variables were normalized and discrete variables were stratified. Finally, the non-intervention variable values were replaced with a weighted average of the $k$-nearest neighbor record values. The hyperparameters (number of $k$ and weights of neighbor records) are detailed in the following section.

HDPD was constructed by plotting the predicted values of the p-mICE-applied data points. The prediction probability at each data point was calculated using the prediction model. If the value was equal to or greater than the threshold value determined in the previous section, the data point was plotted as an onset point in the HDPD; otherwise, it was plotted as a non-onset point. In each record, all bivariate combinations of explanatory variables whose original values were not missing were examined as intervention variables in p-mICE. In addition, HDPD using simple 2d-ICE-applied data points was constructed as a supplementary experiment.

In the case of HDPD construction using active learning, we applied uncertainty sampling[28], which is an effective active learning method based on label uncertainty. We obtained onset/non-onset predictive labels for ten data points as the initial search points. Then, data points with high label uncertainty were selected using label spreading[47] (clamping factor = 0.2), and the actual predictive labels at the data points were calculated using the prediction model. This process was repeated until a defined number of data-point labels was obtained.

**Evaluation of biomarker values defining personal boundaries**

We investigated the differences in biomarker values that defined the personal boundaries among the records. Regarding the dominant biomarker in the prediction model, we analyzed the upper or lower limit values that yielded disease onset prediction. In the detailed assessment of the HDPD of the CKD prediction model, the upper limit of the eGFR value that induced onset prediction was analyzed. Because a low eGFR is generally a risk factor for CKD, whether even high eGFR values could induce disease onset prediction was investigated from a preventive perspective. HDPDs with eGFR and blood glucose as intervention variables were used to determine the upper limit of the eGFR values (Supplementary Fig. 7a). First, the upper limit of the eGFR value, which induced CKD onset prediction, was obtained for each blood glucose level. When all eGFR values induced a non-onset prediction for a given blood glucose level, the minimum eGFR value in the HDPD was acquired as the upper limit. Subsequently, the range of the upper limit values was retrieved for each record. The analysis was also performed using a diabetes prediction model. For the HDPDs with HbA1c and blood glucose as intervention variables, the lower limit of HbA1c that induced the prediction of diabetes onset was investigated (Supplementary Fig. 14a). As high levels are a risk factor for HbA1c, the lower limit of HbA1c was determined. In the cross-disease analysis, the upper limits were analyzed for biomarkers with low values as risk factors and the lower limits for biomarkers with high values (Supplementary Table 7).

**Evaluation of personal boundary patterns in HDPDs**

We categorized HDPDs into three patterns based on personal boundary patterns: no-boundary, univariate boundary, and bivariate boundary (Supplementary Fig. 10). In the no-boundary pattern, all regions were either onset or non-onset, and no personal boundaries were formed (Supplementary Fig. 10a–b). Neither intervention variable contributed to a boundary in the no-boundary pattern. Only a single variable contributed to the personal boundary in the univariate boundary pattern, resulting in a boundary parallel to the x- or y-axis (Supplementary Fig. 10c–d). Other patterns in which personal boundaries were formed based on both intervention variables were categorized as bivariate boundary patterns (Supplementary Fig. 10e–f). Both variables contribute to a personal boundary in a bivariate boundary pattern.

The feature contribution to HDPDs was calculated as the proportion of HDPDs, where the variable contributed to boundary formation. After assigning the contribution of the missing variables in the records as zeros, hierarchical clustering of the feature contributions among the records was performed using the Ward method. For comparison, we performed hierarchical clustering on the original values of the explanatory variables after clipping the upper and lower 0.5th percentile values in the training data as outliers. In addition, the proportion of bivariate boundary patterns in the boundary-formed HDPDs patterns was calculated for each record or bivariate intervention variable. The detailed equation is as follows: (#bivariate boundary patterns) / (#bivariate boundary patterns + #univariate boundary patterns).

**Validation of personal boundaries as individual intervention goals**

We conducted a retrospective time-series analysis using the HDPDs of actual onset records and prevented onset records, which corresponded to the TP and FP records in the confusion matrix of the prediction models (Fig. 5a). The actual future values of the intervention variables over the next three years for the same participants were plotted on the HDPD. Then, it was determined whether each HDPD contained plotted points in the non-onset area, that is, improvement of the intervention variables beyond the personal boundary. Improved HDPDs were regarded as HDPDs, where improvement across personal boundaries was observed. We calculated the improved HDPD proportions for all HDPDs in each record. No-boundary pattern HDPDs were excluded from the analysis because it was difficult to prevent disease onset by any intervention in the HDPD.

**Hyperparameter tuning of p-mICE**

The hyperparameters of p-mICE were determined using predicted disease onset records. The basic strategy was to apply p-mICE to substitute non-intervention variables to get closer to actual future records. The future values of the intervention variables were regarded as perturbed intervention variables, and non-intervention variables were substituted in the projection step of the p-mICE. The distance to actual future records was calculated for each of the 2d-ICE-applied data points and p-mICE-applied data points in the normalized variable space. The approached distance was defined as the difference between these distances, which represented the distance approached to the actual record by substituting non-intervention variables in the p-mICE. We calculated the mean distance based on all the HDPDs for each record. Two hyperparameters must be tuned in the p-mICE: the number of $k$ in $k$-NN and the weights of neighbor records in non-intervention variable substitution. The hyperparameters were tuned using the actual and prevented onset records of the training data to optimize the approach distance. In this situation, future records of the participants themselves were excluded from the training data to search the neighborhood data using $k$-NN. The weights of neighbor records, which decayed exponentially with the distance between two data samples, were adopted because they yielded the closest to the actual record for most diseases (Supplementary Fig. 25). Weights that decay exponentially with distance have also been applied in non-linear dimensionality reduction techniques to represent the relationships between data points[48]. For each disease, the number $k$ was determined to maximize the mean/standard deviation of the approached distance since a large variance is undesirable in a medical context (Supplementary Table 8).

## Data availability

The health checkup data used in this study were collected from the Iwaki Health Promotion Project (IHPP) and transferred to a secure data center with restricted access controls in a de-identified format. The de-identified data are available from the Hirosaki University School of Medicine (contact via e-mail: coi@hirosaki-u.ac.jp) for academic research purposes only and for researchers who meet the criteria for access to the data. Researchers must be approved by the research ethics review committees of both the Hirosaki University School of Medicine and their affiliations. Three months are required for the access request to be approved. All other data are included in this article or available from the corresponding author upon reasonable request.

## Code availability

All codes used in this study are available at https://github.com/clinfo/p-mice.

## Acknowledgements

This work was supported by JST, the Center of Innovation Program (JPMJCE1302 and JPMJCA2201), and Kyowa Hakko Bio Co. Ltd. Kazuki Nakamura thanks Kazushi Shoji, Takashi Ishida, and Kay Moriwaki, employees of Kyowa Hakko Bio Co., Ltd., for their generous support.

## Author information

**Affiliations**

Department of Biomedical Data Intelligence, Graduate School of Medicine, Kyoto University, Kyoto, 606-8507, Japan.
Kazuki Nakamura, Eiichiro Uchino, Noriaki Sato, Ayano Araki, Kei Terayama, Ryosuke Kojima and Yasushi Okuno

Research and Business Development Department, Kyowa Hakko Bio Co., Ltd., Tokyo, 100-0004, Japan.
Kazuki Nakamura

Graduate School of Medical Life Science, Yokohama City University, Kanagawa, 230-0045, Japan.
Kei Terayama

Center of Innovation Research Initiatives Organization (The Center of Healthy Aging Innovation), Graduate School of Medicine, Hirosaki University, Aomori, 036-8562, Japan.
Koichi Murashita

Department of Stress Response Science, Graduate School of Medicine, Hirosaki University, Aomori, 036-8562, Japan.
Ken Itoh

Innovation Center for Health Promotion, Graduate School of Medicine, Hirosaki University, Aomori, 036-8562, Japan.
Tatsuya Mikami and Yoshinori Tamada

**Contributions**

K.N. contributed to data analysis. E.U., N.S., A.A., K.T., R.K., and Y.O. provided substantial contributions to data analysis. K.N., E.U., and N.S. drafted the manuscript. K.T. and K.N. conceived the study. K.M., K.I., T.M., and Y.T. designed the study for the data acquisition.

**Corresponding authors**

Correspondence to Yasushi Okuno.

# Ethics declarations

**Competing interests**

Kazuki Nakamura is an employee of Kyowa Hakko Bio Co. Ltd. The authors declare that they have no conflict of interest.

# Tables

**Table 1. Subject characteristics at first-time participation.**

| Participant characteristics | Male | Female |
| --- | --- | --- |
| Number | 1,281 | 1,957 |
| Age (years) | 50.2 ± 16.2 | 51.5 ± 16.0 |
| Height (cm) | 168.0 ± 7.1 | 154.7 ± 6.7 |
| Body weight (kg) | 66.7 ± 11.2 | 54.1 ± 8.9 |
| BMI (kg/m$^2$) | 23.6 ± 3.3 | 22.6 ± 3.6 |
| eGFR (mL/min/1.73m$^2$) | 81.6 ± 16.5 | 83.0 ± 16.8 |
| SBP (mmHg) | 128.2 ± 18.1 | 124.2 ± 19.6 |
| DBP (mmHg) | 77.2 ± 12.0 | 73.5 ± 11.6 |
| Blood glucose (mg/dL) | 90.4 ± 21.5 | 87.3 ± 19.7 |
| HbA1c (%) | 5.6 ± 0.7 | 5.6 ± 0.7 |
| Triglyceride (mg/dL) | 120.0 ± 98.4 | 82.3 ± 49.8 |
| Total cholesterol (mg/dL) | 197.0 ± 33.7 | 203.2 ± 35.5 |
| HDL cholesterol (mg/dL) | 57.7 ± 15.1 | 66.1 ± 15.2 |
| LDL cholesterol (mg/dL) | 112.9 ± 29.8 | 111.9 ± 30.4 |
| Medicine use | | |
|     Diabetes | 28 (3.7%) | 20 (1.9%) |
|     Hyperlipidemia | 37 (4.9%) | 75 (7.0%) |
|     Hypertension | 121 (15.9%) | 172 (16.0%) |
| Disease | | |
|     Arteriosclerosis | 287 (22.6%) | 402 (20.7%) |
|     CKD | 33 (2.7%) | 48 (2.5%) |
|     COPD | 57 (8.1%) | 34 (3.4%) |
|     Dementia | 23 (3.7%) | 22 (2.4%) |
|     Diabetes | 119 (9.3%) | 109 (5.6%) |
|     Dyslipidemia | 671 (64.6%) | 914 (59.6%) |
|     Hypertension | 483 (37.8%) | 682 (34.9%) |
|     KOA | 48 (13.3%) | 161 (30.0%) |
|     LS | 13 (2.7%) | 46 (7.4%) |
|     Obesity | 381 (29.9%) | 440 (22.6%) |
|     Osteopenia | 303 (24.3%) | 751 (39.1%) |

Means ± standard deviations are presented. BMI, body mass index; eGFR, estimated glomerular filtration rate; SBP, systolic blood pressure; DBP, diastolic blood pressure; HDL, high-density lipoprotein; LDL, low-density lipoprotein; CKD, chronic kidney disease; COPD, chronic obstructive pulmonary disease; KOA, knee osteoarthritis; LS, locomotive syndrome.

**Table 2. Performance of disease onset prediction models within three years.**

| Disease | AUC | TP | FN | FP | TN | Threshold |
|---|---|---|---|---|---|---|
| Arteriosclerosis | 0.898 | 66 | 41 | 122 | 1,121 | 0.226 |
| CKD | 0.871 | 19 | 25 | 84 | 1,535 | 0.070 |
| COPD | 0.668 | 16 | 37 | 58 | 1,209 | 0.119 |
| Dementia | 0.800 | 2 | 27 | 3 | 1,238 | 0.212 |
| Diabetes | 0.919 | 24 | 26 | 38 | 1,521 | 0.132 |
| Dyslipidemia | 0.751 | 54 | 38 | 98 | 281 | 0.249 |
| Hypertension | 0.799 | 101 | 69 | 139 | 709 | 0.256 |
| KOA | 0.781 | 54 | 15 | 91 | 182 | 0.181 |
| LS | 0.656 | 25 | 43 | 137 | 438 | 0.135 |
| Obesity | 0.888 | 28 | 26 | 62 | 1,211 | 0.170 |
| Osteopenia | 0.852 | 82 | 43 | 135 | 706 | 0.240 |

AUC, area under the curve; TP, true positive; FN, false negative; FP, false positive; TN, true negative; CKD, chronic kidney disease; COPD, chronic obstructive pulmonary disease; KOA, knee osteoarthritis; LS, locomotive syndrome. The threshold indicates the model's predict-probabilities that divide the onset/non-onset predictions.

**Table 3. Approached distance with p-mICE-based substitution in test data.**

| Disease | Approached distance | *p*-value | k |
|---|---|---|---|
| Arteriosclerosis | 0.072±0.107 | <0.001 | 4 |
| CKD | 0.059±0.089 | <0.001 | 6 |
| COPD | 0.053±0.093 | <0.001 | 4 |
| Dementia | 0.060±0.086 | 0.190 | 8 |
| Diabetes | 0.062±0.114 | <0.001 | 6 |
| Dyslipidemia | 0.096±0.123 | <0.001 | 6 |
| Hypertension | 0.124±0.128 | <0.001 | 4 |
| KOA | 0.118±0.108 | <0.001 | 8 |
| LS | 0.068±0.093 | <0.001 | 6 |
| Obesity | 0.040±0.074 | <0.001 | 4 |
| Osteopenia | 0.048±0.084 | <0.001 | 4 |

Statistical significance was calculated using two-sided paired *t*-test. CKD, chronic kidney disease; COPD, chronic obstructive pulmonary disease; KOA, knee osteoarthritis; LS, locomotive syndrome.

# Figures

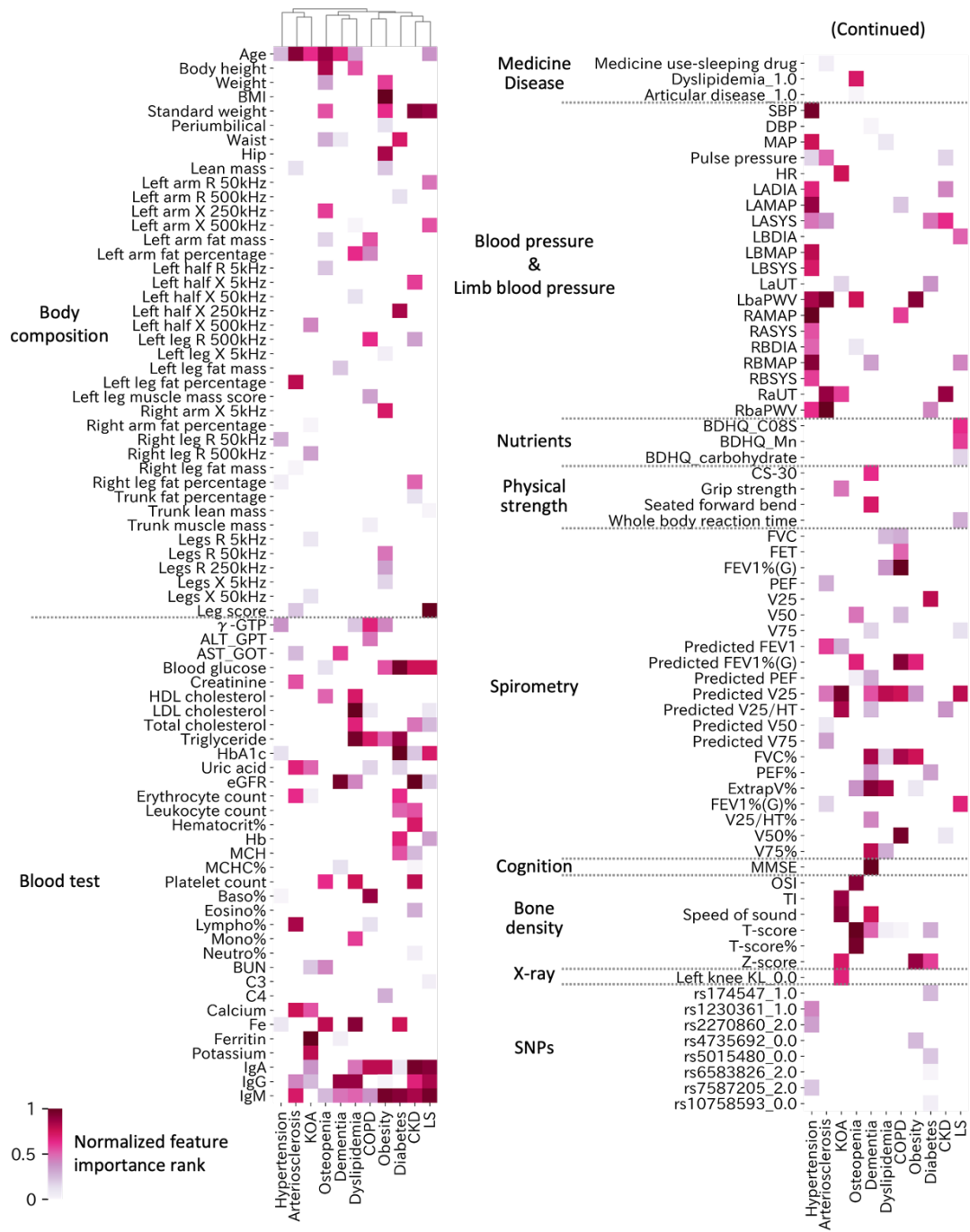

**Figure 1. Selected features for disease prediction models within three years.** Selected features for non-communicable diseases (NCDs) prediction models are mapped. Recursive feature elimination (RFE) was applied to select 25 features for each disease prediction model. RFE was performed with five-fold cross-validation, and the mean feature importance of all the features was calculated when 25 variables remained. For each disease, the selected features were normalized by the rank of feature importance. Hierarchical clustering was then performed among the diseases based on the normalized feature importance rank. CKD, chronic kidney disease; COPD, chronic obstructive pulmonary disease; KOA, knee osteoarthritis; LS, locomotive syndrome.

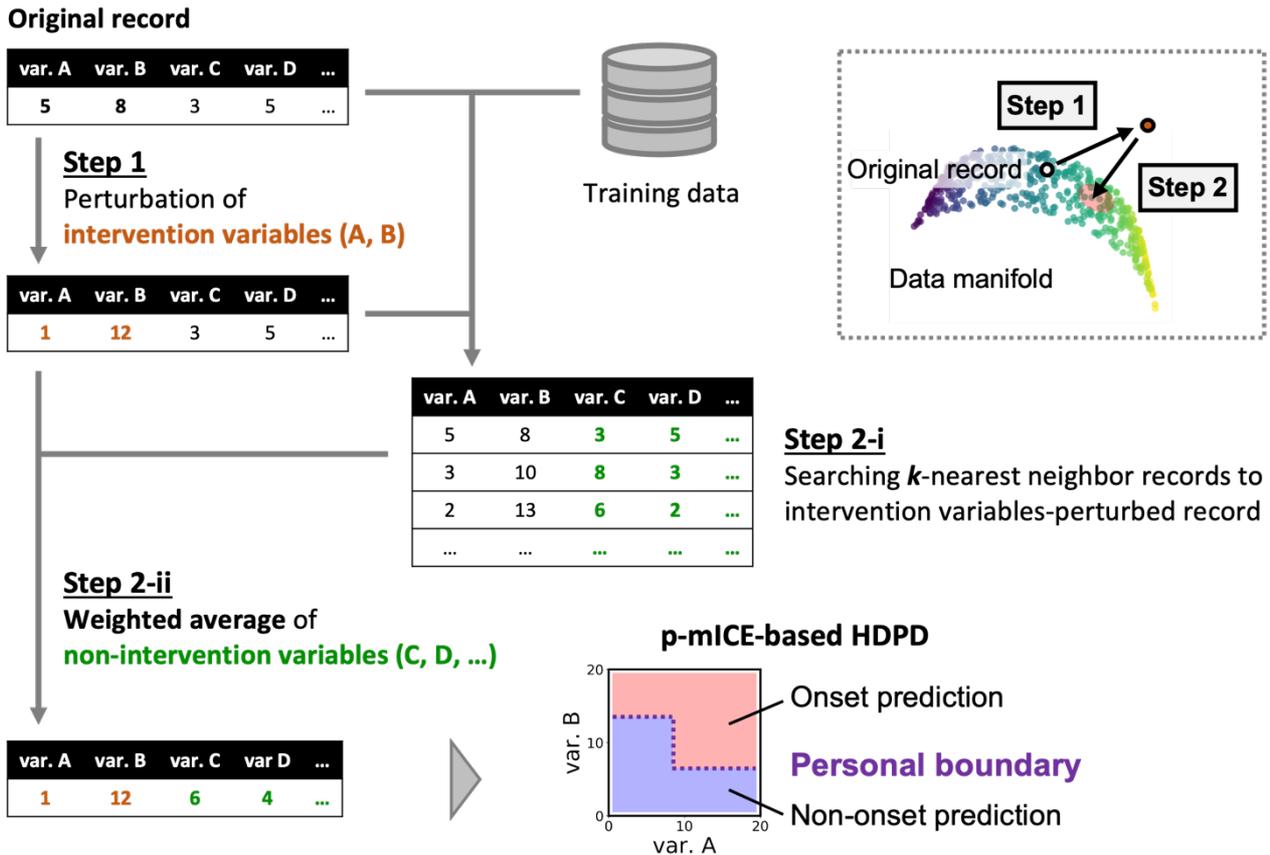

**Figure 2. Schematic representation of projected-multivariable individual conditional expectation (p-mICE) and health–disease phase diagram (HDPD) construction.** A schematic is given as an example in which variables A and B are regarded as intervention variables. The intervention variables are the explanatory variables that are explicitly perturbed when plotting the HDPD using a predictive model. The p-mICE process consists of two steps. In Step 1, intervention variables are perturbed from the original record, corresponding to an extension of the conventional ICE to two dimensions (2d-ICE). In Step 2, the non-intervention variables are projected to follow the data distribution. The nearest neighbors are searched from the training dataset and original record, followed by a weighted average of non-intervention variables. Subsequently, a phase diagram is formed by plotting the predicted values of the p-mICE-applied. The personal boundary is defined as the boundary between onset and non-onset prediction in the HDPD.

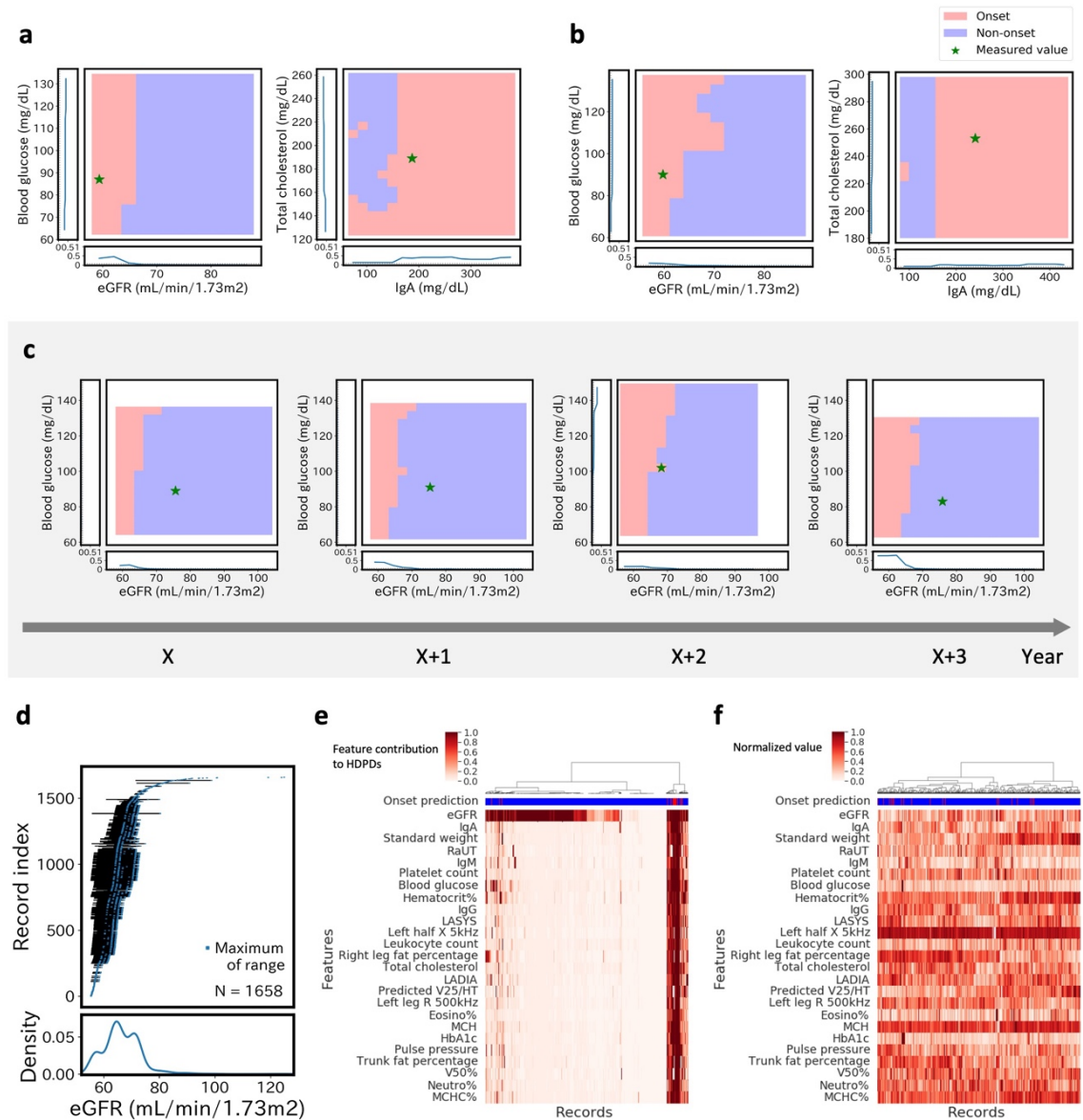

**Figure 3. p-mICE-based personal health–disease phase diagrams (HDPDs) for chronic kidney disease (CKD) prediction model.** Individual differences in the personal boundaries of the HDPD were investigated. **a, b** HDPDs of two randomly selected records. The HDPDs are shown along with the individual conditional expectation (ICE) plot. The dotted line in the ICE plot represents the onset threshold of the predicted probability in the prediction model. For each record, examples of two kinds of combinations of the intervention variables are displayed. **c** Time-series representation of personal HDPDs in a randomly selected participant. The year values were masked for privacy reasons. **d** The range of upper limit eGFR values that yield CKD onset prediction in the HDPDs. The upper limit of the eGFR value that yielded the CKD onset prediction was analyzed based on the eGFR–blood glucose phase diagram. The records are sorted by the median of the range. The maximum values of the ranges are represented in blue and summarized in the density plot. **e** Hierarchically-clustered heatmap for feature contribution to HDPDs among records. The feature contribution to HDPDs was calculated as the proportion of the HDPDs where the variable contributed to the personal boundary. **f** Hierarchically-clustered heatmap for original explanatory variables of records. Explanatory variables were normalized into a range of 0–1 before clustering. In the heatmaps, red predicted labels represent records with onset prediction and blue with non-onset.

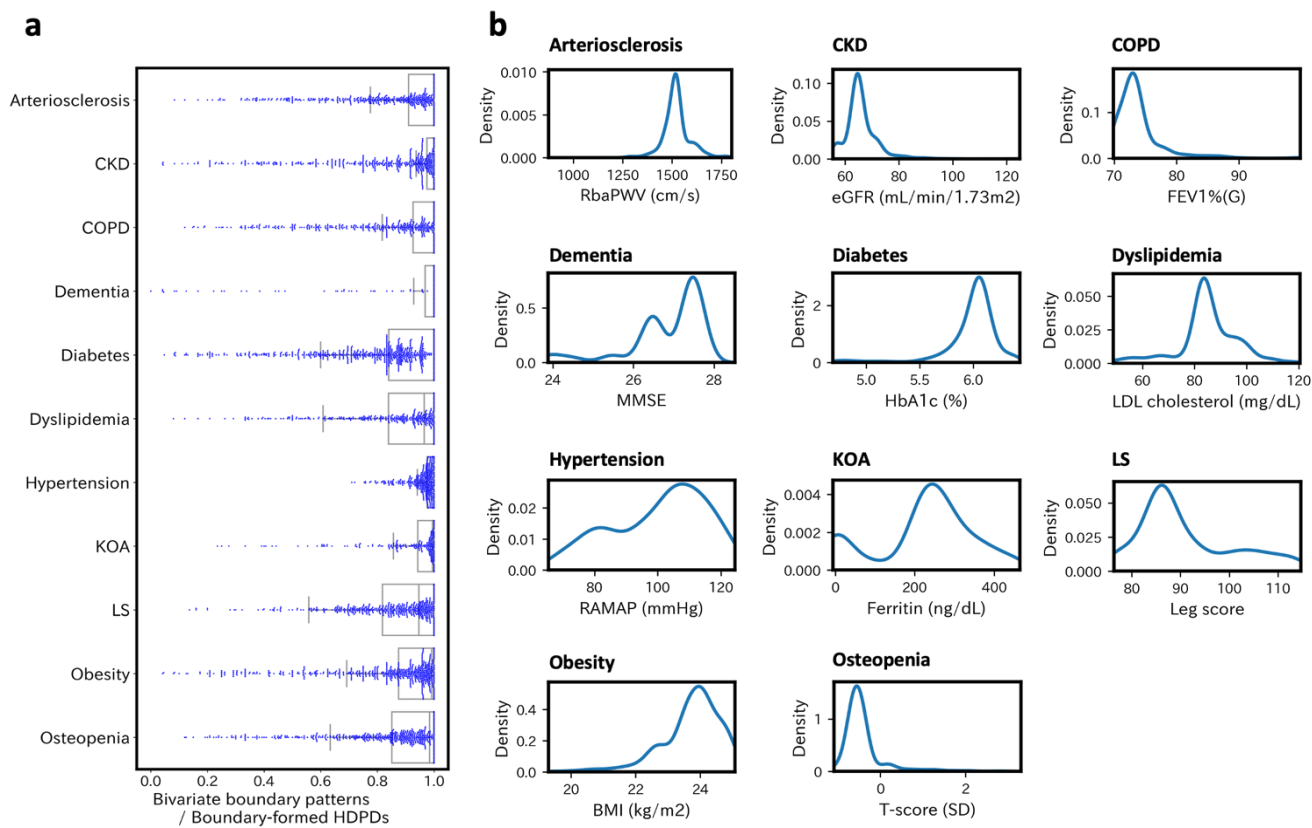

Figure 4. **Cross-disease analysis of personal boundaries. a** Proportion of bivariate boundary patterns in boundary-formed health–disease phase diagrams (HDPDs) was calculated for each record. The boundary-formed HDPD consisted of univariate and bivariate boundary patterns. In bivariate boundary patterns, both two variables contributed to forming the boundary. Each plotted point represents the record. In box-plots, the center line represents the median; box limits, upper and lower quartiles; whiskers, 1.5x interquartile range. **b** Upper or lower limit biomarker values yielding disease onset prediction in the HDPDs. For biomarkers with the highest feature importance in each disease prediction model, the value that yielded disease onset prediction at the furthest value from abnormal values was extracted based on the HDPD for each record. CKD, chronic kidney disease; COPD, chronic obstructive pulmonary disease; KOA, knee osteoarthritis; LS, locomotive syndrome.

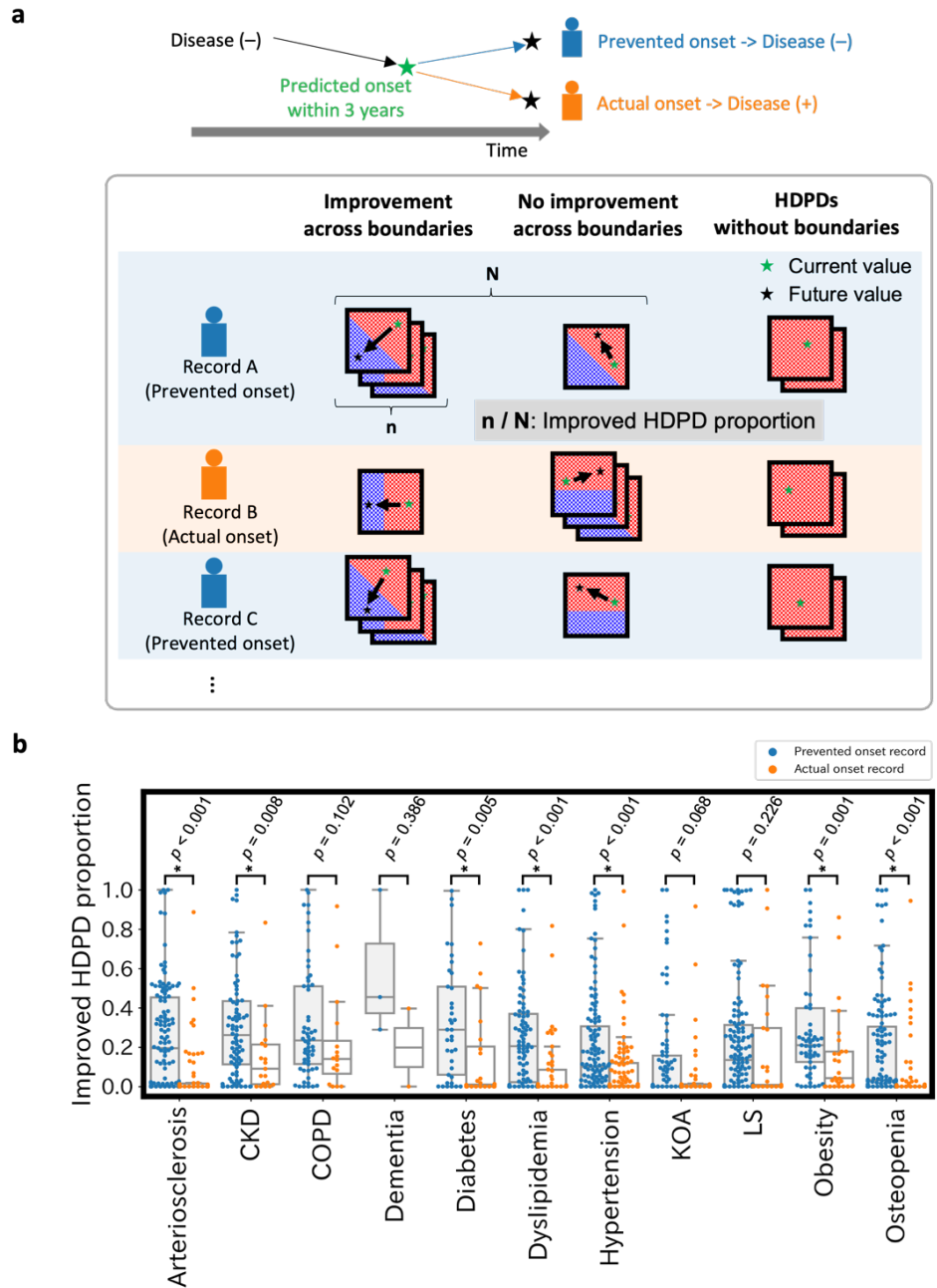

**Figure 5. Evaluation for validity of personal boundaries as intervention goals. a** Schematic representation of the retrospective time-series analysis. The predicted onset records are divided into two groups according to future onset: prevented onset records and actual onset records. It was hypothesized that the future values of the prevented onset records would tend to transition from the onset area to the non-onset area of the health–disease phase diagram (HDPD) compared to the actual onset records. We calculated the improved HDPD proportion for each predicted onset record. **b** Comparison of improved HDPD proportion between prevented onset records and actual onset records. The numbers of prevented onset records and actual onset records for each disease are shown in Table 2. Statistical significance was calculated using two-sided Wilcoxon rank-sum test. $p < 0.05$ is indicated by *. In box-plots, the center line represents the median; box limits, upper and lower quartiles; whiskers, 1.5x interquartile range. CKD, chronic kidney disease; COPD, chronic obstructive pulmonary disease; KOA, knee osteoarthritis; LS, locomotive syndrome.

# Supplementary Information

## Supplementary Methods

### Disease Criteria

Diagnostic criteria applied in this study regarding each disease are described as follows.

### Arteriosclerosis

Arterial stiffness is considered to be a high risk of cardiovascular disease (CVD) events, and brachial-ankle pulse wave velocity (baPWV) is an evaluation method in the general population. In some studies, 14–20 m/s has been proposed as baPWV cutoff, and based on these results, 18 m/s is considered as a high risk of CVD in Japanese guidelines[1]. In this study, 18 m/s was adopted as the cutoff.

### Chronic kidney disease (CKD)

Incident CKD was defined as meeting either of the following criteria: (i) estimated glomerular filtration rate (eGFR) levels <60 ml/min/1.73 m$^2$ in two consecutive times; (ii) Once eGFR < 60 is satisfied and the regression line in all measurements crossing 60 before the final measurement[2]. eGFR value was calculated using the equation for Japanese population[3].

### Chronic obstructive pulmonary disease (COPD)

The Global Initiative for Chronic Obstructive Lung Disease (GOLD) criterion of FEV1/FVC ratio < 0.70 was used as a cut-off[4].

### Dementia

Incident dementia was defined by the Mini-Mental State Examination (MMSE) ≤ 23 points, which is a common cutoff point for screening cognitive impairment[5,6], or taking medication for dementia.

### Diabetes

Incident diabetes was defined as an HbA1c level ≥ 6.5%, a fasting plasma glucose (FPG) level ≥ 126 mg/dL[7], or receiving antidiabetic treatment.

### Dyslipidemia

Diagnostic criteria of Japanese guideline[8] was used. Incident dyslipidemia was defined as LDL-cholesterol (LDL-Cho) ≥ 120 mg/dL, HDL-Cho < 40 mg/dL, triglycerides (TG) ≥ 150 mg/dL, or current medication for the disease.

### Hypertension

Incident hypertension was defined by systolic blood pressure ≥ 140 mmHg, diastolic blood pressure ≥ 90 mmHg, or receiving anti-hypertensive drugs. Although the latest ACC/AHA guidelines[9] has changed the diagnostic criteria from 140/90 mmHg to 130/80 mmHg, the Japanese guideline[10] has kept the 140/90 mmHg criteria, and the data of medical interview results or medication before 2017 was included in this study. Therefore, the conventional criteria were adopted in this study.

**Locomotive syndrome (LS)**

Locomotive syndrome is a condition of reduced mobility due to impairment of locomotive organs[11]. The 25-question Geriatric Locomotive Function Scale (GLFS-25) ≥ 16, which has been developed as early detection and diagnosis for LS[12,13], was used to definition in this study.

**Obesity**

Obesity was defined by body mass index (BMI) level ≥ 25.0 kg/m$^2$, which has been proposed and validated as a cutoff value for the diagnosis of obesity in Asian people[14].

**Knee osteoarthritis (KOA)**

The Kellgren-Lawrence (KL) classification system has been commonly used as a research tool in epidemiological studies of knee OA[15,16]. The incidence of KOA was determined by KL grade ≥ 2.

**Osteopenia**

From World Health Organization (WHO) diagnostic categories of bone mineral density, the criteria of low bone mass (osteopenia) or osteoporosis[17] was adopted. Incident osteopenia was defined as dual-energy x-ray absorptiometry (DXA) or Quantitative ultrasound (QUS) T-score < -1.0[18].

# Supplementary References

## Supplementary Tables

### Supplementary Table 1. Disease possession status of participants during all periods.

| Disease | Negative in all years | Initially positive | New onset during period | No labels in all years |
|---|---|---|---|---|
| Arteriosclerosis | 2,192 | 693 | 340 | 13 |
| CKD | 2,967 | 85 | 115 | 71 |
| COPD | 2,362 | 209 | 147 | 520 |
| Dementia | 2,392 | 110 | 73 | 663 |
| Diabetes | 2,862 | 229 | 142 | 5 |
| Dyslipidemia | 871 | 1,894 | 337 | 136 |
| Hypertension | 1,572 | 1,167 | 499 | 0 |
| KOA | 1,097 | 587 | 324 | 1,230 |
| LS | 1,802 | 226 | 186 | 1,024 |
| Obesity | 2,239 | 823 | 168 | 18 |
| Osteopenia | 1,705 | 1,069 | 427 | 37 |

N = 3,238. Some records lacked enough data to label over the entire period. CKD, chronic kidney disease; COPD, chronic obstructive pulmonary disease; KOA, knee osteoarthritis; LS, locomotive syndrome.

### Supplementary Table 2. Area under the curve (AUC) scores for disease prediction models.

| Disease | Current onset | Onset within 1 year | Onset within 2 years | Onset within 3 years |
|---|---|---|---|---|
| Arteriosclerosis | 1.000 | 0.898 | 0.887 | 0.898 |
| CKD | 0.996 | 0.912 | 0.904 | 0.871 |
| COPD | 1.000 | 0.691 | 0.645 | 0.668 |
| Dementia | 0.992 | 0.825 | 0.776 | 0.800 |
| Diabetes | 1.000 | 0.923 | 0.920 | 0.919 |
| Dyslipidemia | 1.000 | 0.737 | 0.788 | 0.751 |
| Hypertension | 1.000 | 0.806 | 0.821 | 0.799 |
| KOA | 1.000 | 0.567 | 0.709 | 0.781 |
| LS | 0.811 | 0.744 | 0.645 | 0.656 |
| Obesity | 1.000 | 0.936 | 0.908 | 0.888 |
| Osteopenia | 0.999 | 0.831 | 0.841 | 0.852 |

CKD, chronic kidney disease; COPD, chronic obstructive pulmonary disease; KOA, knee osteoarthritis; LS, locomotive syndrome.

**Supplementary Table 3. Number of HDPDs with each boundary pattern.**

| | 2d-ICE | | | p-mICE | | |
|---|---|---|---|---|---|---|
| Disease | No-boundary patterns | Univariate boundary patterns | Bivariate boundary patterns | No-boundary patterns | Univariate boundary patterns | Bivariate boundary patterns |
| Arteriosclerosis | 330,540 (92.9%) | 15,948 (4.5%) | 9,421 (2.6%) | 308,233 (86.6%) | 6,266 (1.8%) | 41,410 (11.6%) |
| CKD | 408,924 (88.7%) | 14,231 (3.1%) | 38,114 (8.3%) | 392,648 (85.1%) | 3,425 (0.7%) | 65,196 (14.1%) |
| COPD | 361,793 (92.5%) | 17,267 (4.4%) | 12,145 (3.1%) | 349,986 (89.5%) | 4,495 (1.1%) | 36,724 (9.4%) |
| Dementia | 354,744 (99.3%) | 1,602 (0.4%) | 762 (0.2%) | 354,685 (99.3%) | 650 (0.2%) | 1,773 (0.5%) |
| Diabetes | 396,110 (91.7%) | 13,303 (3.1%) | 22,577 (5.2%) | 395,293 (91.5%) | 5,727 (1.3%) | 30,970 (7.2%) |
| Dyslipidemia | 107,908 (81.0%) | 13,984 (10.5%) | 11,277 (8.5%) | 87,002 (65.3%) | 4,155 (3.1%) | 42,012 (31.5%) |
| Hypertension | 230,953 (79.7%) | 8,771 (3.0%) | 50,107 (17.3%) | 202,687 (69.9%) | 2,614 (0.9%) | 84,530 (29.2%) |
| KOA | 66,476 (73.4%) | 6,930 (7.7%) | 17,103 (18.9%) | 46,948 (51.9%) | 1,798 (2.0%) | 41,763 (46.1%) |
| LS | 131,476 (69.6%) | 25,717 (13.6%) | 31,720 (16.8%) | 104,686 (55.4%) | 8,028 (4.2%) | 76,199 (40.3%) |
| Obesity | 317,598 (88.8%) | 14,871 (4.2%) | 25,359 (7.1%) | 302,473 (84.5%) | 4,389 (1.2%) | 50,966 (14.2%) |
| Osteopenia | 243,000 (90.8%) | 14,024 (5.2%) | 10,610 (4.0%) | 219,930 (82.2%) | 5,943 (2.2%) | 41,761 (15.6%) |

CKD, chronic kidney disease; COPD, chronic obstructive pulmonary disease; KOA, knee osteoarthritis; LS, locomotive syndrome.

**Supplementary Table 4. Number of predicted onset records without health improvement toward non-onset area.**

| | | No non-onset prediction records | | |
|---|---|---|---|---|
| Disease | All records | ICE | 2d-ICE | p-mICE |
| Arteriosclerosis | 188 | 0 | 0 | 0 |
| CKD | 103 | 0 | 0 | 0 |
| COPD | 74 | 1 | 0 | 0 |
| Dementia | 5 | 1 | 0 | 0 |
| Diabetes | 62 | 4 | 0 | 0 |
| Dyslipidemia | 152 | 21 | 1 | 0 |
| Hypertension | 240 | 73 | 7 | 5 |
| KOA | 145 | 61 | 13 | 7 |
| LS | 162 | 26 | 3 | 1 |
| Obesity | 90 | 0 | 0 | 0 |
| Osteopenia | 217 | 9 | 0 | 0 |

The number of records without health improvement toward the non-onset area for any combination of intervention variables were counted. The intervention variables were univariate in ICE, while bivariate in 2d-ICE and p-mICE. ICE, individual conditional expectation; CKD, chronic kidney disease; COPD, chronic obstructive pulmonary disease; KOA, knee osteoarthritis; LS, locomotive syndrome.

**Supplementary Table 5. Number of SNPs used in each disease dataset.**

| Disease | Number of SNPs |
|---|---|
| Arteriosclerosis | 41 |
| CKD | 151 |
| COPD | 144 |
| Dementia | 5 |
| Diabetes | 476 |
| Dyslipidemia | 286 |
| Hypertension | 178 |
| KOA | 13 |
| LS | 0 |
| Obesity | 130 |
| Osteopenia | 11 |

CKD, chronic kidney disease; COPD, chronic obstructive pulmonary disease; KOA, knee osteoarthritis; LS, locomotive syndrome.

**Supplementary Table 6. Number of records in training and test datasets.**

| | Current onset | | Onset within 1 year | | Onset within 2 years | | Onset within 3 years | |
|---|---|---|---|---|---|---|---|---|
| Disease | Train | Test | Train | Test | Train | Test | Train | Test |
| Arteriosclerosis | 10,924 | 2,808 | 6,081 | 1,658 | 5,548 | 1,514 | 4,984 | 1,350 |
| CKD | 9,958 | 2,560 | 7,168 | 1,839 | 7,175 | 1,838 | 6,527 | 1,663 |
| COPD | 8,501 | 2,202 | 5,840 | 1,510 | 5,823 | 1,502 | 5,237 | 1,320 |
| Dementia | 7,768 | 2,022 | 4,842 | 1,259 | 5,585 | 1,451 | 4,968 | 1,270 |
| Diabetes | 11,023 | 2,830 | 7,746 | 1,937 | 7,167 | 1,786 | 6,543 | 1,609 |
| Dyslipidemia | 10,157 | 2,634 | 2,560 | 632 | 2,256 | 554 | 1,968 | 471 |
| Hypertension | 11,033 | 2,831 | 4,461 | 1,247 | 4,071 | 1,133 | 3,648 | 1,018 |
| KOA | 3,905 | 996 | 642 | 160 | 1,364 | 340 | 1,373 | 342 |
| LS | 5,837 | 1,556 | 3,584 | 942 | 3,044 | 803 | 2,447 | 643 |
| Obesity | 10,959 | 2,819 | 6,202 | 1,620 | 5,742 | 1,483 | 5,210 | 1,327 |
| Osteopenia | 10,923 | 2,804 | 4,736 | 1,185 | 4,338 | 1,081 | 3,912 | 966 |

Participant assignments to training and test datasets were common to all diseases, but the number of records was different because each disease had records that were difficult to label. CKD, chronic kidney disease; COPD, chronic obstructive pulmonary disease; KOA, knee osteoarthritis; LS, locomotive syndrome.

**Supplementary Table 7. Risk factors in dominant biomarkers.**

| Disease | Biomarker | Risk factor |
|---|---|---|
| Arteriosclerosis | RbaPWV | High |
| CKD | eGFR | Low |
| COPD | FEV1%(G) | Low |
| Dementia | MMSE | Low |
| Diabetes | HbA1c | High |
| Dyslipidemia | LDL cholesterol | High |
| Hypertension | RAMAP | High |
| KOA | Ferritin | Low |
| LS | Leg score | Low |
| Obesity | BMI | High |
| Osteopenia | T-score | Low |

CKD, chronic kidney disease; COPD, chronic obstructive pulmonary disease; KOA, knee osteoarthritis; LS, locomotive syndrome.

**Supplementary Table 8. Mean/standard deviation of approached distance calculated using training data at weights decaying exponentially with distance.**

| Disease | k=1 | 2 | 4 | 6 | 8 | 10 | 12 | 16 | 20 | 24 | 28 | 32 | 40 | 48 | 56 | 64 |
|---|---|---|---|---|---|---|---|---|---|---|---|---|---|---|---|---|
| Arteriosclerosis | -0.204 | 0.468 | **0.601** | 0.588 | 0.573 | 0.551 | 0.527 | 0.480 | 0.442 | 0.405 | 0.371 | 0.340 | 0.285 | 0.237 | 0.196 | 0.161 |
| CKD | -0.171 | 0.503 | 0.632 | **0.668** | 0.668 | 0.652 | 0.639 | 0.605 | 0.573 | 0.543 | 0.517 | 0.489 | 0.438 | 0.388 | 0.348 | 0.312 |
| COPD | -0.198 | 0.670 | **0.880** | 0.857 | 0.859 | 0.842 | 0.827 | 0.787 | 0.750 | 0.717 | 0.685 | 0.658 | 0.606 | 0.559 | 0.519 | 0.486 |
| Dementia | -0.153 | 0.593 | 0.623 | 0.618 | **0.628** | 0.603 | 0.606 | 0.571 | 0.549 | 0.539 | 0.522 | 0.507 | 0.483 | 0.473 | 0.464 | 0.457 |
| Diabetes | -0.170 | 0.480 | 0.500 | **0.520** | 0.498 | 0.479 | 0.466 | 0.426 | 0.394 | 0.358 | 0.330 | 0.302 | 0.261 | 0.226 | 0.198 | 0.178 |
| Dyslipidemia | -0.102 | 0.472 | 0.723 | **0.730** | 0.708 | 0.696 | 0.648 | 0.615 | 0.570 | 0.528 | 0.493 | 0.457 | 0.396 | 0.352 | 0.312 | 0.271 |
| Hypertension | 0.044 | 0.854 | **1.002** | 0.966 | 0.986 | 0.938 | 0.930 | 0.858 | 0.818 | 0.773 | 0.745 | 0.714 | 0.659 | 0.619 | 0.586 | 0.558 |
| KOA | -0.070 | 0.635 | 0.755 | 0.768 | **0.808** | 0.787 | 0.765 | 0.736 | 0.708 | 0.666 | 0.644 | 0.616 | 0.575 | 0.528 | 0.495 | 0.471 |
| LS | -0.129 | 0.764 | 0.593 | **0.840** | 0.815 | 0.790 | 0.741 | 0.695 | 0.655 | 0.616 | 0.586 | 0.553 | 0.492 | 0.444 | 0.398 | 0.364 |
| Obesity | -0.228 | 0.316 | **0.513** | 0.493 | 0.460 | 0.422 | 0.392 | 0.323 | 0.268 | 0.226 | 0.186 | 0.152 | 0.100 | 0.058 | 0.024 | -0.006 |
| Osteopenia | -0.169 | 0.479 | **0.604** | 0.587 | 0.547 | 0.513 | 0.473 | 0.390 | 0.326 | 0.272 | 0.225 | 0.181 | 0.105 | 0.045 | -0.003 | -0.046 |

Maximum values are shown in bold. CKD, chronic kidney disease; COPD, chronic obstructive pulmonary disease; KOA, knee osteoarthritis; LS, locomotive syndrome.

# Supplementary Figures

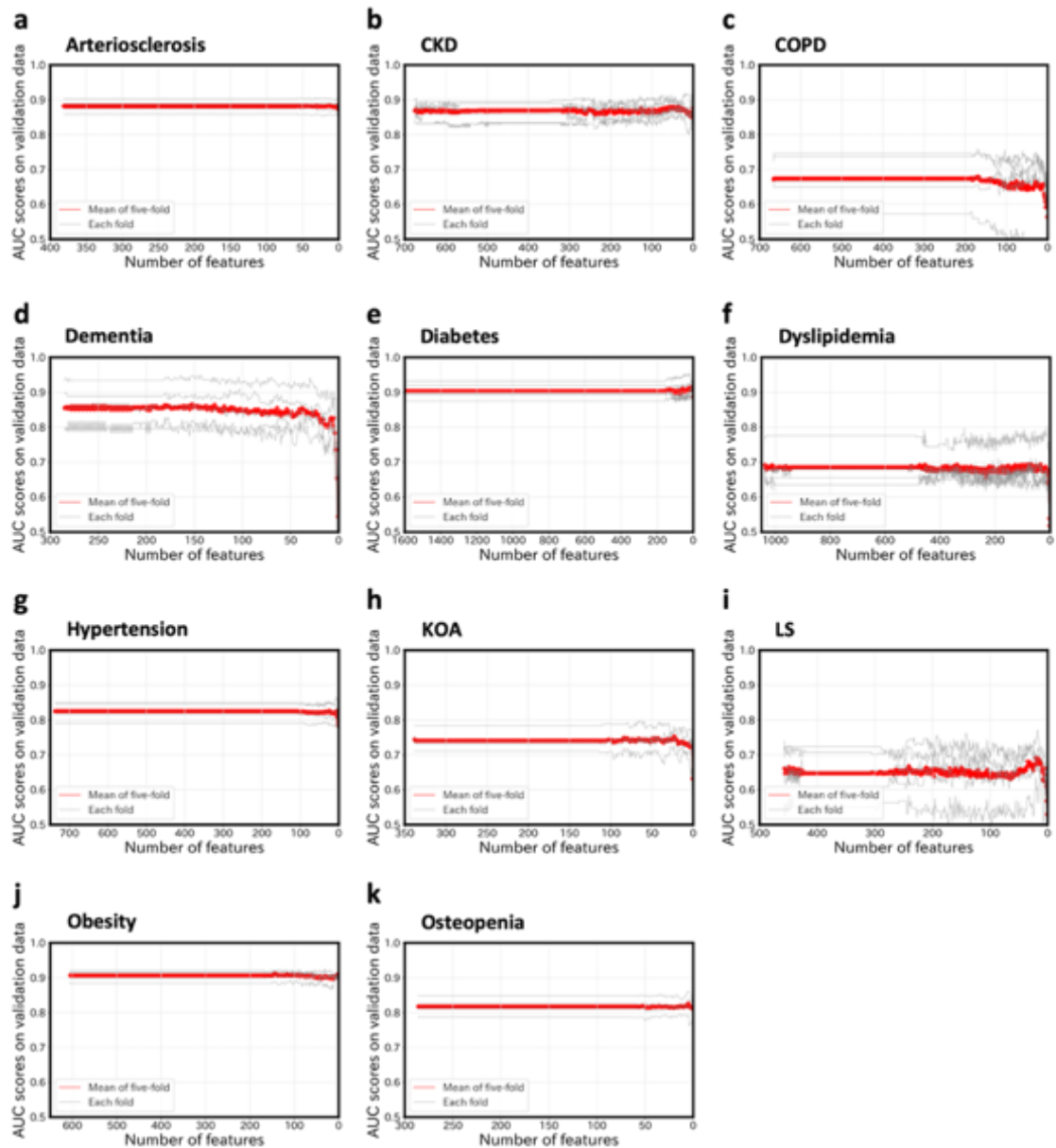

**Supplementary Figure 1. Performance of predictive models within three years during recursive feature elimination (RFE).** Features are gradually reduced in RFE, and the scores on the validation data at each stage are shown. Area under the curve (AUC) scores on arteriosclerosis (**a**), chronic kidney disease (CKD) (**b**), chronic obstructive pulmonary disease (COPD) (**c**), dementia (**d**), diabetes (**e**), dyslipidemia (**f**), hypertension (**g**), knee osteoarthritis (KOA) (**h**), locomotive syndrome (LS) (**i**), obesity (**j**), and osteopenia (**k**) are shown.

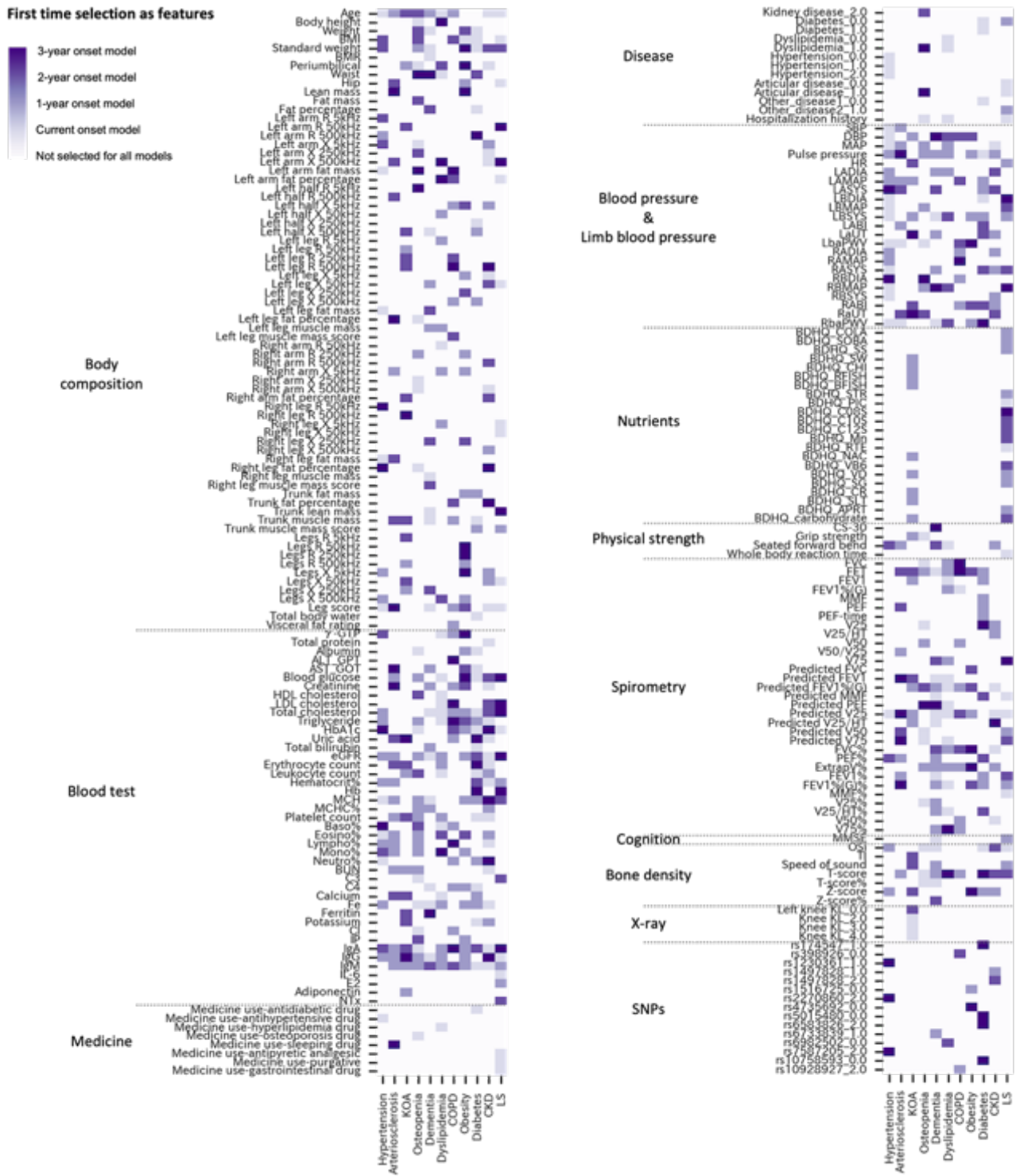

**Supplementary Figure 2. First time selection of biomarkers in recursive feature elimination (RFE) process.** Each biomarker was colored according to the first selected model. Darker colors suggest that the biomarker is an early marker.

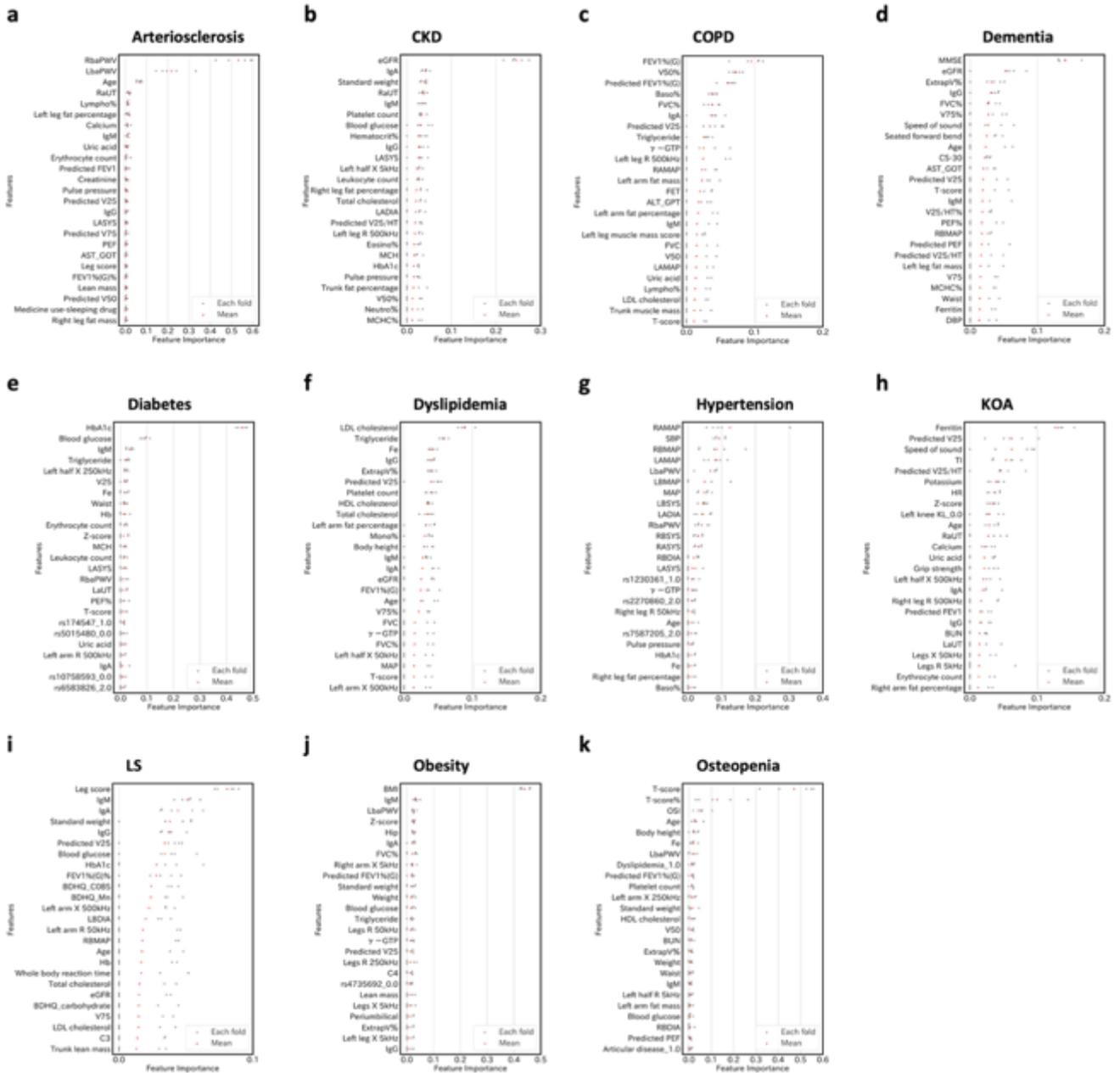

**Supplementary Figure 3. Feature importance for disease prediction models within three years.** These 25 features were selected by recursive feature elimination (RFE). RFE was performed with five-fold cross-validation, and the feature importance when 25 variables remained is shown for each fold (*n* = 5). **a** Arteriosclerosis, **b** chronic kidney disease (CKD), **c** chronic obstructive pulmonary disease (COPD), **d** dementia, **e** diabetes, **f** dyslipidemia, **g** hypertension, **h** knee osteoarthritis (KOA), **i** locomotive syndrome (LS), **j** obesity, and **k** osteopenia.

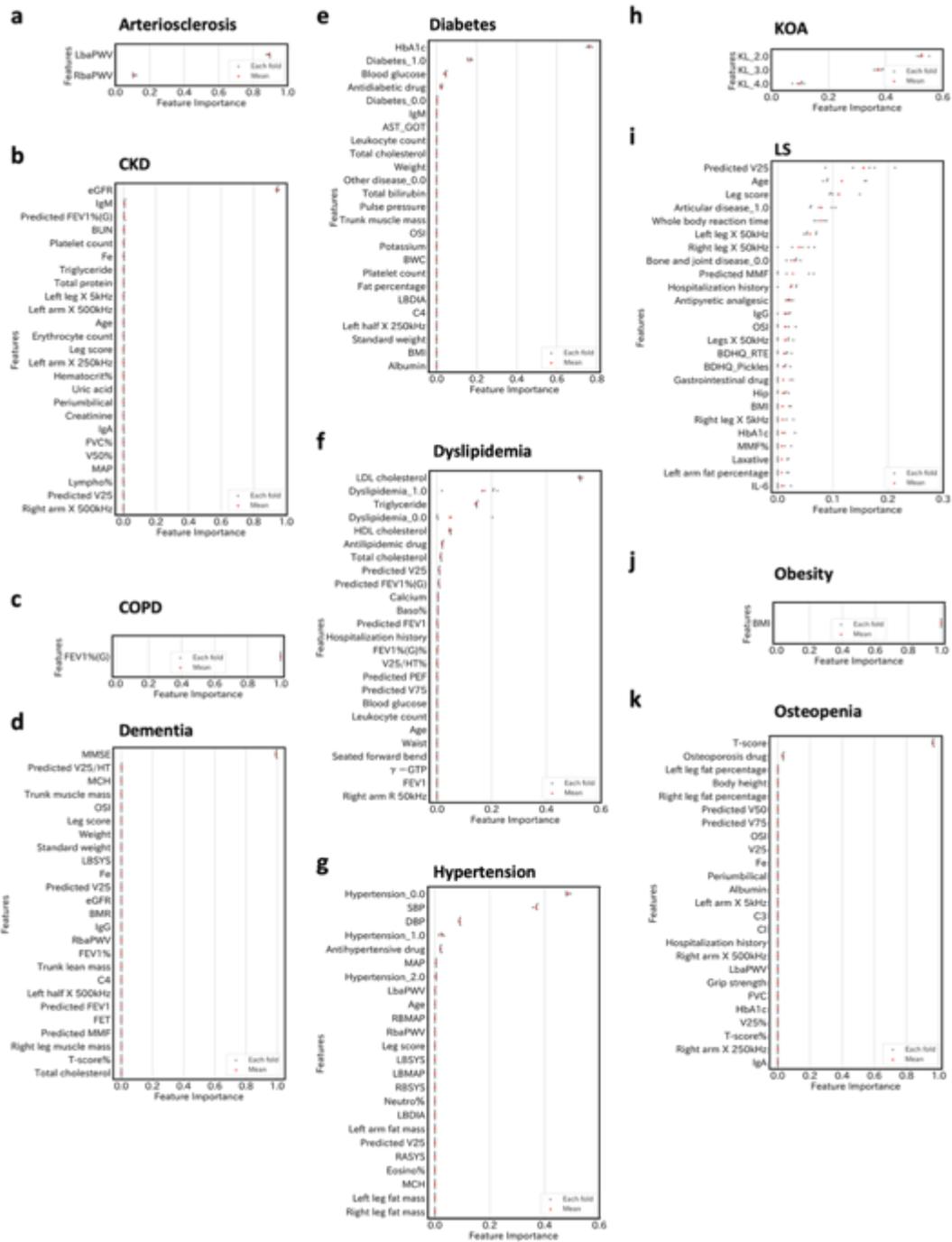

**Supplementary Figure 4. Feature importance for current disease state prediction models.** Recursive feature elimination (RFE) was performed with five-fold cross-validation, and the feature importance when 25 variables remained is shown for each fold ($n = 5$). Features with importance of zero were omitted in the plot, even if remained in RFE process. **a** Arteriosclerosis, **b** chronic kidney disease (CKD), **c** chronic obstructive pulmonary disease (COPD), **d** dementia, **e** diabetes, **f** dyslipidemia, **g** hypertension, **h** knee osteoarthritis (KOA), **i** locomotive syndrome (LS), **j** obesity, and **k** osteopenia.

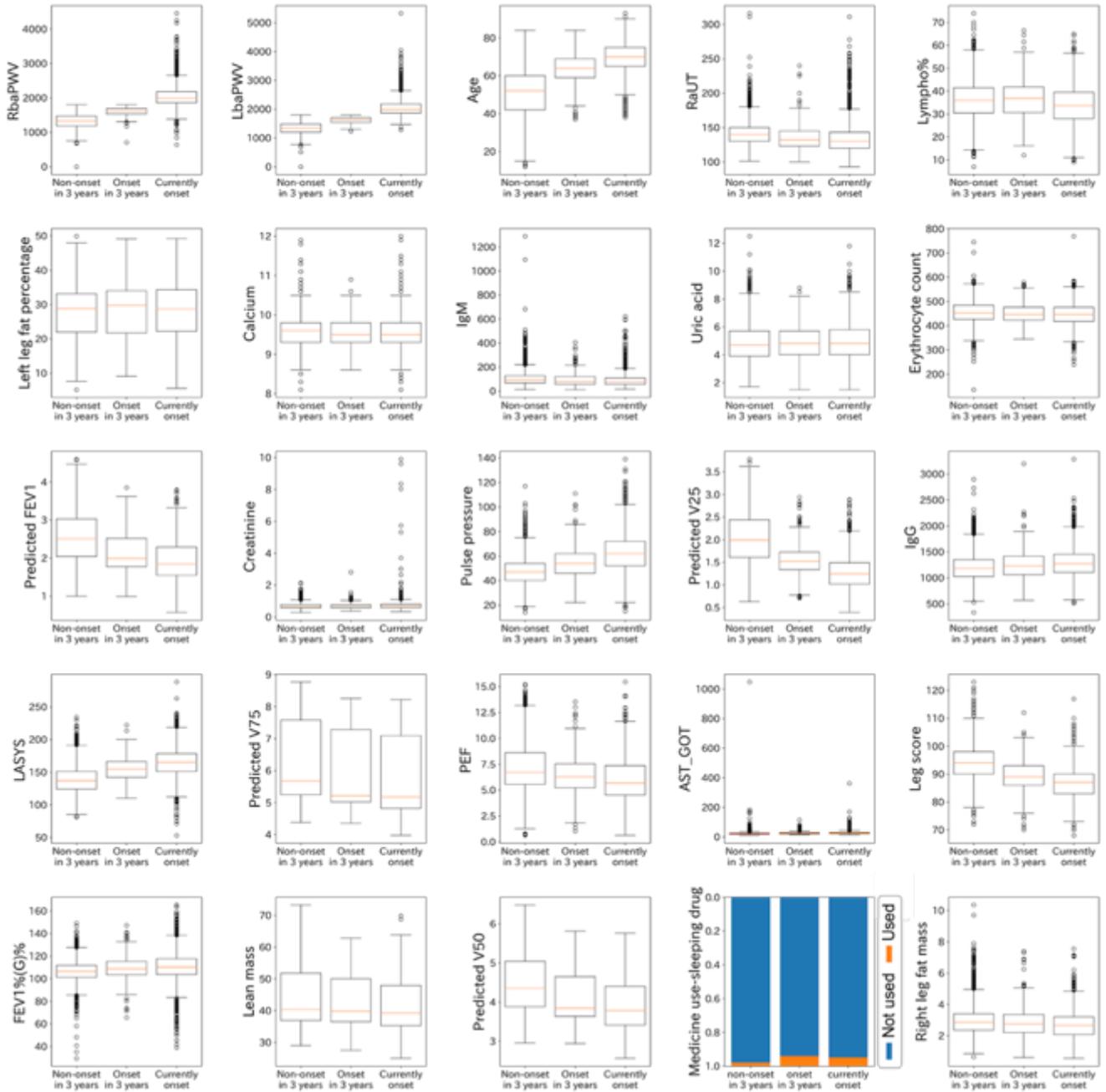

**Supplementary Figure 5 (Continued).**

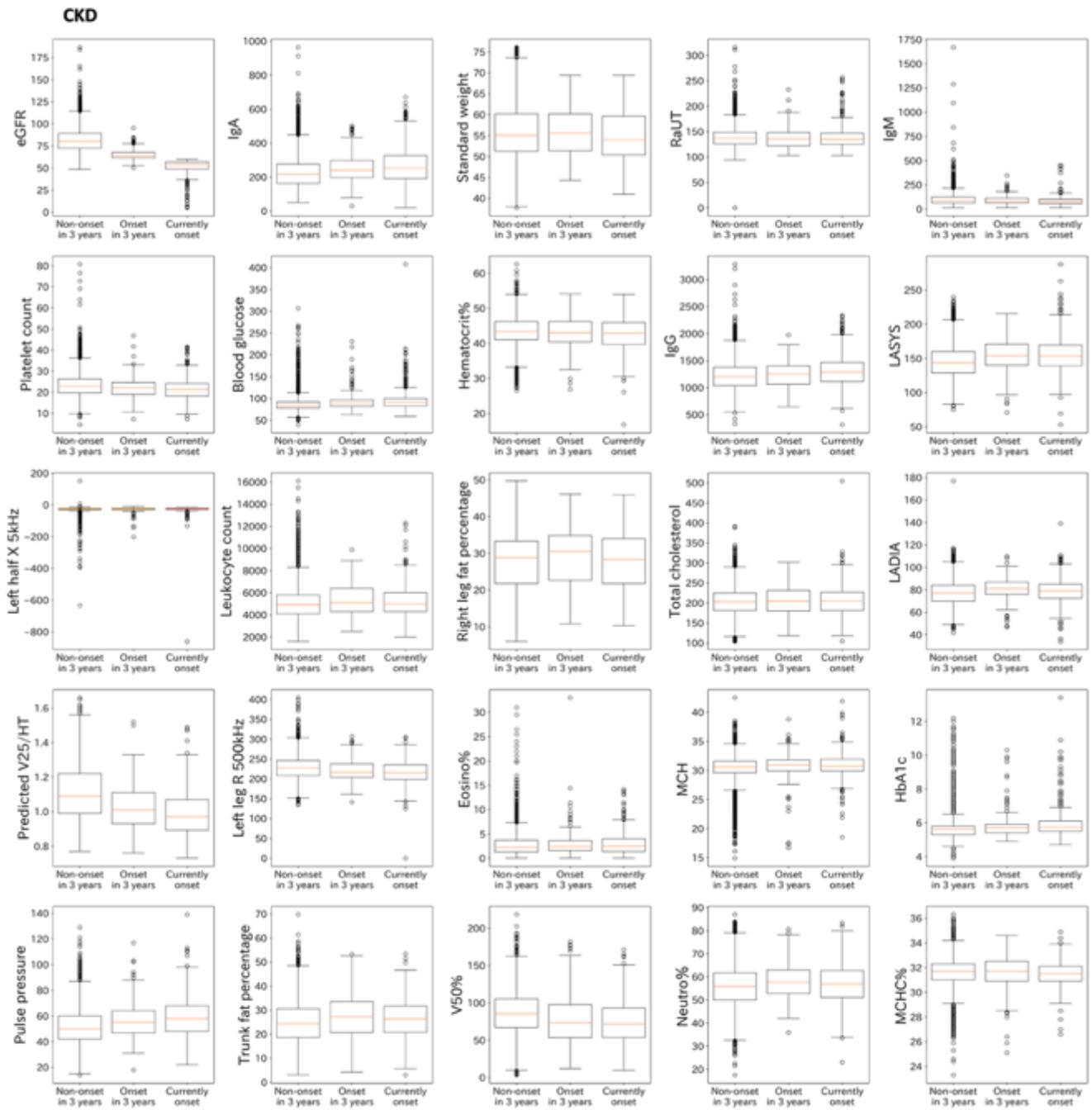

**Supplementary Figure 5 (Continued).**

**c**

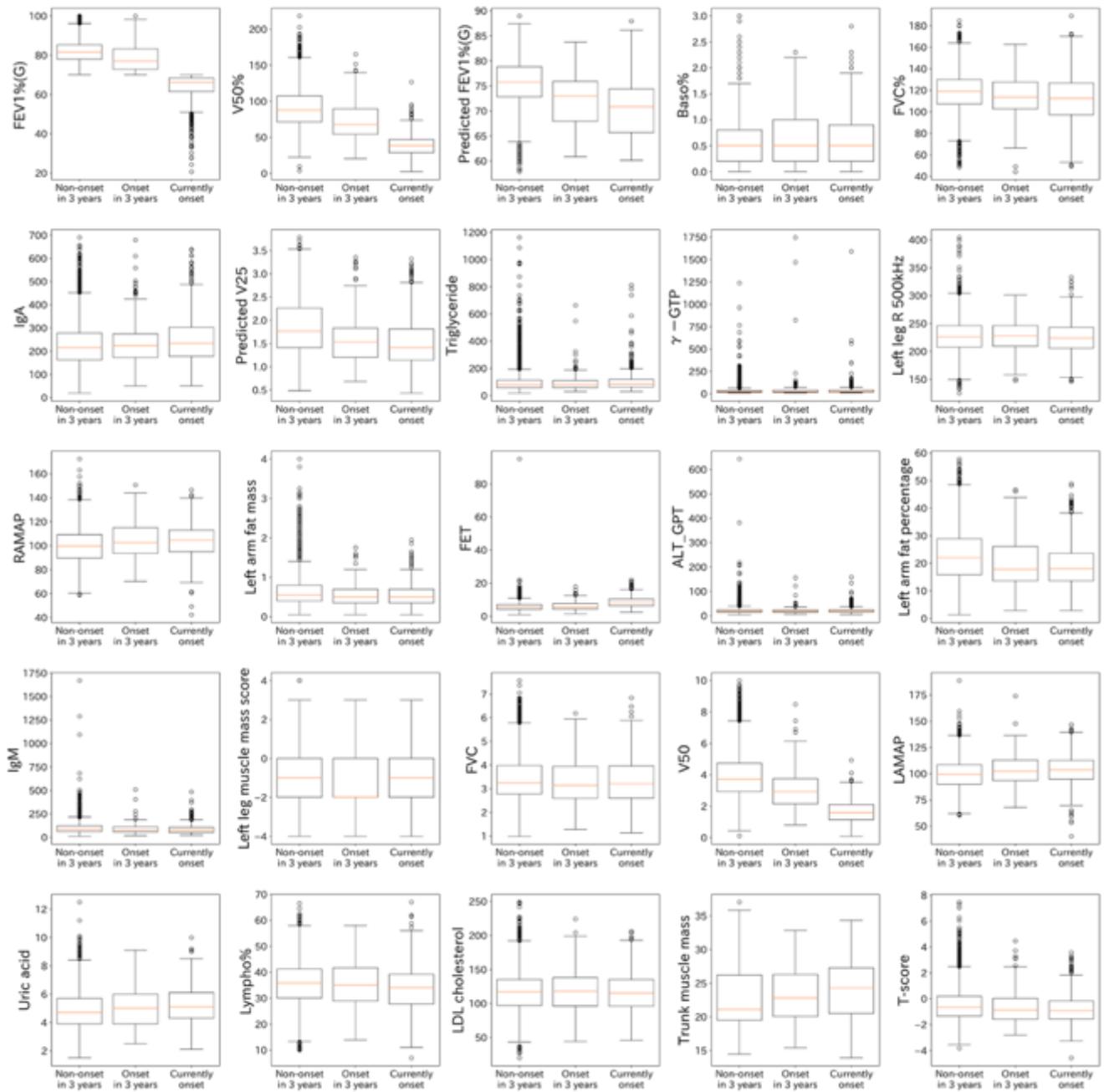

**Supplementary Figure 5 (Continued).**

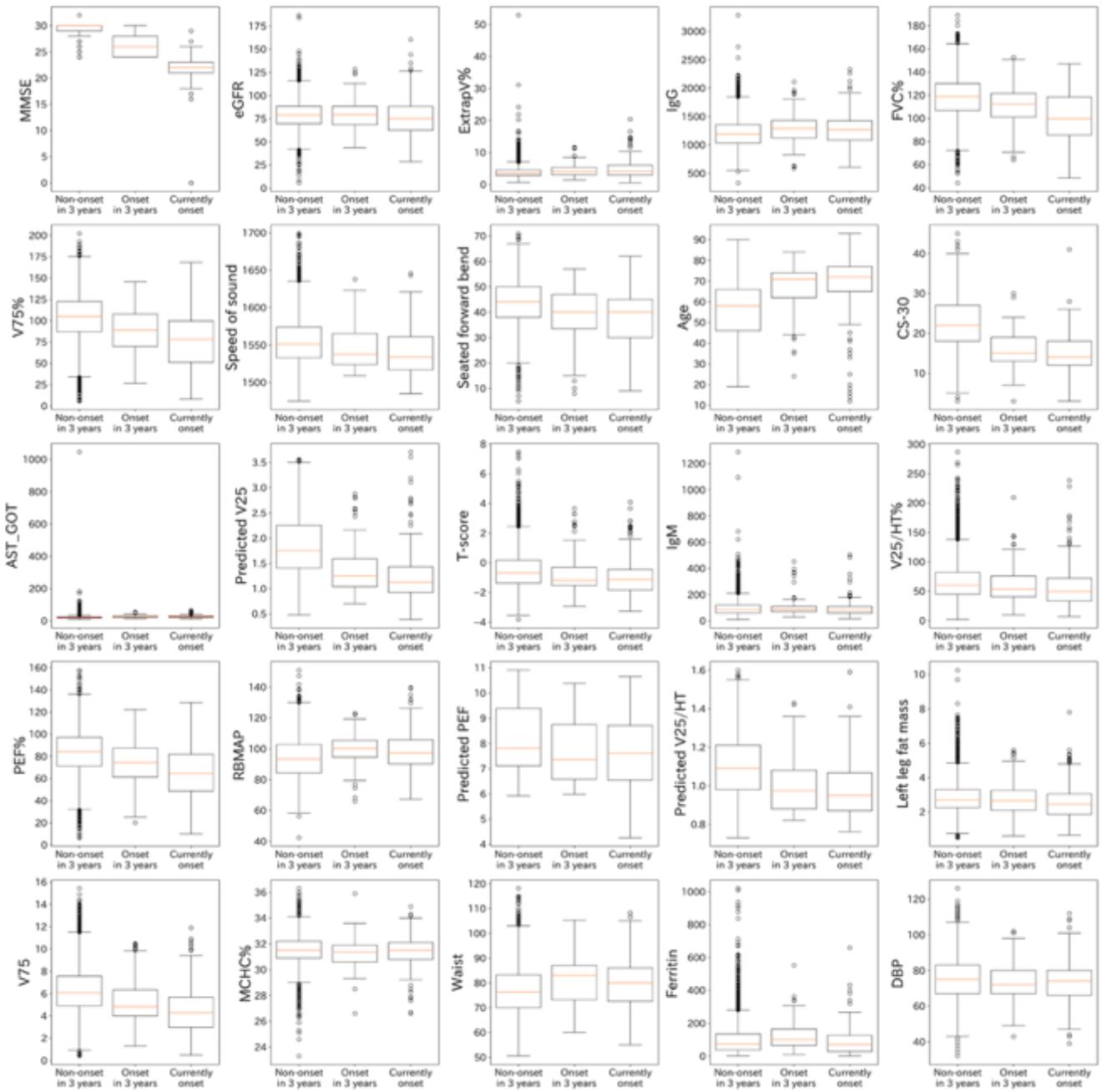

**Supplementary Figure 5 (Continued).**

e

**Diabetes**

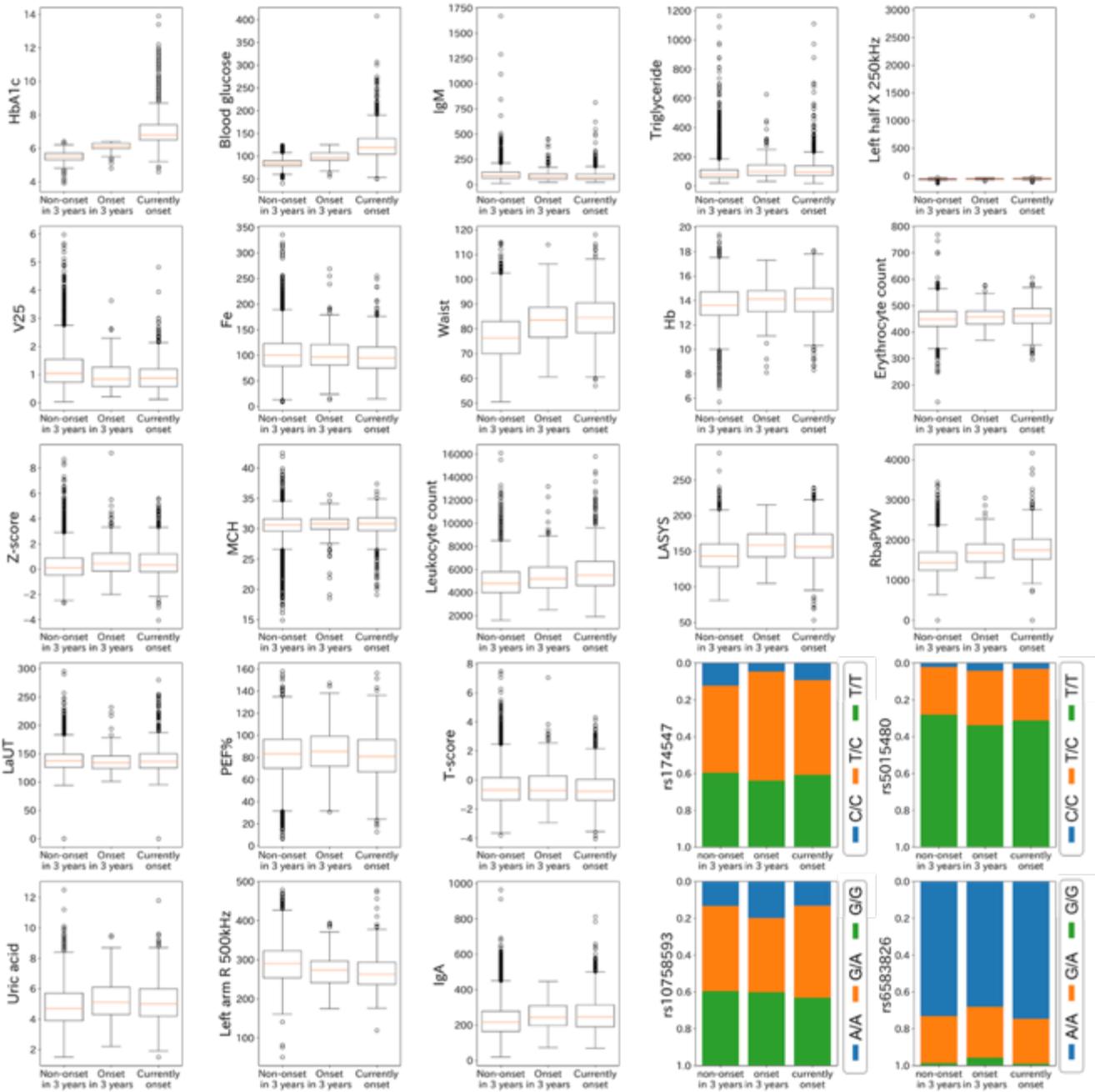

**Supplementary Figure 5 (Continued).**

## f
**Dyslipidemia**

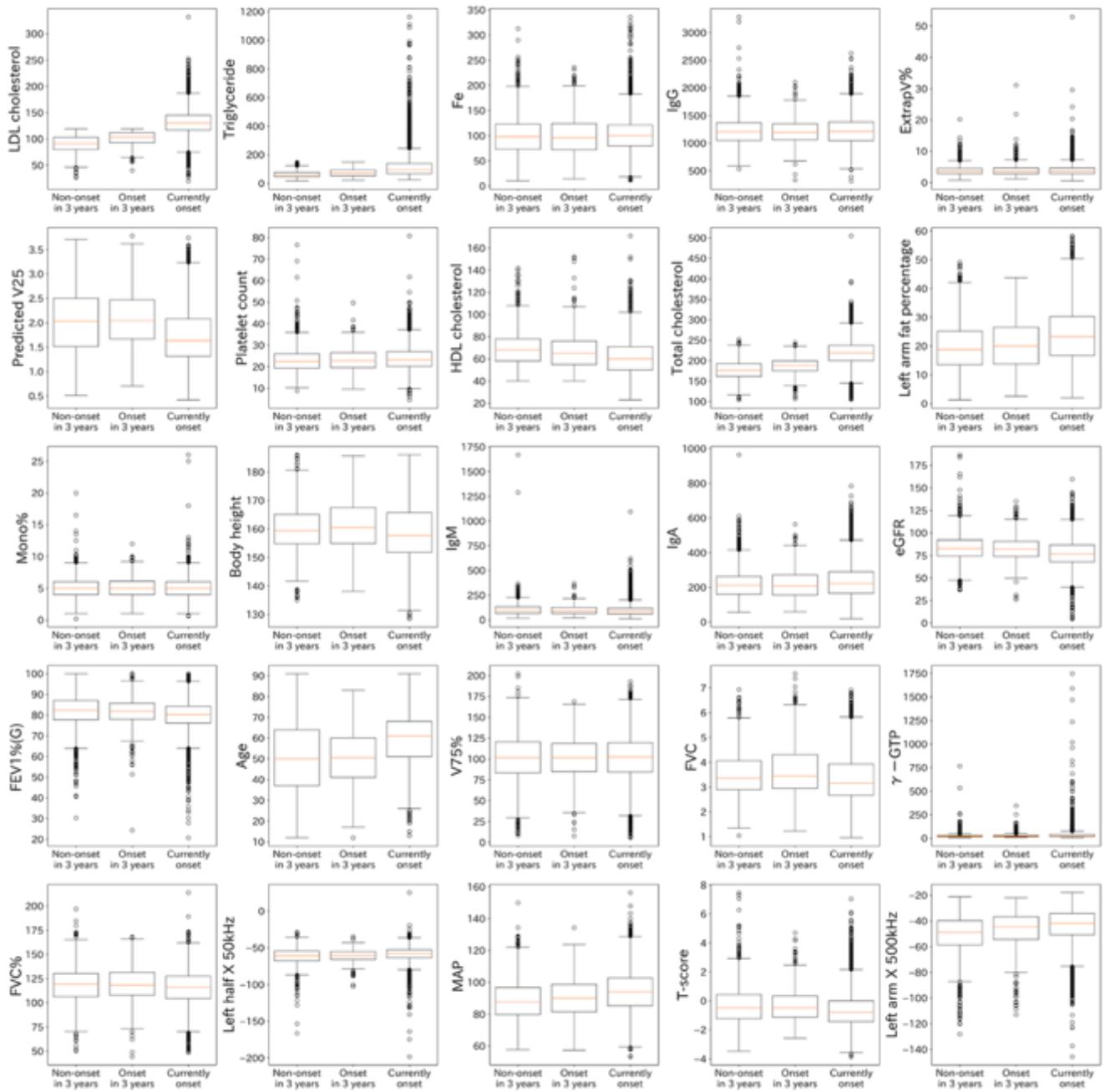

**Supplementary Figure 5 (Continued).**

g

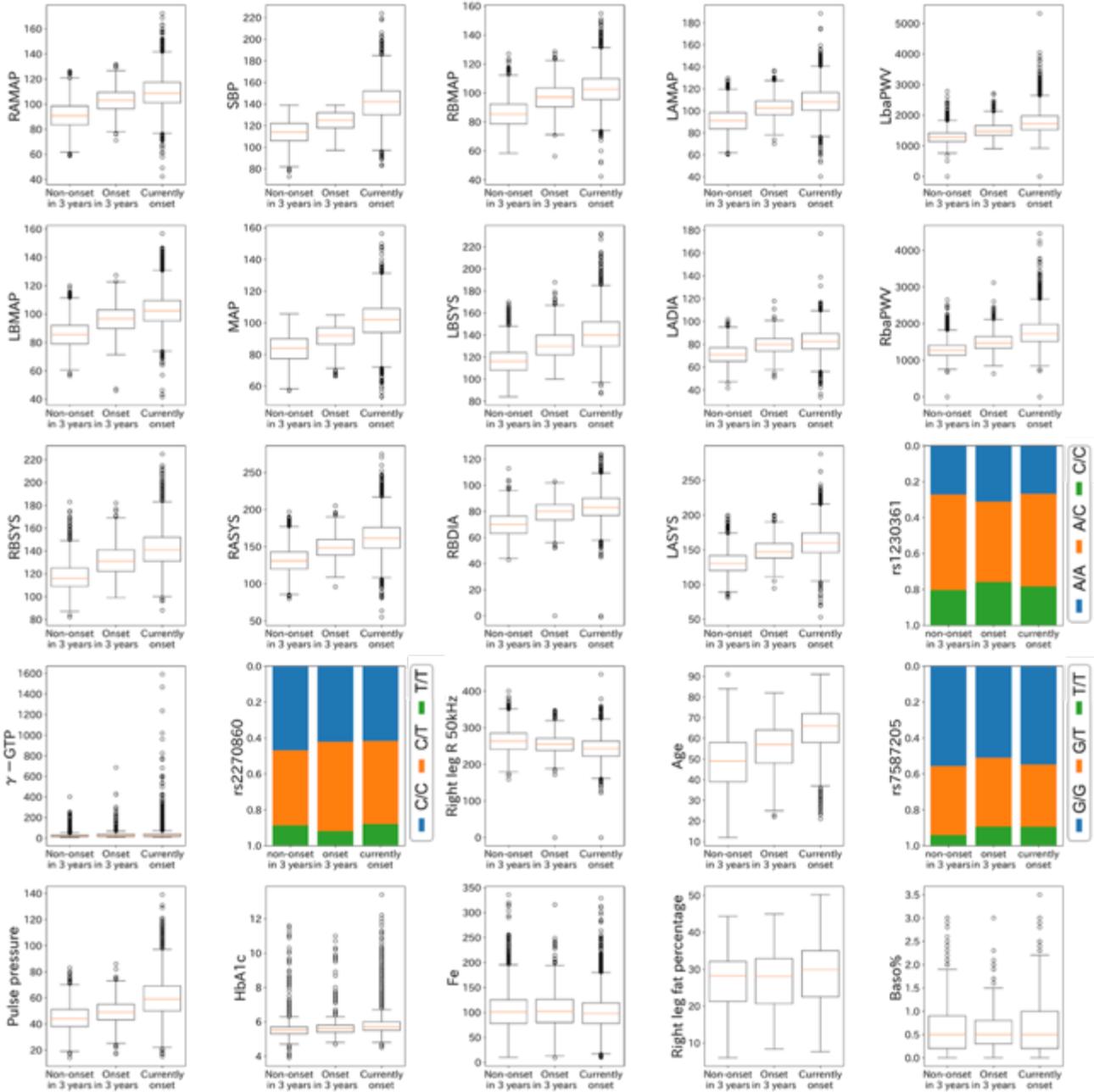

**Supplementary Figure 5 (Continued).**

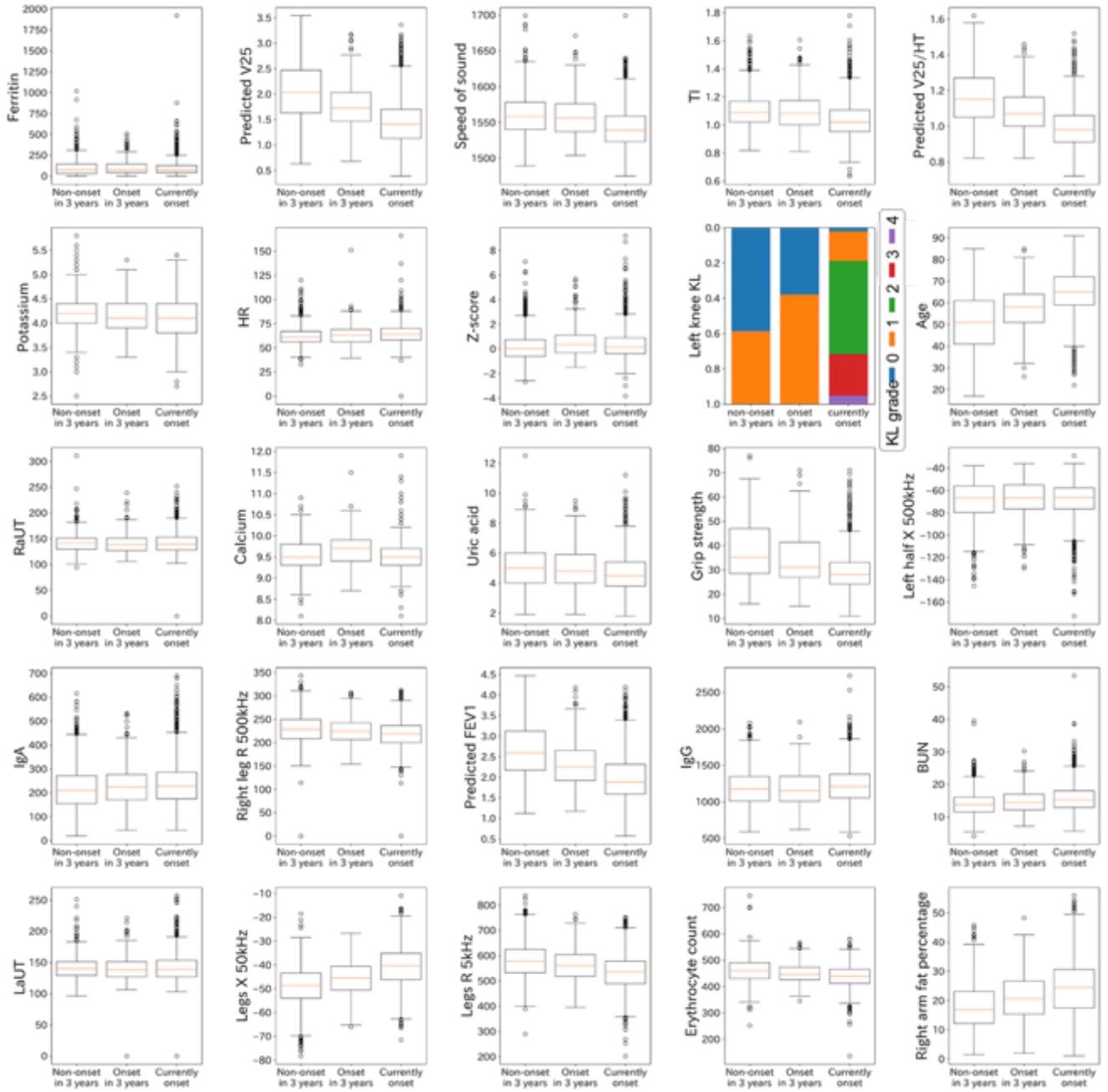

**Supplementary Figure 5 (Continued).**

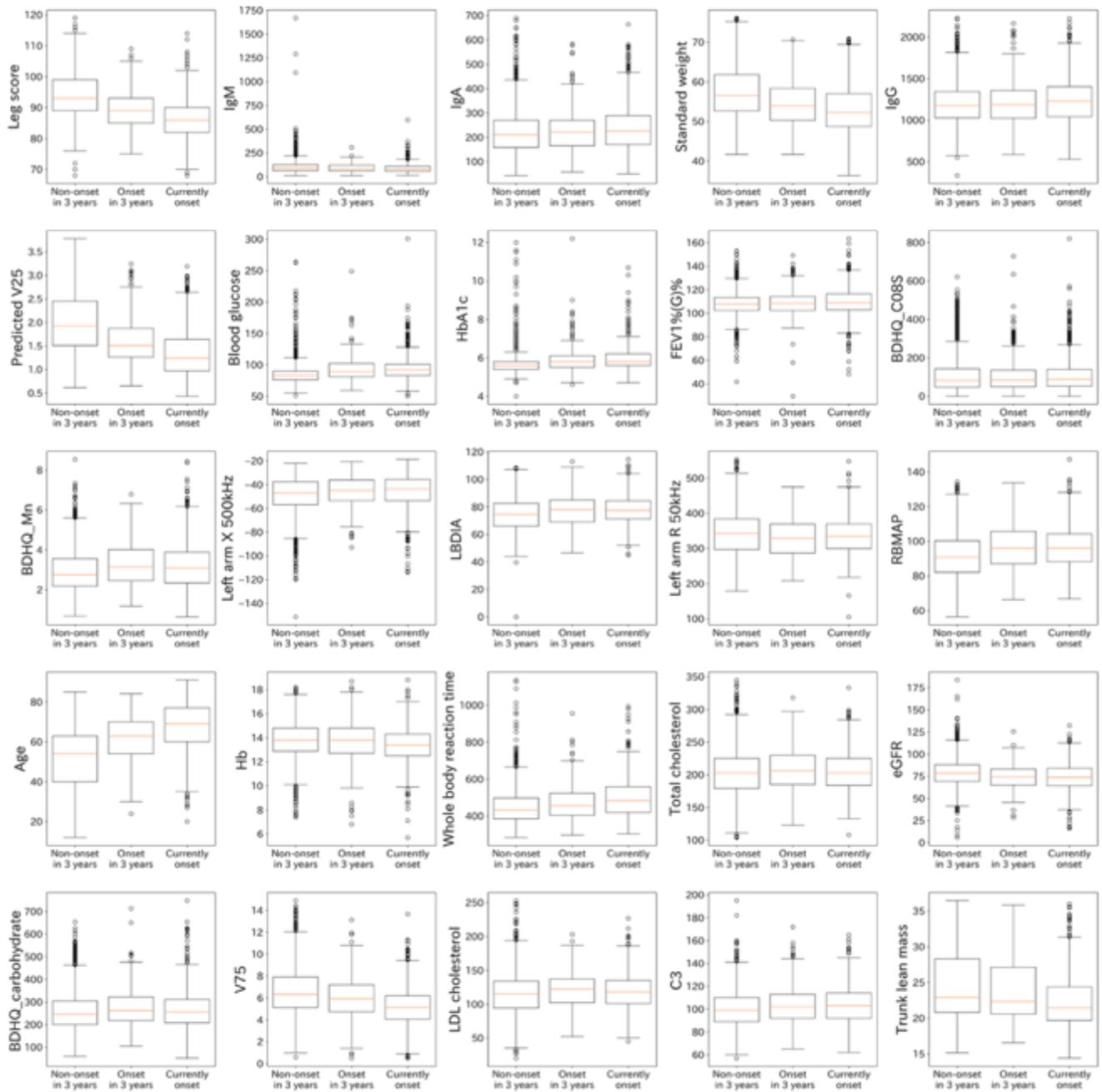

**Supplementary Figure 5 (Continued).**

j

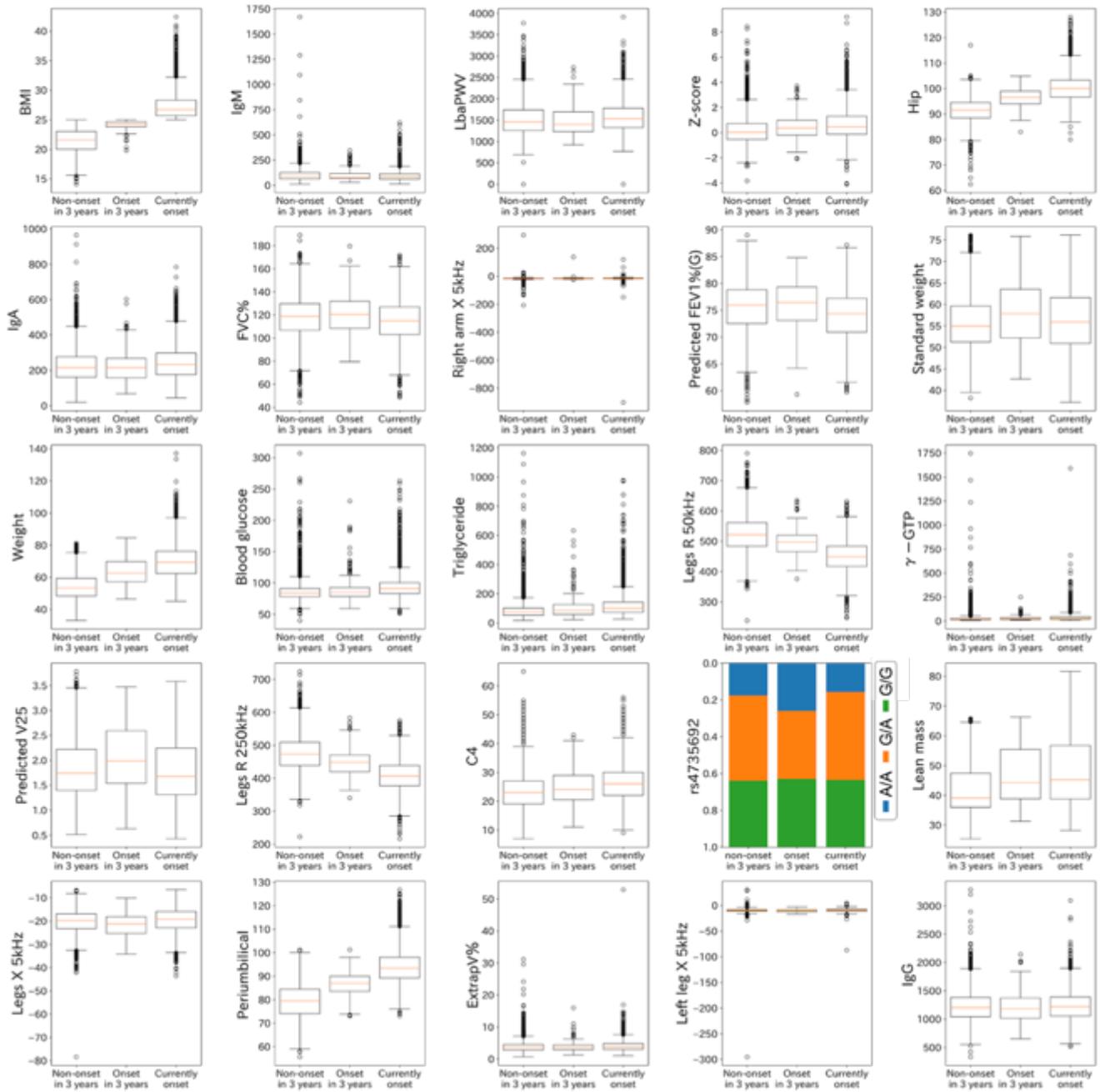

**Supplementary Figure 5 (Continued).**

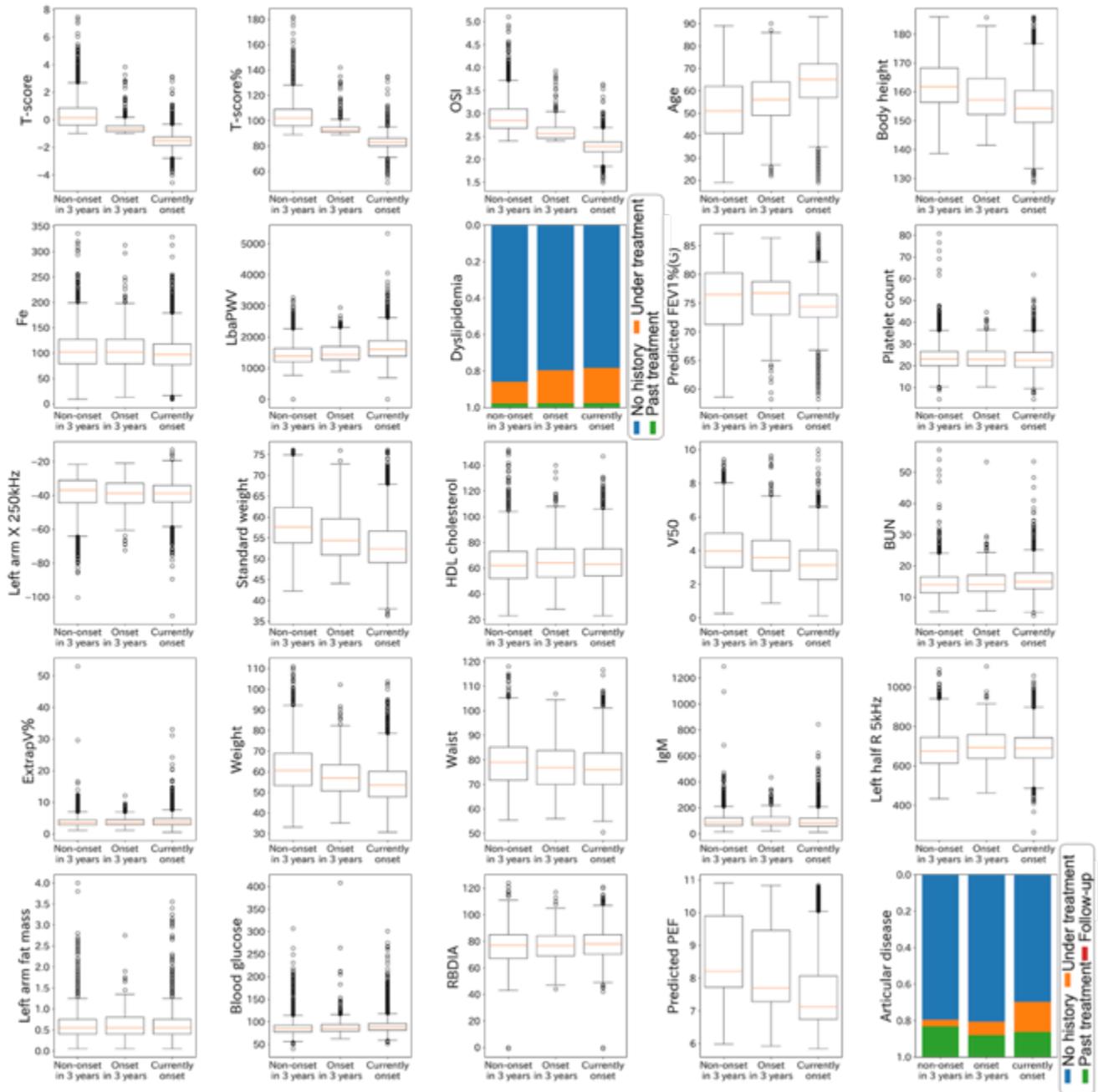

**Supplementary Figure 5. Distributions of variables among three disease status groups.** The distributions of variables selected for the disease-onset prediction models within three years were compared for three groups: non-onset within three years, onset within three years, and currently onset. In box-plots, the center line represents the median; box limits, upper and lower quartiles; whiskers, 1.5x interquartile range. **a** Arteriosclerosis, **b** chronic kidney disease (CKD), **c** chronic obstructive pulmonary disease (COPD), **d** dementia, **e** diabetes, **f** dyslipidemia, **g** hypertension, **h** knee osteoarthritis (KOA), **i** locomotive syndrome (LS), **j** obesity, and **k** osteopenia.

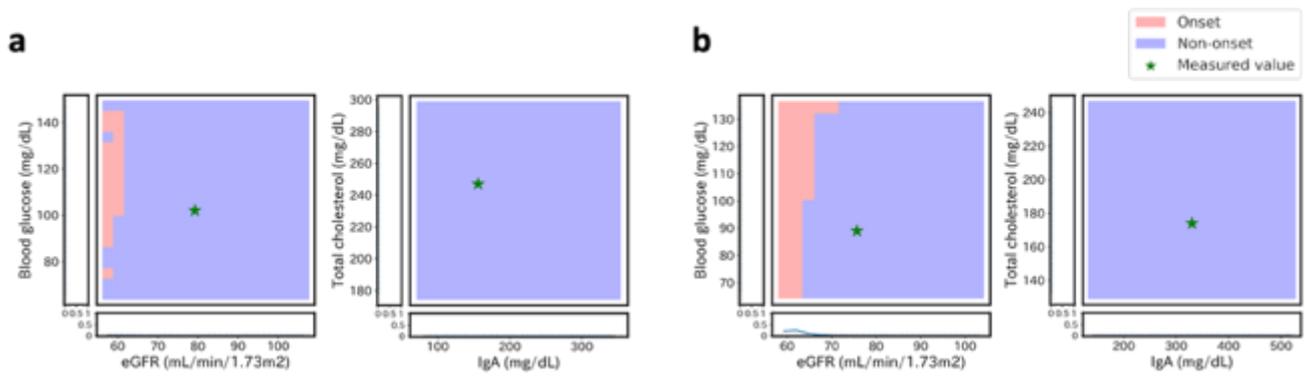

**Supplementary Figure 6. Examples of p-mICE-based HDPDs for chronic kidney disease (CKD) prediction model.** The HDPDs are shown along with the individual conditional expectation (ICE) plots. The dotted line in the ICE plot represents the onset threshold of the predicted probability in the prediction model. **a** HDPDs of randomly selected records. **b** HDPDs of the participant whose time-series analysis was performed in Fig. 3c. For each record, HDPDs of the combination of estimated glomerular filtration rate (eGFR)–blood glucose and immunoglobulin A (IgA)–total cholesterol are displayed.

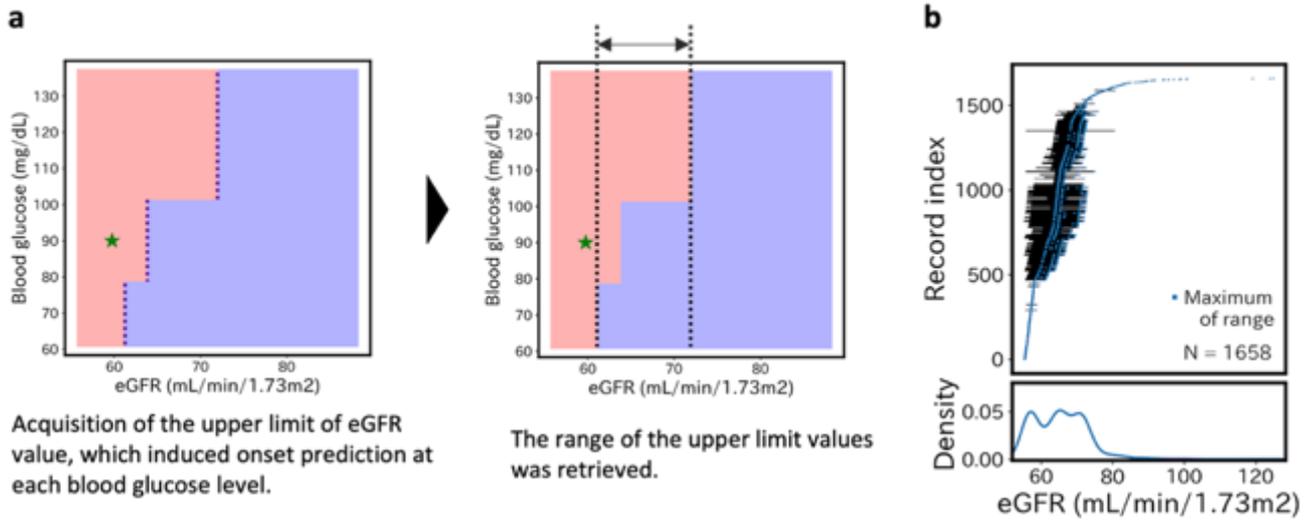

**Supplementary Figure 7. Individual differences in biomarker values forming personal boundaries in HDPDs for chronic kidney disease (CKD) prediction model.** HDPDs with estimated glomerular filtration rate (eGFR) and blood glucose as intervention variables were subjected to determine the upper limit of eGFR values inducing the CKD onset prediction. **a** Schematic representation of the acquisition of the upper limit eGFR value. First, the upper limit of eGFR value, which induced the CKD onset prediction, was obtained for each blood glucose level. Subsequently, the range of the upper limit values was retrieved. **b** Plots of the range of the upper limit eGFR values that yield CKD onset prediction in the 2d-ICE-based phase diagrams. The records were sorted by the median of the range. The maximum values of the ranges are represented in blue and summarized in the density plot.

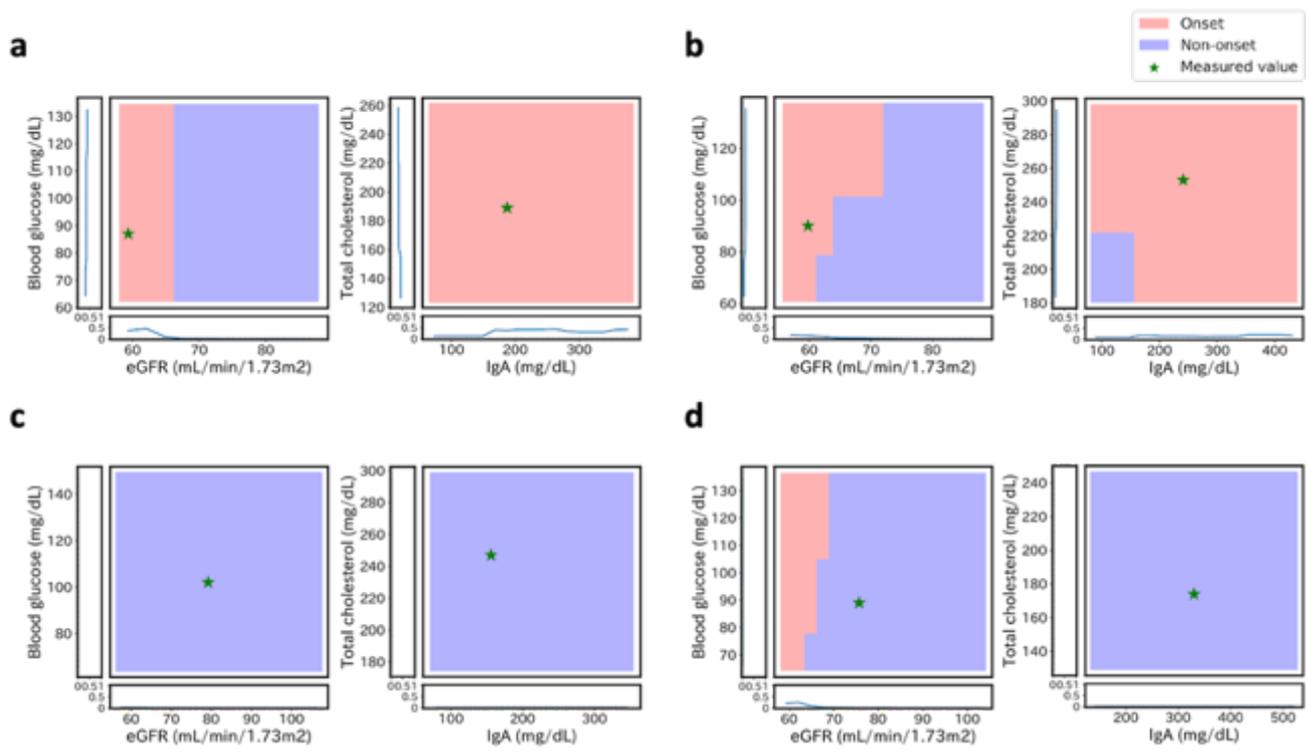

**Supplementary Figure 8. Examples of 2d-ICE-based HDPDs for chronic kidney disease (CKD) prediction model.** The phase diagrams are shown along with the individual conditional expectation (ICE) plots. The dotted line in the ICE plot represents the onset threshold of the predicted probability in the prediction model. **a–d** HDPDs of four records. The records for symbols **a** and **b** correspond to the records for the same symbols in Fig. 3, and the records for **c** and **d** correspond to the records in Supplementary Fig. 6. For each record, HDPDs of the combination of estimated glomerular filtration rate (eGFR)–blood glucose and immunoglobulin A (IgA)–total cholesterol are displayed.

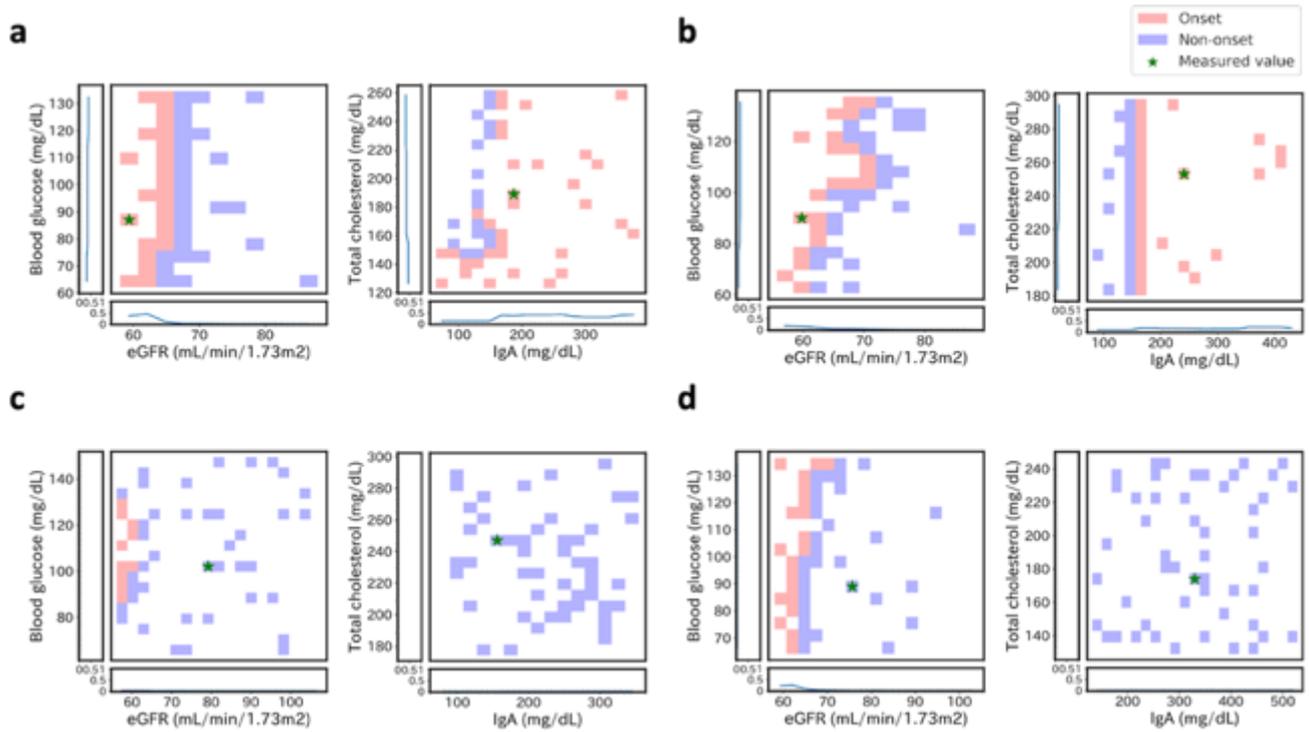

**Supplementary Figure 9. Examples of p-mICE-based HDPDs using active learning for chronic kidney disease (CKD) prediction model.** The phase diagrams using active learning are shown along with the individual conditional expectation (ICE) plots. The dotted line in the ICE plot represents the onset threshold of the predicted probability in the prediction model. **a**–**d** HDPDs of four records. The records for symbols **a** and **b** correspond to the records for the same symbols in Fig. 3, and the records for **c** and **d** correspond to the records in Supplementary Fig. 6. For each record, HDPDs of the combination of estimated glomerular filtration rate (eGFR)–blood glucose and immunoglobulin A (IgA)–total cholesterol are displayed.

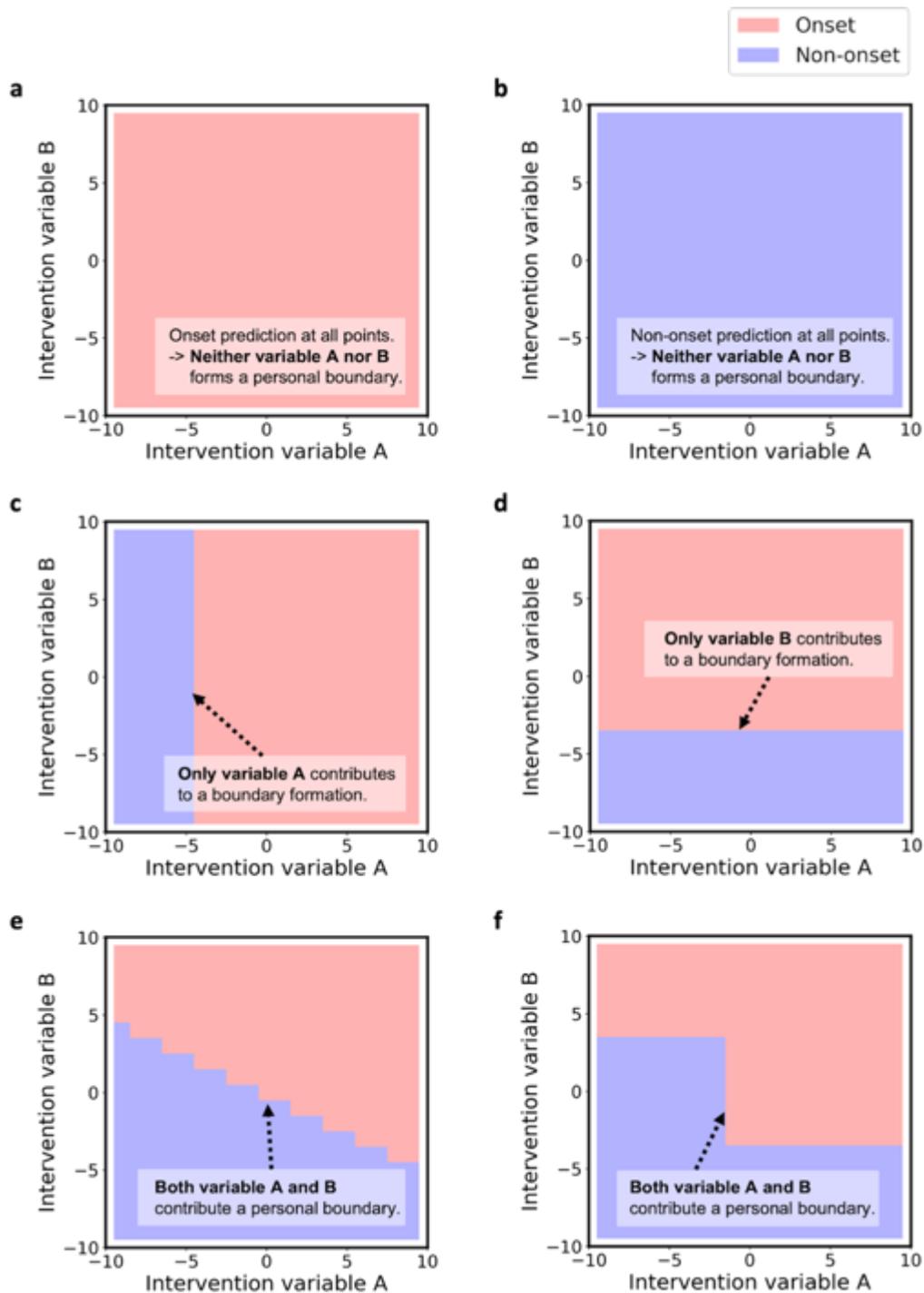

**Supplementary Figure 10. Three patterns of personal boundaries in HDPDs. a**, **b** No-boundary patterns. Entire region of the phase diagram consists only onset or non-onset points. Neither intervention variables contribute to forming the boundary. **c**, **d** Univariate boundary patterns. The boundary is defined by a single intervention variable. Only intervention variable A contributes the boundary (**c**) and only intervention variable B (**d**). **e**, **f** Bivariate boundary patterns. Both intervention variables contribute the boundary.

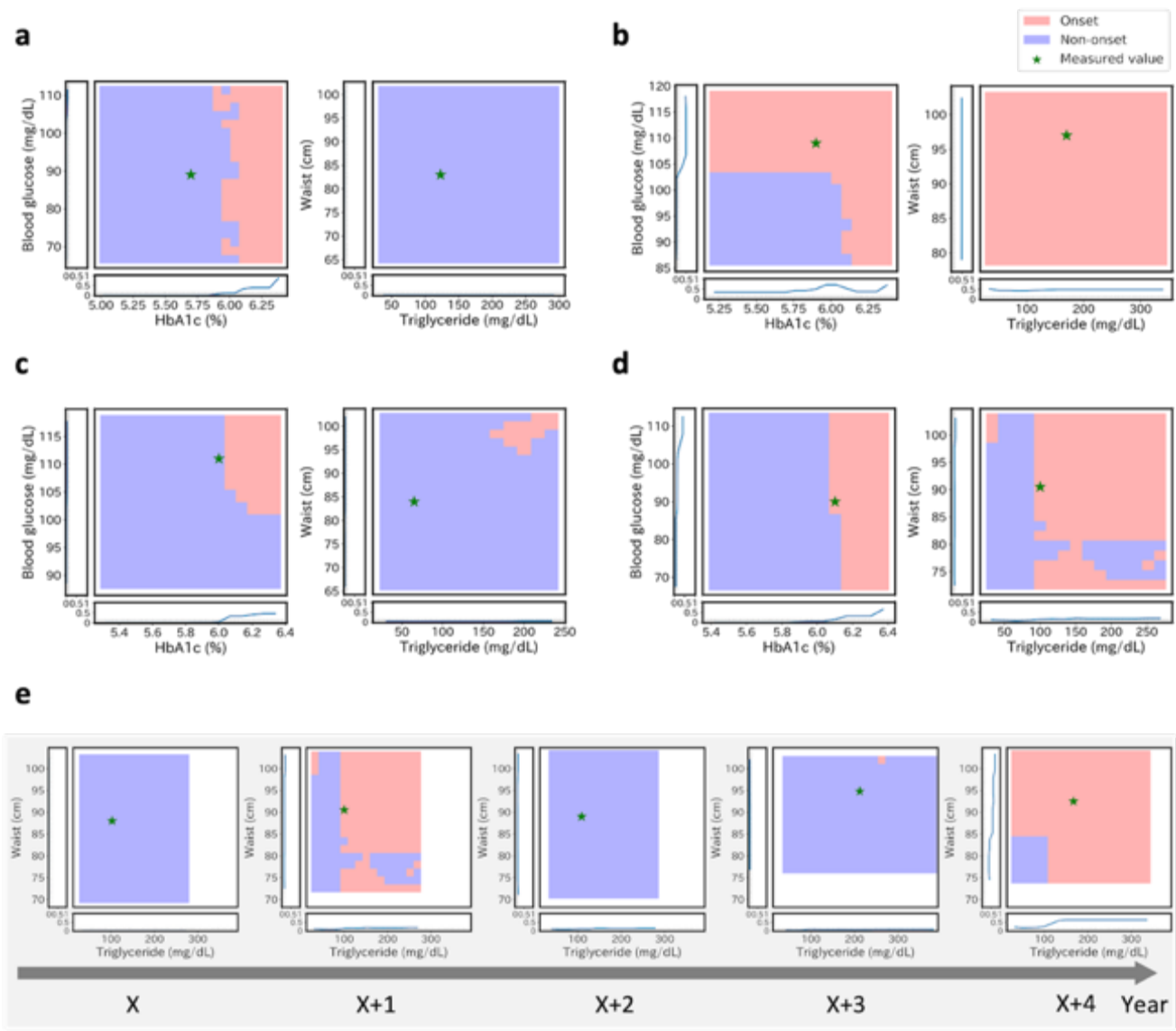

**Supplementary Figure 11. Examples of p-mICE-based HDPDs for diabetes prediction model.** The phase diagrams are shown along with the individual conditional expectation (ICE) plots. The dotted line in the ICE plot represents the onset threshold of the predicted probability in the prediction model. **a**–**d** HDPDs of four randomly selected records. For each record, HDPDs of the combination of HbA1c–blood glucose and triglyceride–waist are displayed. **e** Time-series representation of HDPDs in the same participant. The participant of the record (**d**) is selected for the plot. Personal HDPDs of the combination of triglyceride–waist are displayed. The year values were masked for privacy reasons.

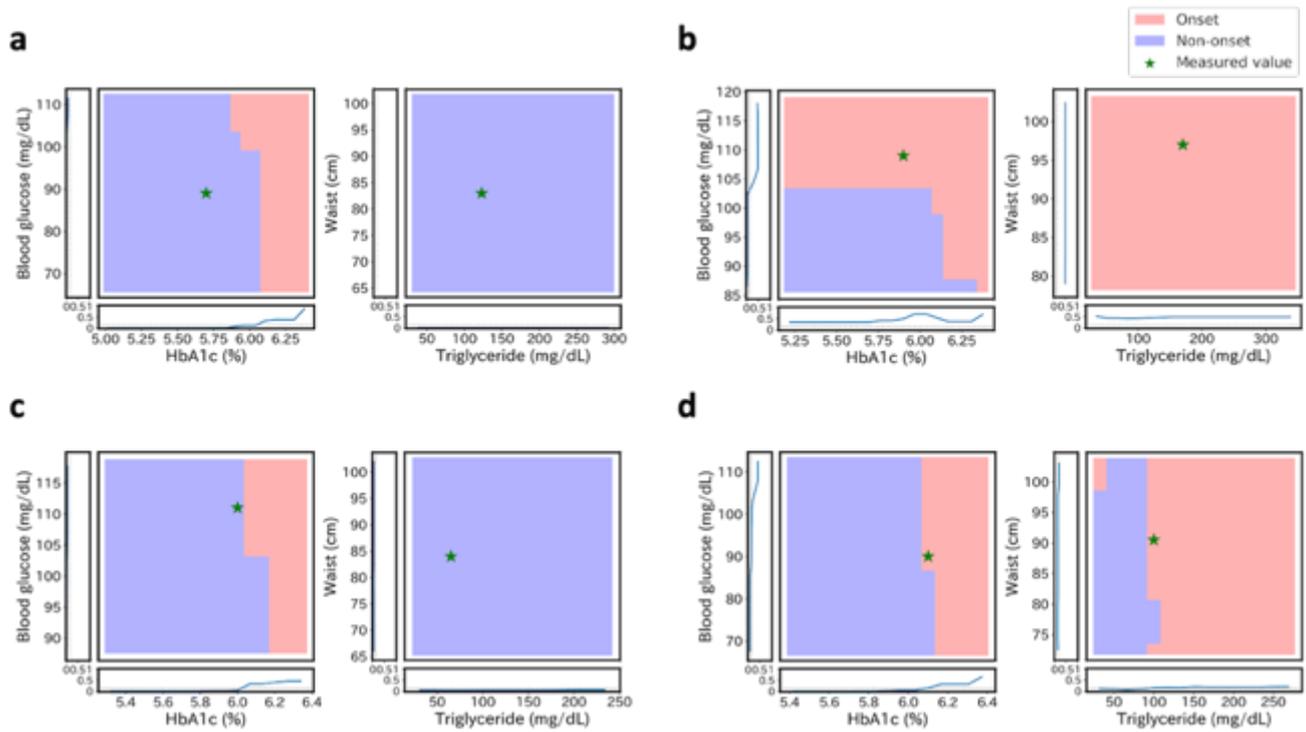

**Supplementary Figure 12. Examples of 2d-ICE-based HDPDs for diabetes prediction model.** The phase diagrams are shown along with the individual conditional expectation (ICE) plots. The dotted line in the ICE plot represents the onset threshold of the predicted probability in the prediction model. **a**–**d** HDPDs of four records. The records are identical to Supplementary Fig. 11, and the symbols correspond. For each record, HDPDs of the combination of HbA1c–blood glucose and triglyceride–waist are displayed.

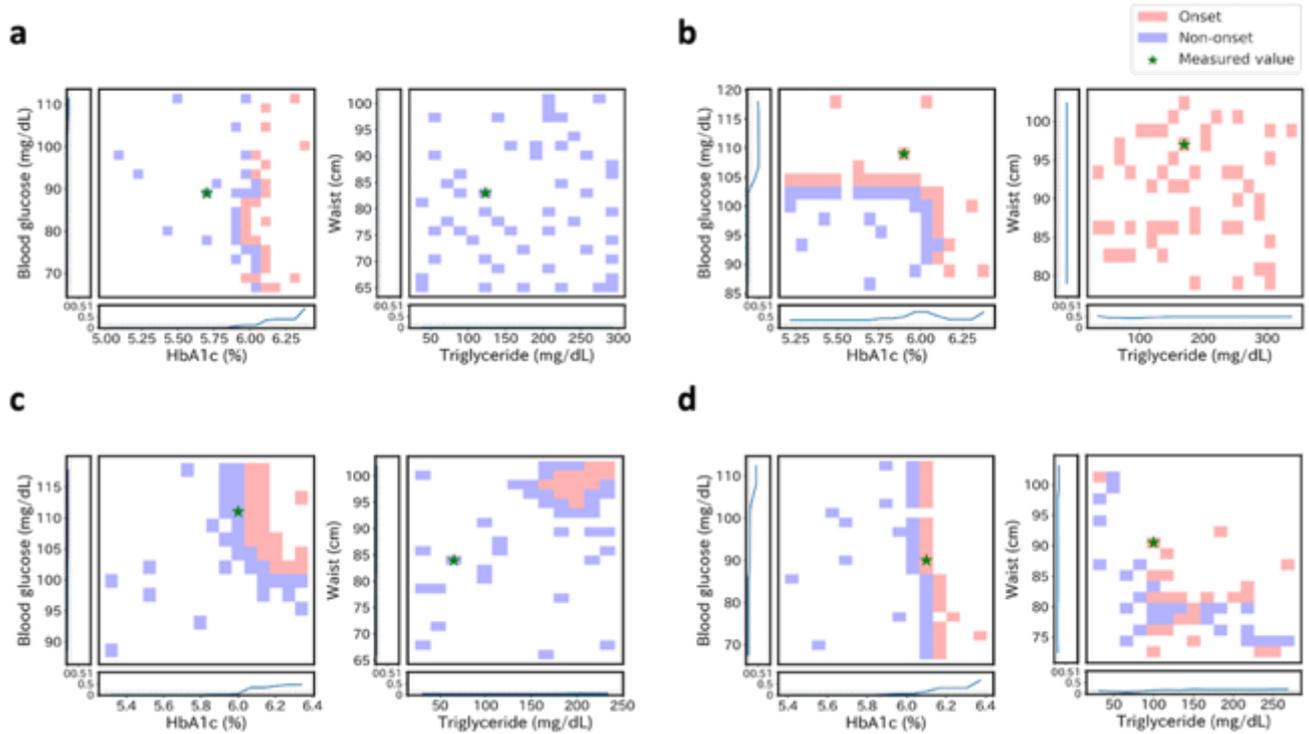

**Supplementary Figure 13. Examples of p-mICE-based HDPDs using active learning for diabetes prediction model.** The HDPDs using active learning are shown along with the individual conditional expectation (ICE) plots. The dotted line in the ICE plot represents the onset threshold of the predicted probability in the prediction model. **a**–**d** HDPDs of four records. The records are identical to Supplementary Fig. 11, and the symbols correspond. For each record, HDPDs of the combination of HbA1c–blood glucose and triglyceride–waist are displayed.

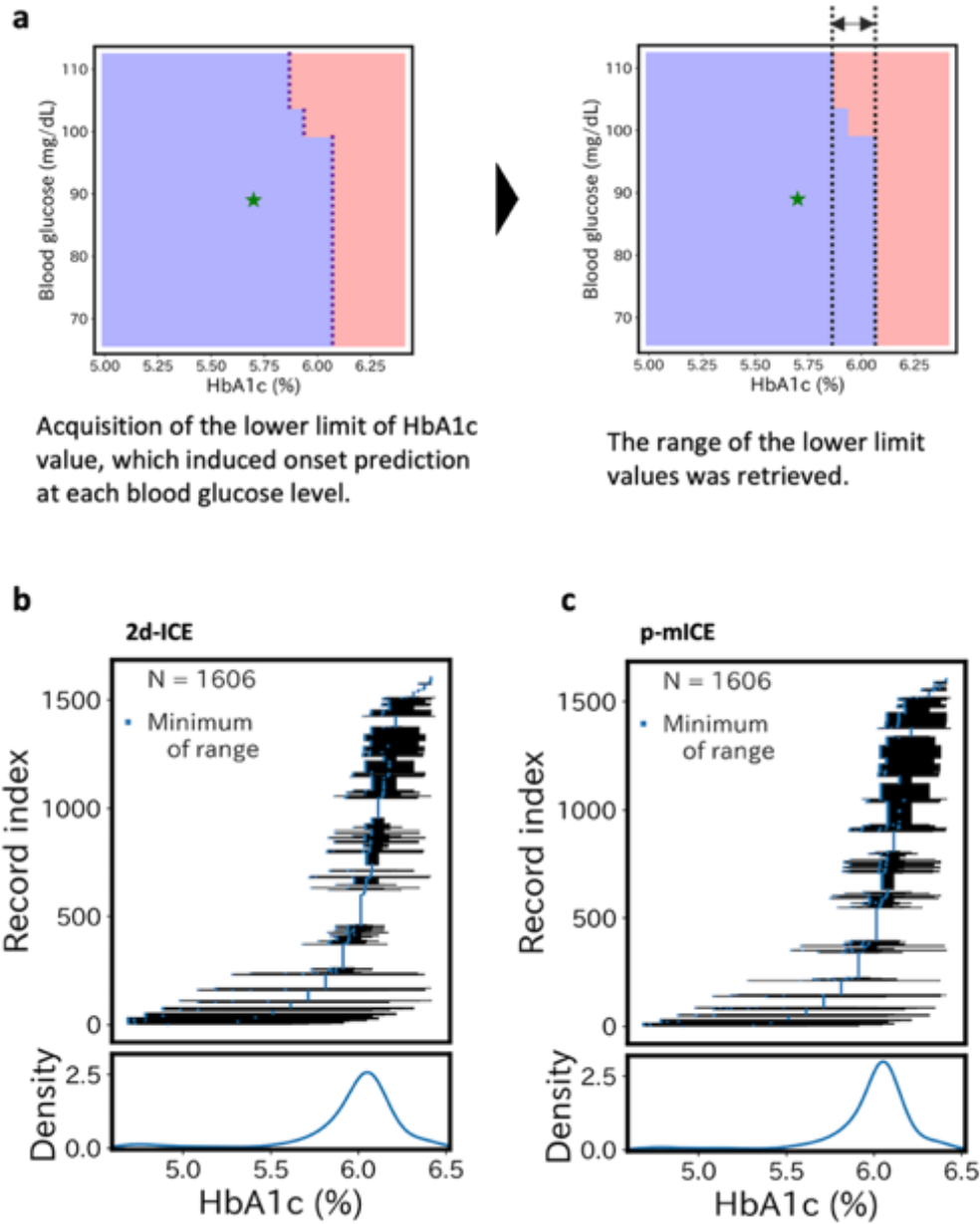

**Supplementary Figure 14. Individual differences in biomarker values forming personal boundaries in phase diagrams for diabetes prediction model.** HDPDs with HbA1c and blood glucose as intervention variables were subjected to determine the lower limit of HbA1c values inducing the diabetes onset prediction. **a** Schematic representation of the acquisition of the lower limit HbA1c value. First, the lower limit of HbA1c value, which induced the diabetes onset prediction, was obtained for each blood glucose level. Subsequently, the range of the lower limit values was retrieved. **b, c** Plots of the range of the lower limit HbA1c values that yield diabetes onset prediction in the 2d-ICE-based (**b**) and p-mICE-based phase diagrams (**c**). The records were sorted by the median of the range. The minimum values of the ranges are represented in blue and summarized in the density plot.

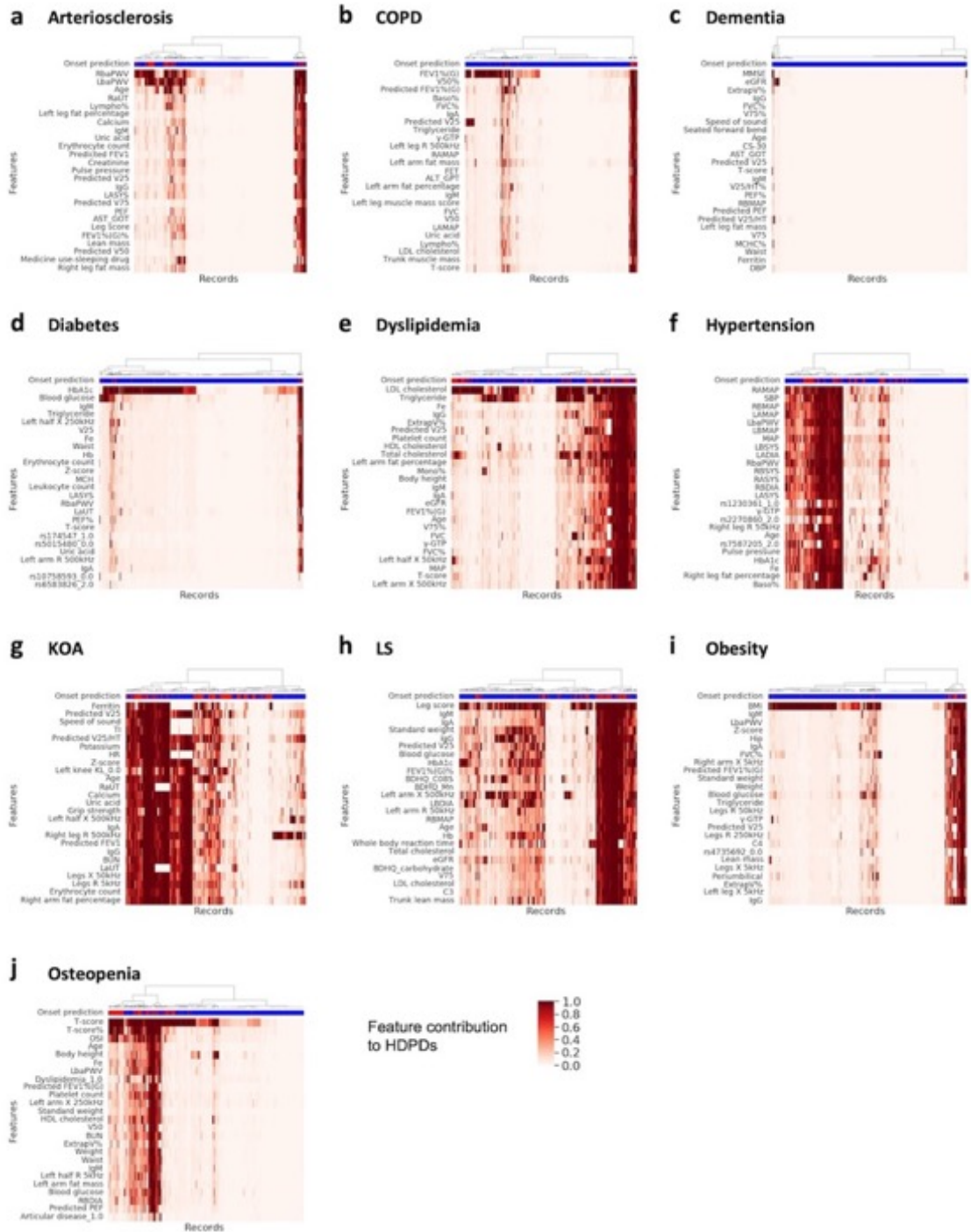

**Supplementary Figure 15. Hierarchically-clustered heatmap for feature contribution to p-mICE-based HDPDs among records.** The feature contribution to HDPDs was calculated as the proportion of the HDPDs where the variable contributed to the personal boundary (detailed in Supplementary Fig. 10). In the heatmaps, red predicted labels represent records with onset prediction and blue with non-onset. **a** Arteriosclerosis, **b** chronic obstructive pulmonary disease (COPD), **c** dementia, **d** diabetes, **e** dyslipidemia, **f** hypertension, **g** knee osteoarthritis (KOA), **h** locomotive syndrome (LS), **i** obesity, and **j** osteopenia.

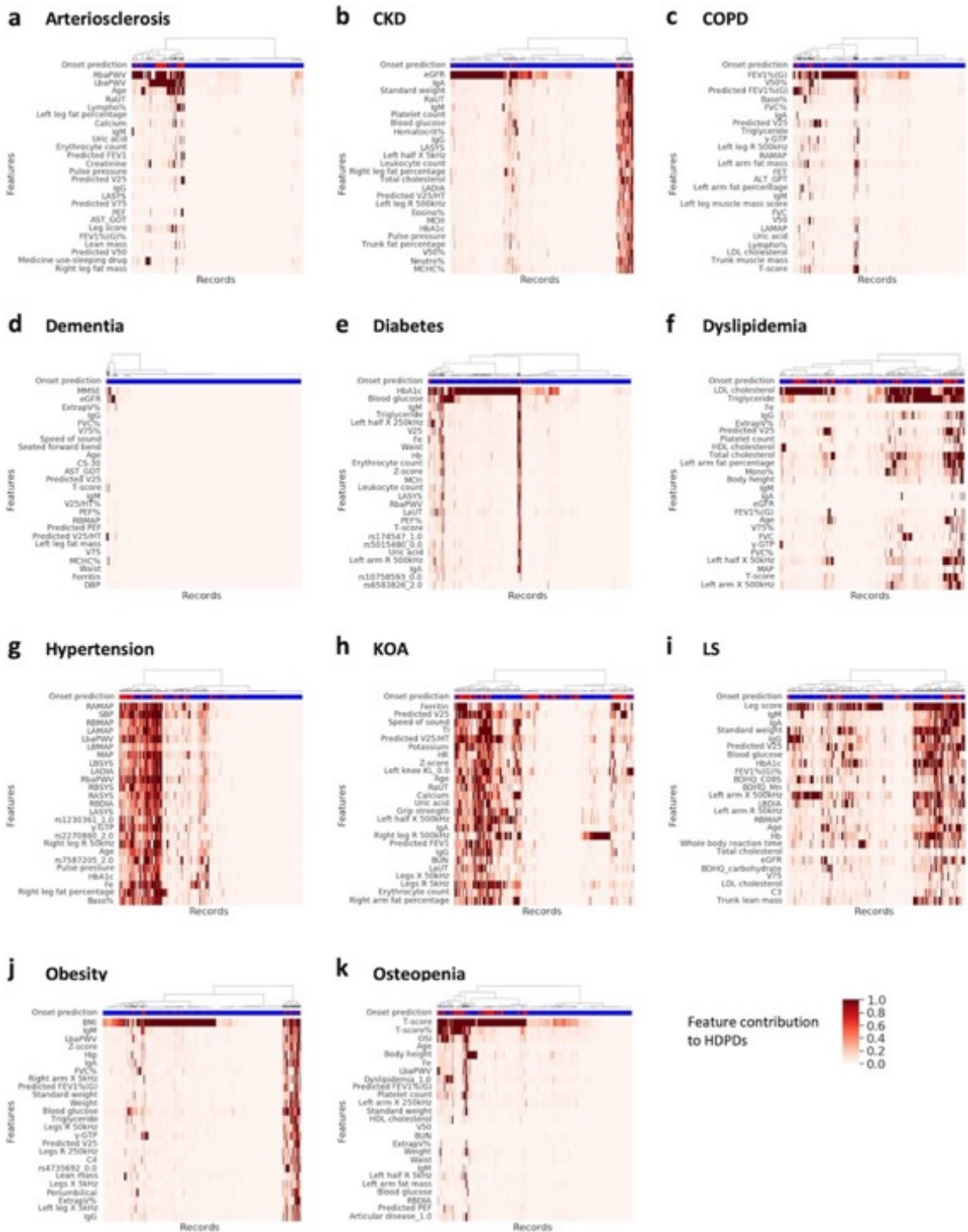

**Supplementary Figure 16. Hierarchically-clustered heatmap for feature contribution to 2d-ICE-based HDPDs among records.** The feature contribution to HDPDs was calculated as the proportion of the HDPDs where the variable contributed to the personal boundary (detailed in Supplementary Fig. 10). In the heatmaps, red predicted labels represent records with onset prediction and blue with non-onset. **a** Arteriosclerosis, **b** chronic kidney disease (CKD), **c** chronic obstructive pulmonary disease (COPD), **d** dementia, **e** diabetes, **f** dyslipidemia, **g** hypertension, **h** knee osteoarthritis (KOA), **i** locomotive syndrome (LS), **j** obesity, and **k** osteopenia.

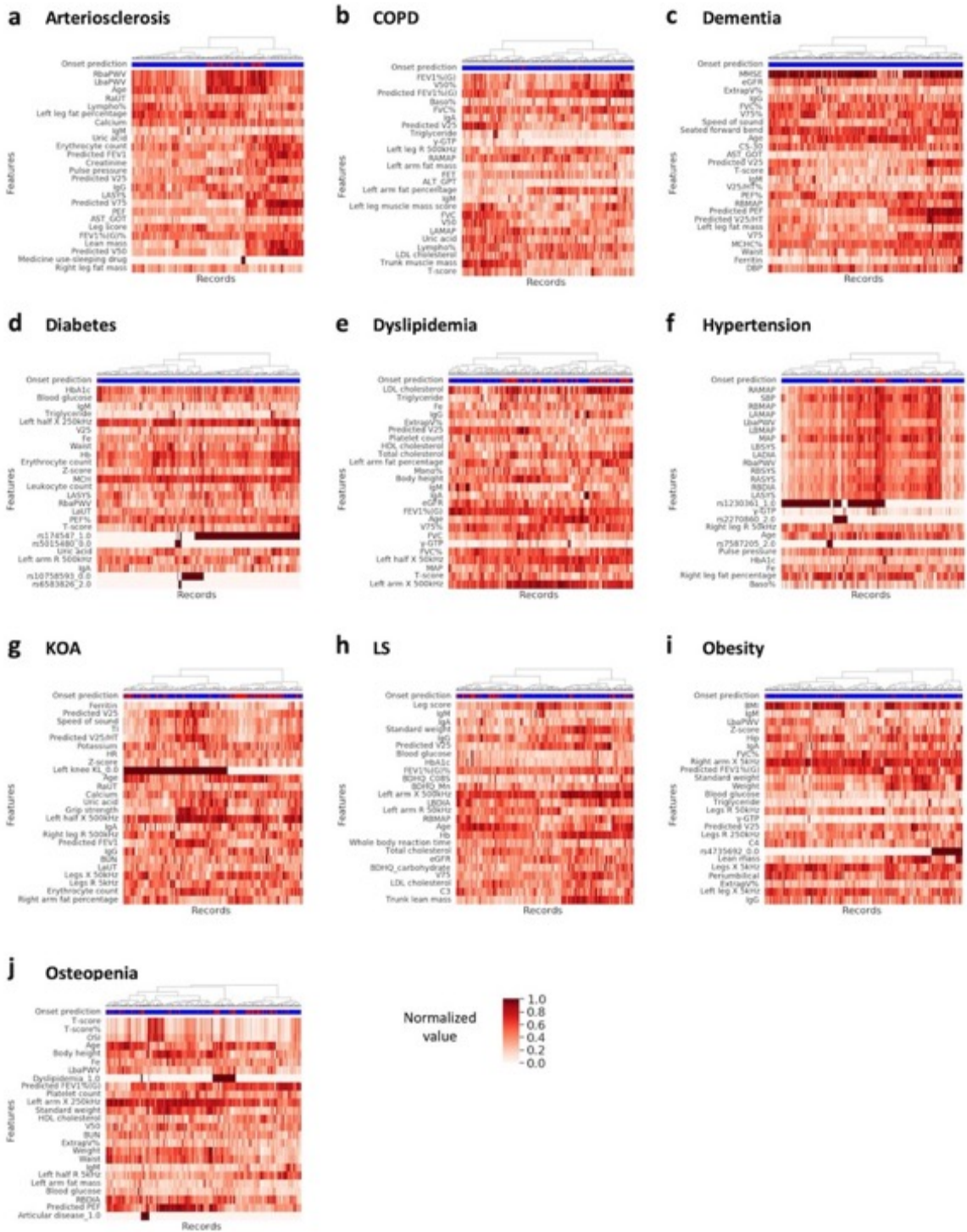

**Supplementary Figure 17. Hierarchically-clustered heatmap for original explanatory variables of records in the dataset.** Explanatory variables were normalized into a range of 0–1 before clustering. In the heatmaps, red predicted labels represent records with onset prediction and blue with non-onset. **a** Arteriosclerosis, **b** chronic obstructive pulmonary disease (COPD), **c** dementia, **d** diabetes, **e** dyslipidemia, **f** hypertension, **g** knee osteoarthritis (KOA), **h** locomotive syndrome (LS), **i** obesity, and **j** osteopenia.

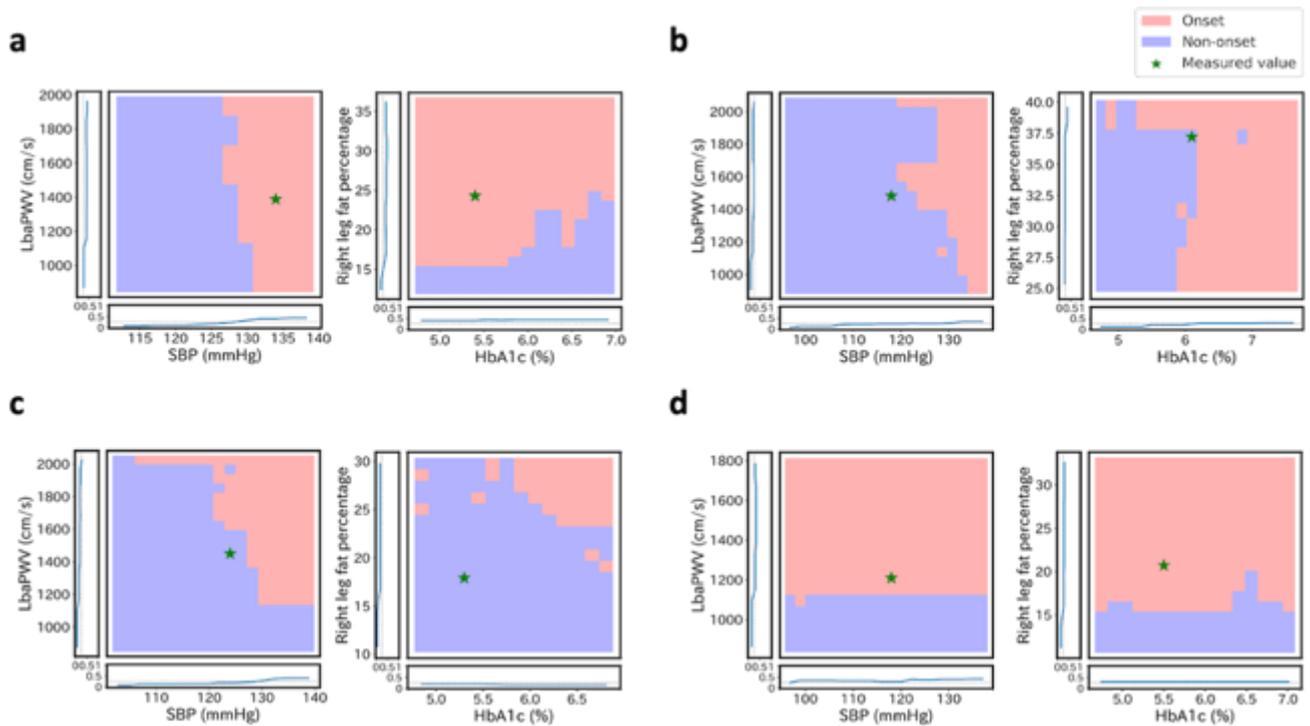

**Supplementary Figure 18. Examples of p-mICE-based HDPDs for hypertension prediction model.** The phase diagrams are shown along with the individual conditional expectation (ICE) plots. The dotted line in the ICE plot represents the onset threshold of the predicted probability in the prediction model. **a**–**d** HDPDs of four randomly selected records. For each record, HDPDs of the combination of systolic blood pressure (SBP)–left brachial-ankle pulse wave velocity (LbaPWV) and HbA1c–right leg fat percentage are displayed.

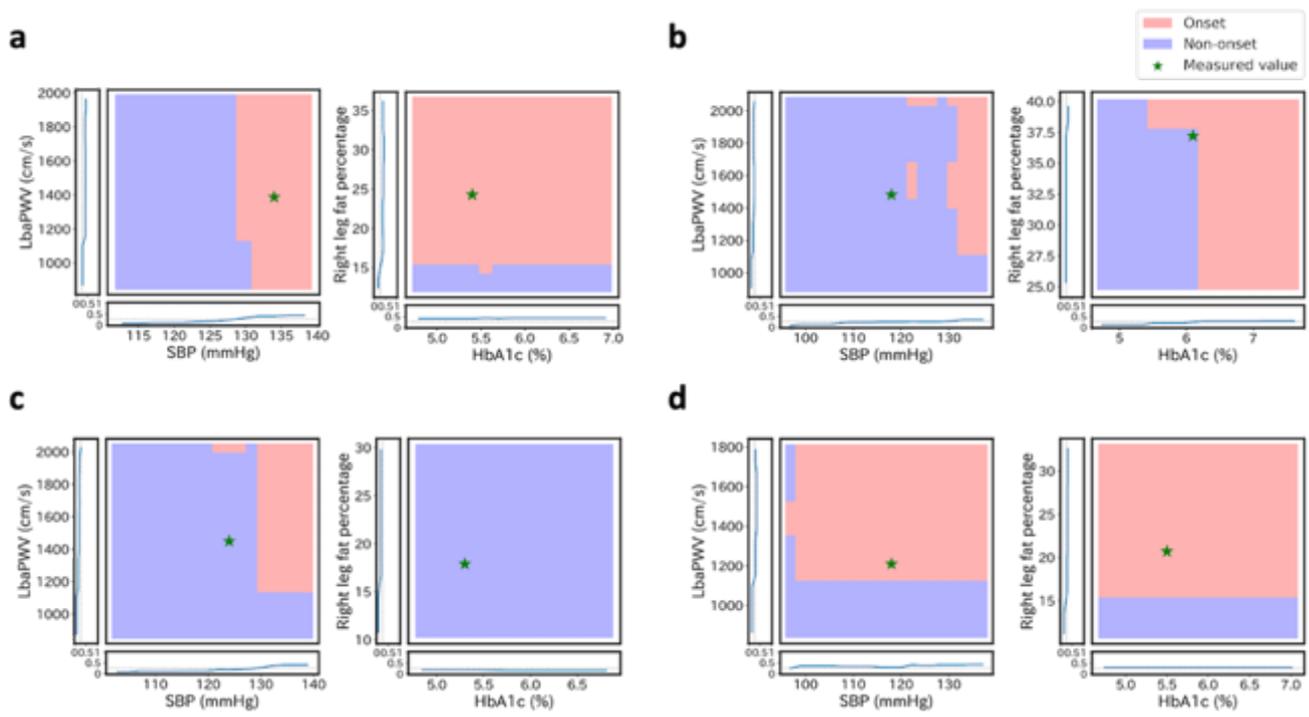

**Supplementary Figure 19. Examples of 2d-ICE-based HDPDs for hypertension prediction model.** The phase diagrams are shown along with the individual conditional expectation (ICE) plots. The dotted line in the ICE plot represents the onset threshold of the predicted probability in the prediction model. **a**–**d** HDPDs of four records. The records are identical to Supplementary Fig. 18, and the symbols correspond. For each record, HDPDs of the combination of systolic blood pressure (SBP)–left brachial-ankle pulse wave velocity (LbaPWV) and HbA1c–right leg fat percentage are displayed.

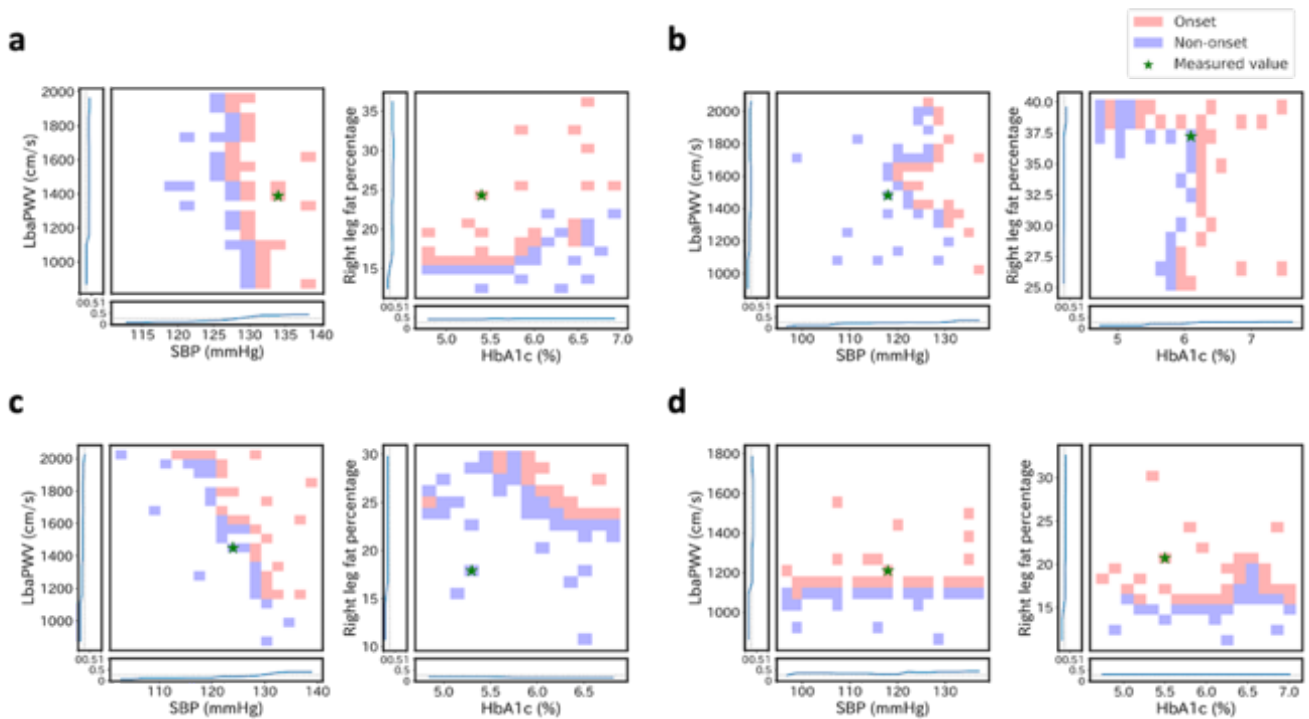

**Supplementary Figure 20. Examples of p-mICE-based HDPDs using active learning for hypertension prediction model.** The phase diagrams using active learning are shown along with the individual conditional expectation (ICE) plots. The dotted line in the ICE plot represents the onset threshold of the predicted probability in the prediction model. **a**–**d** HDPDs of four records. The records are identical to Supplementary Fig. 18, and the symbols correspond. For each record, HDPDs of the combination of systolic blood pressure (SBP)–left brachial-ankle pulse wave velocity (LbaPWV) and HbA1c–right leg fat percentage are displayed.

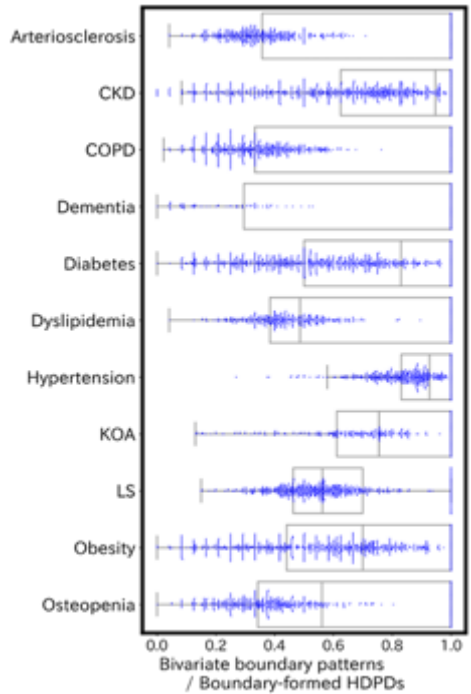

**Supplementary Figure 21. Bee swarm plot for proportion of bivariate boundary patterns in boundary-formed 2d-ICE HDPDs for a record.** The proportion of bivariate boundary patterns in boundary-formed phase diagrams was calculated for each record. In box-plots, the center line represents the median; box limits, upper and lower quartiles; whiskers, 1.5x interquartile range. CKD, chronic kidney disease; COPD, chronic obstructive pulmonary disease; KOA, knee osteoarthritis; LS, locomotive syndrome.

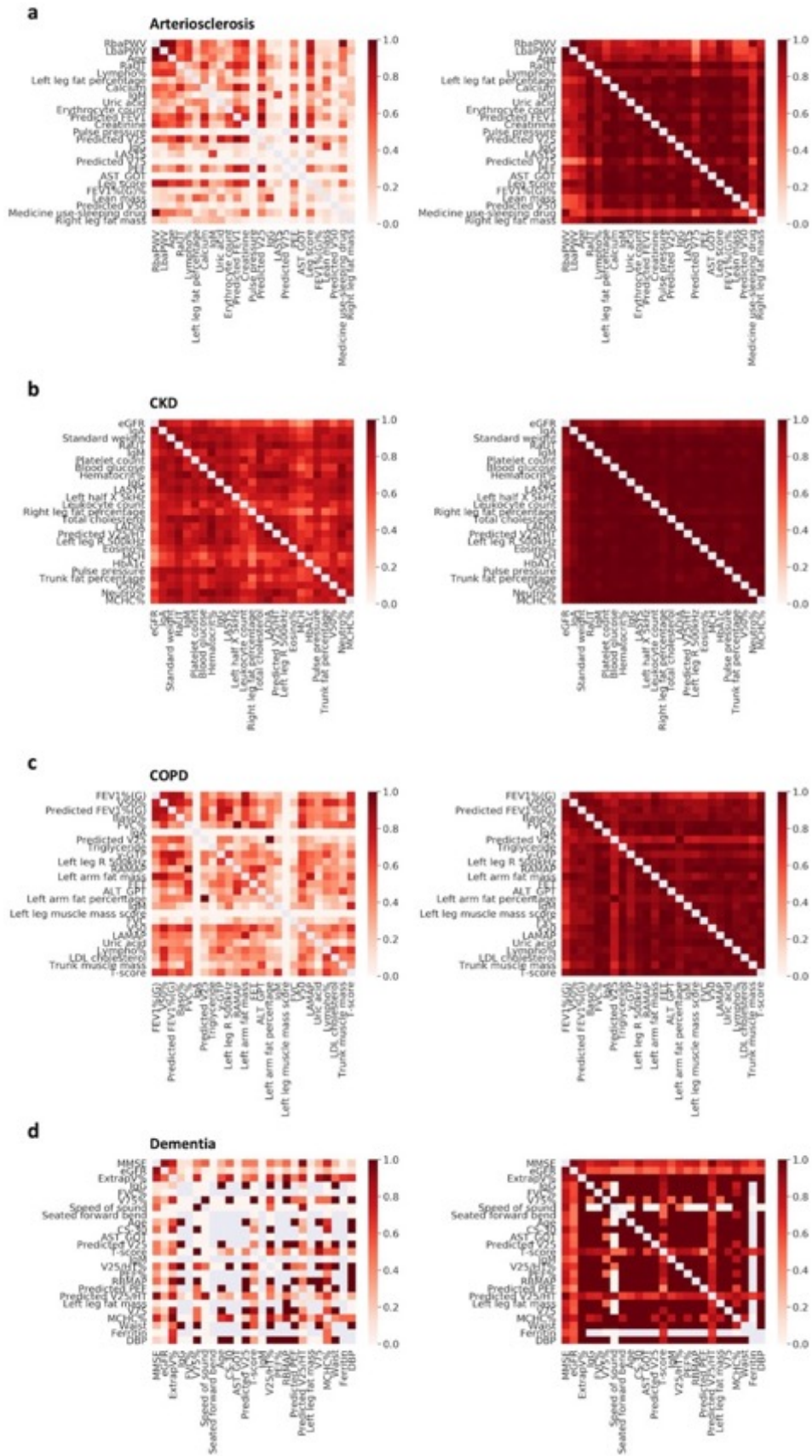

**Supplementary Figure 22 (continued).**

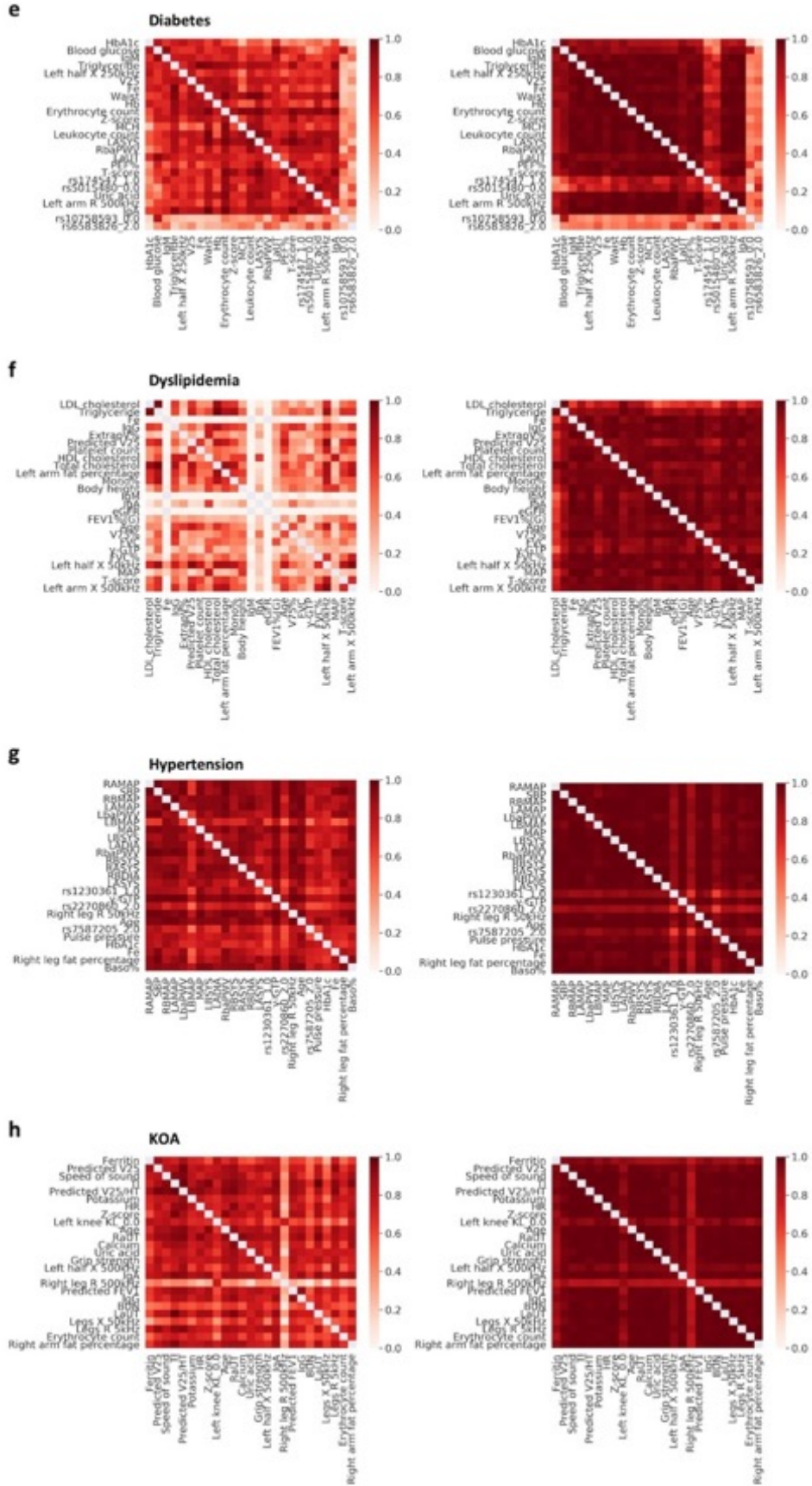

**Supplementary Figure 22 (continued).**

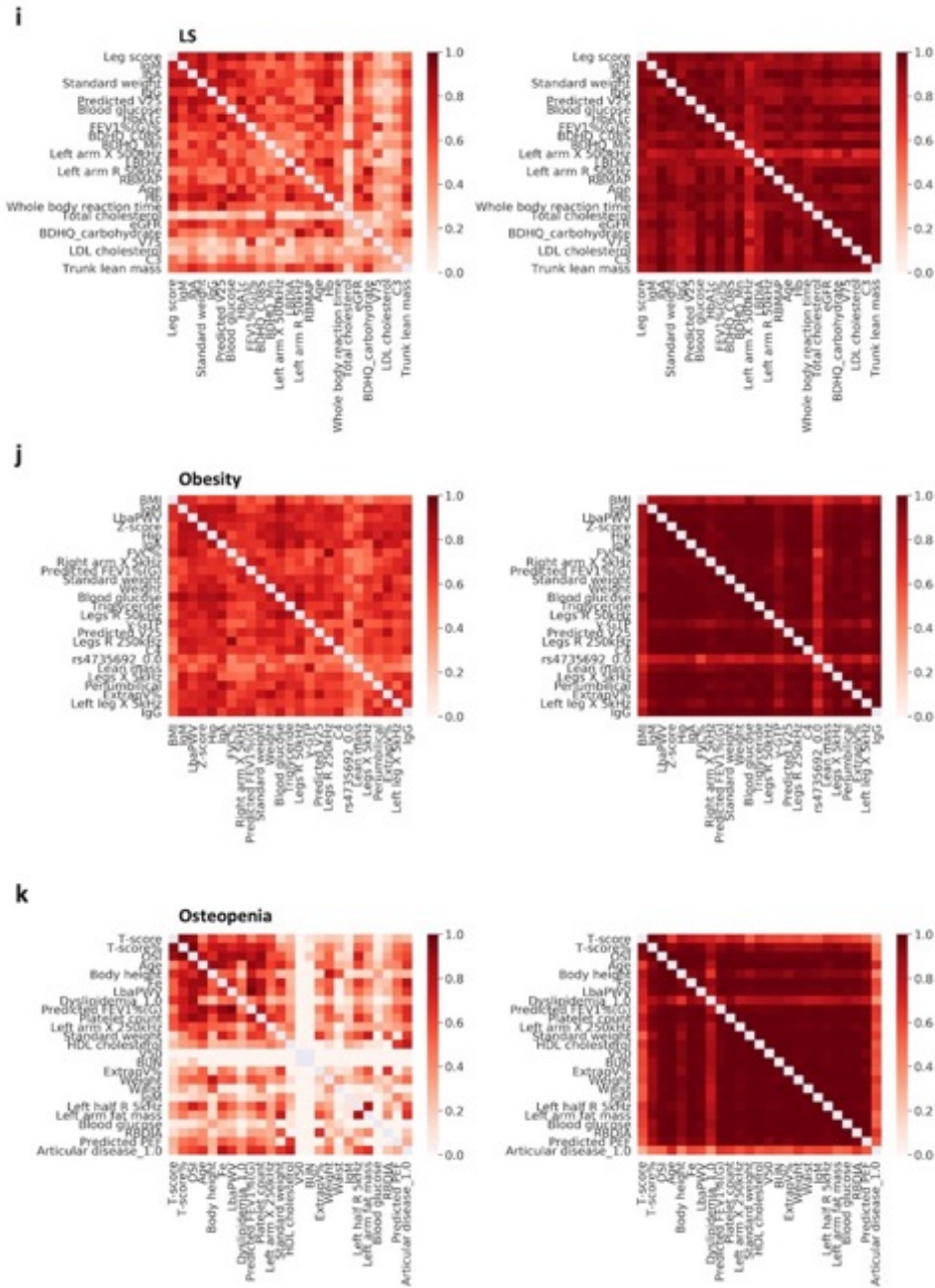

**Supplementary Figure 22. Heatmap for proportion of bivariate boundary patterns in boundary-formed HDPDs for each combination of intervention variables.** The proportion of bivariate boundary patterns in the boundary-formed phase diagrams in the test data are shown for each combination of intervention variables. **a** Arteriosclerosis, **b** chronic kidney disease (CKD), **c** chronic obstructive pulmonary disease (COPD), **d** dementia, **e** diabetes, **f** dyslipidemia, **g** hypertension, **h** knee osteoarthritis (KOA), **i** locomotive syndrome (LS), **j** obesity, and **k** osteopenia. Regarding each disease, the left heatmap is constructed based on 2d-ICE-based HDPDs, and the right one is based on p-mICE-based phase diagrams.

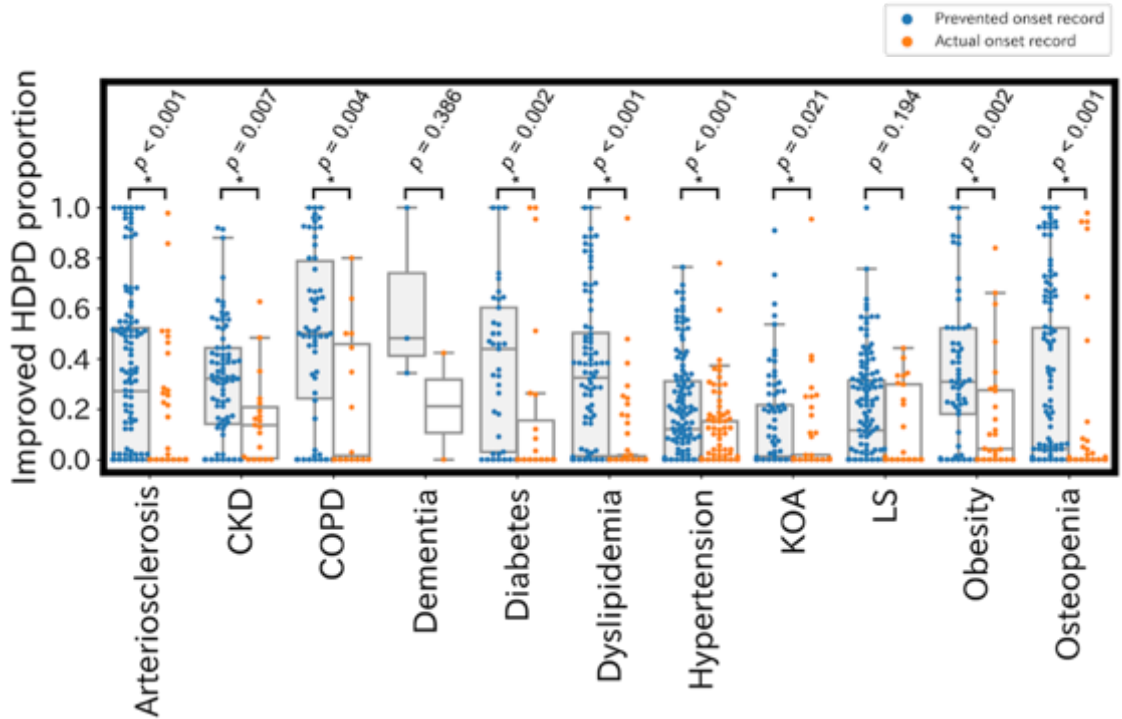

**Supplementary Figure 23. Evaluation for validity of 2d-ICE-based personal boundaries as intervention goals.** Improved HDPD proportion was compared between prevented onset records and actual onset records. The numbers of prevented onset records and actual onset records for each disease are shown in Table 2. Statistical significance was calculated using two-sided Wilcoxon rank-sum test. $p < 0.05$ is indicated by *. In box-plots, the center line represents the median; box limits, upper and lower quartiles; whiskers, 1.5x interquartile range.

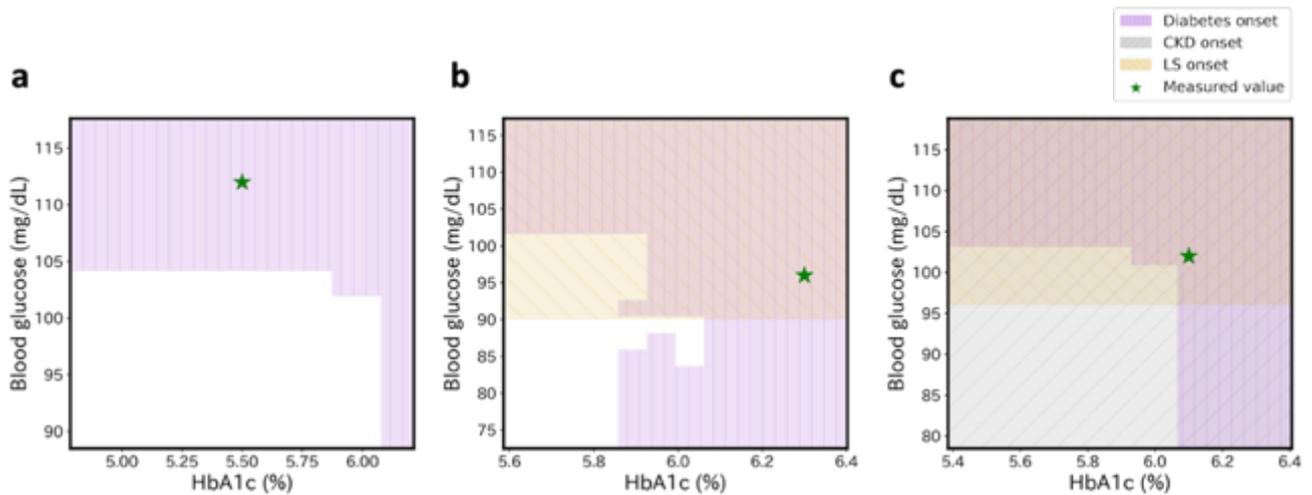

**Supplementary Figure 24. Superimposition of onset area in phase diagrams for multiple diseases.** The onset areas in HDPDs were superimposed for diabetes, chronic kidney disease (CKD), and locomotive syndrome (LS), where HbA1c and blood glucose were selected as model features. We assumed a scenario where interventions would be made on records with diabetes onset prediction (**a–c**). While some records only required improvements based on the predictive onset of diabetes (**a**), other records required simultaneous prevention of multiple diseases (**b**). Also, some cases represented that simultaneous prevention of all diseases could not be expected when HbA1c and serum blood glucose were selected as intervention variables (**c**).

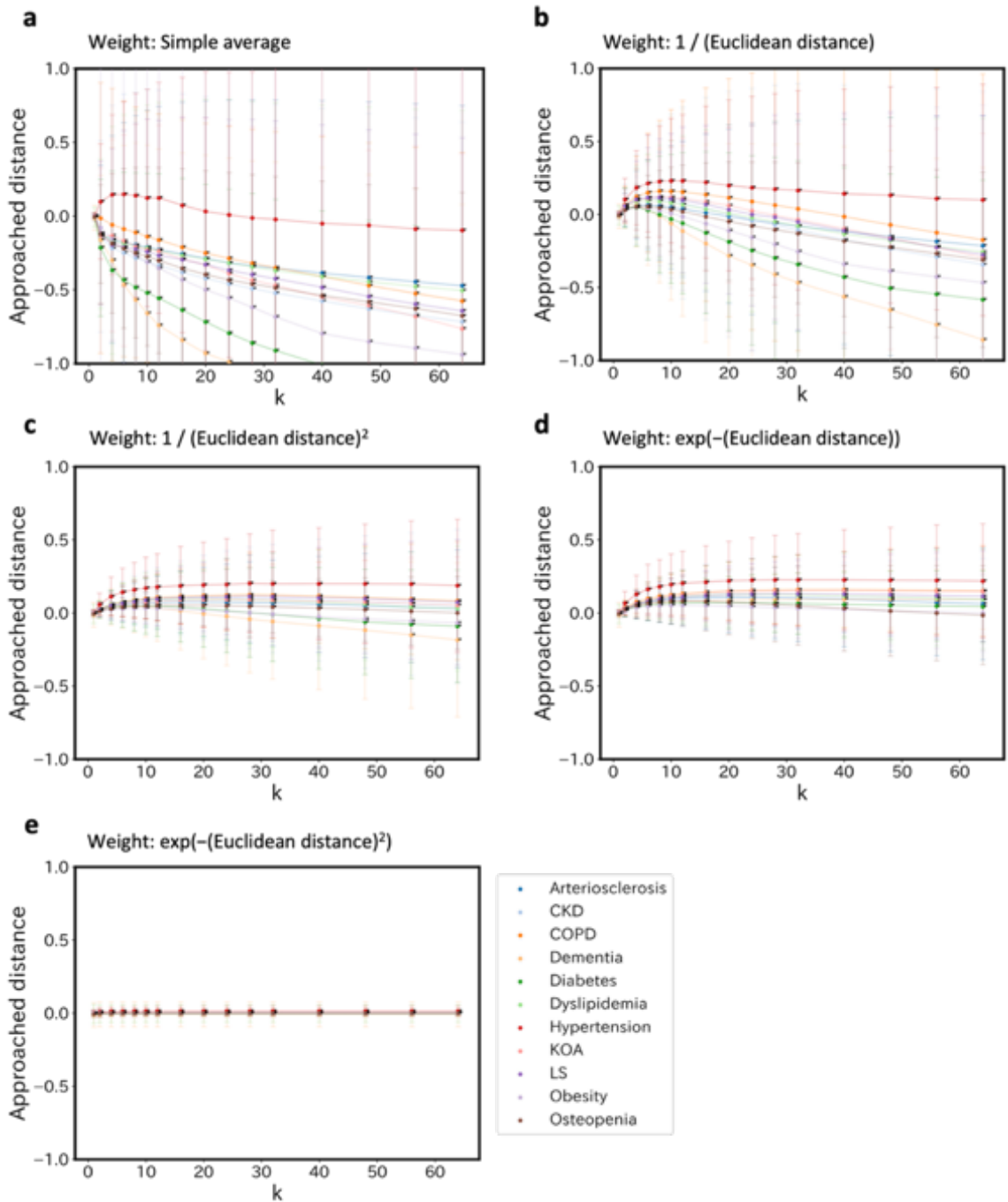

**Supplementary Figure 25. Approached distance at various averaging weights in training data.** The distance to the actual future records was calculated for each of 2d-ICE-applied data and p-mICE-applied data in the normalized variable space. From the difference of these distances, the distance approached by the projection step of the p-mICE was calculated for each record. $k$-nearest neighbor records were subjected to the projection in the p-mICE. Statistical significance was calculated using two-sided paired $t$-test. $p < 0.05$ is indicated by *. CKD, chronic kidney disease; COPD, chronic obstructive pulmonary disease; KOA, knee osteoarthritis; LS, locomotive syndrome.